# LLMs learn scientific taste from institutional traces across the social sciences

Ziqing Gong[1], Ning Li[1*], Huaikang Zhou[1]
[1] School of Economics and Management, Tsinghua University, Beijing, China
Corresponding author. Email: lining@sem.tsinghua.edu.cn

## Abstract

Reinforcement-learned reasoning has powered recent AI leaps on verifiable tasks — mathematics, code, structure prediction. The harder bottleneck is *evaluative judgment* in low-verifiability domains, where no oracle anchors reward and which untested ideas deserve attention is the question. We test whether *institutional traces* — the record of what fields published, where, and at which tier — can serve as a training signal for AI evaluators. Across eight social science disciplines (psychology, economics, communication, sociology, political science, management, business and finance, public administration), we built held-out four-tier research-pitch benchmarks and fine-tuned LLMs on field-specific publication outcomes. The fine-tuned models cleared the 25% chance baseline and exceeded frontier-model performance by wide margins, with best single-model accuracy ranging from 55.0% in public administration to 85.5% in psychology. In management — evaluated against 48 expert gatekeepers, 174 junior researchers, and 11 frontier reasoning models — the best single fine-tuned model (Qwen3-4B) reached 59.2%, 17.6 percentage points above expert majority vote (41.6%, non-tied) and 28.1 above the frontier mean (31.1%). The fine-tuned models also showed calibrated confidence: confidence rose when predictions were correct and fell when wrong, mirroring how a skilled reviewer can say "I'm sure" versus "I'm guessing". Selective triage on this signal reached very high accuracy on the highest-confidence subsets in every field. Reinforcement learning with chain-of-thought plateaued below direct fine-tuning in the management mechanism probe, and a contemporaneous GPT-5.5 comparison found no accuracy gain from high-reasoning inference over chat/log-probability classification across all eight fields. The approach offers a cost-effective way to train evaluative models from institutional records. Institutional traces, we conclude, encode a scalable training signal for the low-verifiability judgment on which science depends.

## Introduction

Artificial intelligence has advanced fastest in scientific domains where candidate outputs can be checked. Protein structures can be matched against experimental constraints[1]; mathematical proofs can be verified by formal systems[2]; computer code can be executed against tests[3]. The remarkable AI-for-science results of the last five years all share this property: an unambiguous oracle exists, and the model's job is to find an output the oracle accepts.

Most of science, however, depends on a prior judgment with no such oracle. Before any experiment, proof, or benchmark exists, scientists must decide which untested ideas deserve scarce attention. This evaluative judgment, the capacity sometimes called *taste*, is what editors, reviewers, hiring committees, and grant panels enact whenever they choose what to pursue and what to set aside. As AI systems generate hypotheses, analyses, and manuscripts at unprecedented scale[4, 5], the bottleneck in science is shifting from production toward evaluation[6]: the rate-limiting step is no longer writing the next plausible idea, but deciding which plausible idea is worth working on. Scientific taste defines the next critical boundary in artificial intelligence, where capability drops sharply from domains with clear answers to those requiring discrimination and judgment under uncertainty[7].

Evaluative judgment is a paradigmatic case of *collective tacit knowledge*: understanding embedded in institutional practices that no individual participant fully articulates, yet that the system reliably enacts[8, 9]. Individual peer reviewers agree on quality assignments at barely above chance (meta-analytic Fleiss' kappa of 0.17 across 48 studies[10]), and neither career stage nor editorial experience improves this reliability[11, 12]. Ethnographic evidence confirms that academic judgment operates through intuitive assessment and disciplinary sensibility, not rule application[13]. Yet the same institutional system, integrating thousands of such noisy judgments over decades, produces consistent quality stratification across publication tiers[14]. That apparent paradox, agreement-poor at the individual level and signal-rich at the institutional level, is the central observation that motivates this work.

Frontier language models confront the same boundary. They can summarize, reason, and write fluent reviews, but prompting them with evaluative criteria does not by itself produce selective judgment: prior AI-review studies document bias and blind spots, and the present evaluations show frontier models clustering near chance, over-predicting favorable categories, and compressing judgments into middle tiers[15, 16, 17, 18]. The results below argue against a prompt-engineering-only explanation and point instead to a transmission problem. Tacit standards do not survive the channel of explicit instruction, and reinforcement learning from human preferences pushes models toward agreeable, lenient assessments[19, 20].

We propose institutional traces (the historical record of what fields published, where, and in which prestige tier) as an *alignment signal* for AI evaluators when no verifiable reward is available. Individual reviewers are noisy, but fields *repeatedly* select, reject, and rank work through editorial decisions and journal hierarchies. These records are imperfect proxies for quality, but they preserve a learnable trace of field-level evaluative practice. If a model can be aligned to those traces, then low-verifiability scientific judgment may be made computationally accessible without first reducing it to a written rubric. Figure 1 summarizes this conceptual frame.

This paper is empirically novel against an active literature. Recent work has used reinforcement learning from verifiable reward to push frontier reasoning on math, code, and other check-able tasks[21]; specialized continued pretraining has produced LLMs that surpass human experts at predicting empirical neuroscience outcomes[22]; LLMs have been shown to generate research ideas judged more novel than expert-generated ones[4], to produce reviewer-quality qualitative feedback on Nature-family submissions[23], and to power end-to-end automated research pipelines that can complete AI-research workflows and pass workshop peer review[24, 25]. Concurrent work by Tong et al.[26] also pursues "scientific taste" through a related framing, using reinforcement learning from community feedback with citation counts on accepted papers as the supervision signal. Citations are a problematic proxy for evaluative judgment at the moment of decision: they accumulate over years, are confounded by venue prestige and author networks, and are not part of what reviewers see when deciding whether an idea is worth pursuing. Institutional selection traces (which venues accepted which work in the first place) capture the prior gatekeeping decision itself, which is the variable evaluators are actually computing. None of the prior systems treats *institutional selection traces* as a transferable training signal for an AI *evaluator* in a low-verifiability domain, and none has shown that such an evaluator generalizes across disciplines or matches the pattern that the present results display: direct fine-tuning on outcome traces beats reinforcement-learned chain-of-thought on the same task, the inverse of the verifiable-reward result. The empirical gap we close is the conjunction of (i) cross-field generalization across eight social science disciplines, (ii) head-to-head comparison with senior human gatekeepers in the field where we could recruit them, (iii) calibrated selective prediction in a low-verifiability evaluative task, and (iv) a mechanism finding in which direct fine-tuning beats RL, dual to the verifiable-reward paradigm.

Here we test whether fine-tuning on institutional traces can teach LLMs to discriminate research-idea quality across the social sciences. We construct held-out four-tier research-pitch benchmarks in eight disciplines (psychology, economics, communication, sociology, political science, management, business and finance, and public administration), fine-tune three cost-effective models per field, and compare them against architecture-matched base controls and frontier reasoning systems. In management, where two of the authors hold domain expertise and could recruit qualified evaluators, we additionally compare against 48 expert gatekeepers, 174 junior researchers, and an 11-model frontier reasoning cohort, and conduct deeper mechanism probes. This design uses cross-field breadth to establish the phenomenon and a management deep dive to benchmark it against humans and mechanism probes. Four claims follow from the data. (1) Institutional traces are learnable across all eight fields. (2) In management, the field where human benchmarks were collected, fine-tuned models recover judgment beyond both expert majority vote and the frontier reasoning baseline. (3) The model knows what it knows: higher-confidence subsets are substantially more accurate than the full prediction set, enabling triage that fast-tracks the most reliable cases and routes the genuinely uncertain remainder to human review. (4) In the management mechanism probe, the signal is robust under input compression and temporal shift, with explicit boundary conditions. Scientific taste, the results suggest, was never purely individual. It was always being deposited in the institutional record, awaiting a learning procedure simple enough to extract it.

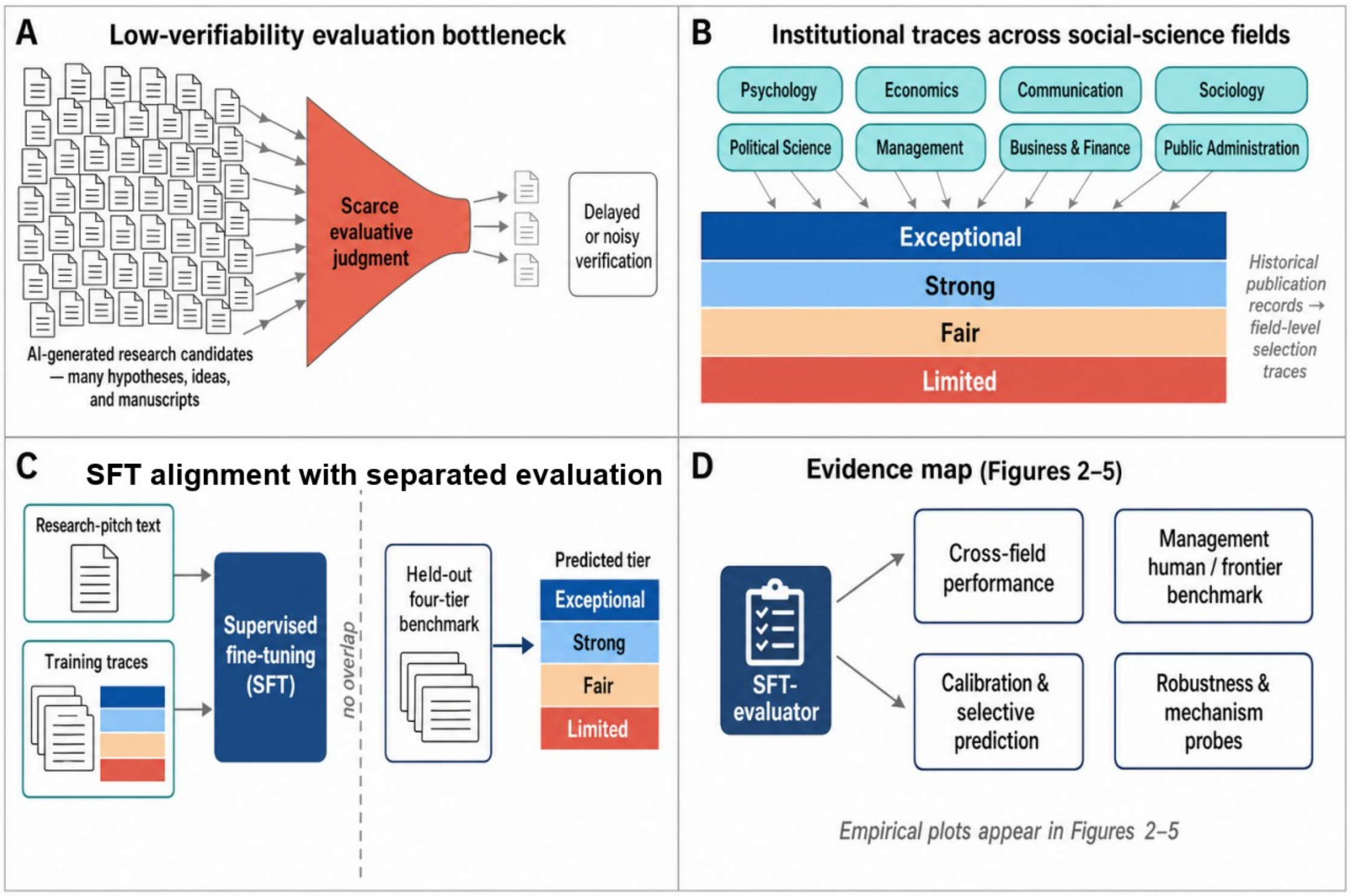

**Figure 1 | Institutional traces as a training signal for evaluative judgment in low-verifiability science.**
a, Low-verifiability evaluation bottleneck: AI systems can generate many plausible hypotheses, ideas, and manuscripts, but scarce evaluative judgment determines which candidates receive attention before delayed or noisy verification arrives. b, Institutional traces across eight social-science fields: historical publication records encode field-level selection into four prestige tiers (exceptional, strong, fair, limited). c, SFT alignment with separated training and evaluation: research-pitch text and tier-labeled training traces are used to supervised-fine-tune an evaluator, which is then evaluated on non-overlapping four-tier benchmarks. d, Evidence map linking the SFT evaluator to the empirical tests in Figures 2–5. Benchmark items were excluded from SFT training; all source articles were published after 30 June 2025, but proprietary frontier-model pre-training exposure cannot be independently verified.

## Methods

**Benchmarks.** We constructed balanced four-tier evaluation benchmarks in eight social science fields: management (organizational psychology and management), economics, business and finance, communication, political science, psychology, public administration, and sociology. Field scope follows Web of Science discipline categories. All source articles were published after 30 June 2025 and all benchmark items were excluded from SFT training; because frontier providers do not disclose complete training corpora, this date screen is a temporal-control measure rather than a guarantee of exclusion from every proprietary model. Management contained 120 pitches (30 per tier) drawn from a 19-journal source universe; the smaller size reflects the recruitment ceiling of the matched expert panel. The remaining seven fields each contained 200 pitches (50 per tier) drawn from field-specific journal universes. Benchmark tiers (exceptional, strong, fair, limited) are properties of the source journal, assigned by a human pre-construction procedure that is logically separate from the AI evaluation. In management, the management subject-matter experts on the author team directly assigned each of the 19 source journals to a tier — no external numerical proxy was used, since the management hierarchy is sufficiently codified in tenure and editorial-board norms — consulting additional field experts on boundary cases. In the other seven fields, field experts nominated candidate journals for each tier (drawing on editorial standing, tenure-credit norms, and submission/acceptance patterns), and the nominations were cross-checked against an external benchmark of citation impact metrics and recognized journal-quality lists used by tenure committees and field associations; inconsistencies between expert nomination and the external benchmark were resolved through additional expert consultation, iterated until convergence. The author team's domain experts retained final decision authority on all tier boundaries. Crucially, AI models being evaluated never see journal identity — they predict tier from research-pitch text alone (full per-field mappings in SI Tables ST3–ST10). Each source article was transformed into a standardized research-pitch text using a fixed extraction workflow that exposes only the core research question and theoretical framing; methods, empirical results, journal identity, and author identity were removed, to isolate idea-level assessment across all evaluator classes. Throughout, "publication tier" is treated as an institutional proxy for field-level evaluation, not as ground-truth quality. Full benchmark construction, Web of Science scoping, and journal-mapping detail are in SI SM5.

**Supervised fine-tuning.** In management, where we conducted the full human, frontier-model, mechanism, and transfer analyses, we fine-tuned four architectures: GPT-4.1, GPT-4.1-nano, Qwen3-30B-A3B-Instruct (30B-parameter mixture-of-experts, 3B active at inference), and Qwen3-4B-Instruct (4B-parameter dense). The four management SFT models occupied a narrow in-domain accuracy band (55.0–59.2%), while GPT-4.1 accounted for most external fine-tuning spend; we therefore used the three cost-effective architectures (Qwen3-30B-A3B, Qwen3-4B, and GPT-4.1-nano) for the full eight-field core. GPT-4.1 was retained only for management mechanism probes (pairwise discrimination and input-compression transfer) and the cross-field-transfer test, where larger representational capacity was expected to matter. Architecture-matched un-fine-tuned base controls were evaluated on every field's benchmark to isolate the SFT effect. Each training example paired a research-pitch text, presented through a pre-specified field-agnostic evaluation prompt, with a single tier-label completion token. Training minimized label-token negative log-likelihood with input tokens masked from gradient updates, forcing the model to learn the mapping from research content to quality tier. Training corpora were field-specific and disjoint from evaluation benchmarks: management used 4,479 pitch–outcome pairs; the seven other fields used 2,094–5,593 pairs each (per-field sizes in SI Table ST11). Qwen checkpoints were trained locally via TRL (learning rate 1e-5 for the 30B model, defaults for 4B), requiring approximately eight and one A100-GPU-hours per field for the 30B and 4B models respectively. GPT-4.1-nano was trained via the OpenAI fine-tuning API (3 epochs, auto-selected hyperparameters; ~$10 per field). Total fine-tuning cost across the

eight-field, three-architecture core was therefore approximately $80 in API spend (eight GPT-4.1-nano fine-tunes) plus ~72 A100 GPU-hours of local compute (eight 30B-A3B + eight 4B fine-tunes); the additional management-only GPT-4.1 fine-tune contributed approximately $200 in API spend. This design uses smaller cost-effective evaluators for the breadth result while preserving a larger-model probe for generalization tests; later analyses show that model scale buys out-of-format and cross-field robustness despite similar in-domain performance. Full hyperparameters and decoding settings are in SI SM1.

**Frontier model evaluation.** In management, we evaluated a March 2026 cohort of 11 frontier reasoning models under one pre-specified expert-derived instruction scaffold: Gemini 3.1 Pro, Claude Opus 4.6, GPT-5.2 High, Gemini 2.5 Pro, Qwen 3.5 Plus, Grok 4.1 Fast, Kimi K2.5, DeepSeek V3.2, GLM-5, Seed 2.0 Pro, and MiniMax M2.5. Each model was sampled eight times per pitch; the primary metric was pitch-mean eight-sample accuracy. Separately, in May 2026, we ran an additional GPT-5.5 High evaluation across all eight fields; GPT-5.5 High, together with Gemini 3.1 Pro, serves as the cross-field frontier-control pair. Management GPT-5.5 High used the expert-derived management rubric prompt; the seven other fields used the field-agnostic social-science evaluation prompt. GPT-5.5 chat/log-probability classification was also run across all eight fields with reasoning disabled to support confidence and selective-prediction analyses. Full prompt text and per-model decoding parameters are in SI SM2 and ST19.

**Human evaluation.** The human study was approved by the Tsinghua University institutional review board (Project No. THU-04-2026-0034). Two panels evaluated the management benchmark under the same four-tier rubric used for AI models. The expert panel size was set by the empirical recruitment ceiling for senior gatekeepers in management; we maximized junior-researcher recruitment to quantify how much aggregation (wisdom-of-crowds size) lifts a noisy individual signal. The expert panel comprised 48 gatekeepers (current journal editors and editorial board members) recruited through direct professional contact, contributing 384 ratings (8 pitches per rater; mean 3.2 ratings per pitch). The junior panel comprised 174 doctoral and postdoctoral researchers contributing 2,530 ratings (mean 14.5 pitches per rater; mean 21.1 ratings per pitch); raters spending fewer than one minute per pitch were excluded as perfunctory. Compensation was 100 RMB and/or access to a research tool. For each pitch, raters reported tier assignment using field-familiar labels (Top, Top−, Good, Fair, deterministically mapped to the unified four-tier scheme), confidence (5-point Likert), and topic familiarity (5-point Likert). Recruitment, panel demographics, and full rating instruments are detailed in SI SM3. Human benchmarks were collected only in management, the field where the authors' domain expertise enabled recruitment of qualified expert evaluators, and SFT accuracies in other fields are interpreted against this management anchor.

**Ensemble pair selection.** A two-model probability-averaging ensemble (GPT-4.1-nano + Qwen3-30B-A3B) was evaluated as a sensitivity analysis on the best single fine-tuned model; alternative pairings yield comparable accuracy and are reported in SI SM6.

**Exploratory transfer probe.** As a secondary robustness analysis, all four management-trained fine-tuned models (GPT-4.1, GPT-4.1-nano, Qwen3-30B-A3B, Qwen3-4B) were evaluated on the held-out benchmarks of the seven other fields with no target-field training. This probe tests whether any component of the management-learned signal transfers without field-specific traces; it is not used as evidence for the primary cross-field claim, which rests on field-specific fine-tuning and matched base/frontier controls in each field. Full per-architecture × per-field cross-transfer accuracies (with Wilson 95% CIs) are reported in SI SM9 and Supplementary Table ST17.

**Statistical analyses.** The primary endpoint was four-class exact-match accuracy. Bootstrap resampling (10,000 draws, procedure described in SI SM4) produced 95% confidence intervals on accuracy and macro-F1. Paired evaluator comparisons used exact McNemar tests on correctness vectors. Above-chance tests used exact binomial tests for single field-level prediction lists; cross-field SFT-versus-frontier summaries are reported as descriptive field-level contrasts with bootstrap uncertainty over fields.

Frontier-cohort heterogeneity used Cochran's Q. Inter-rater reliability used Fleiss' kappa (categorical) and Krippendorff's alpha (ordinal). Confidence-gap significance between correct and incorrect predictions used Mann–Whitney U at the rating level. Calibration was quantified via expected calibration error (ECE) at the model–field level; we report ECE for GPT-4.1-nano in the main text because it consistently produced the lowest cross-field ECE among the three SFT architectures and therefore serves as a principled calibration anchor. Per-architecture ECEs are in SI Table ST2. Majority-vote results excluded tied pitches and report effective non-tied sample sizes. Monte Carlo matched-$N$ analyses (5,000 draws) compared junior and expert majority voting at equivalent panel sizes. Multiple-testing corrections are noted where applied; focused pairwise comparisons report raw unadjusted $P$ values. Full statistical detail and sensitivity analyses are in SI SM4.

**Code and data availability.** Benchmark pitches, anonymized human ratings, AI predictions, training-corpus summaries, journal-tier mappings, analysis code, and reproduction scripts are provided in the accompanying public reproducibility package (https://github.com/FutureTech-OB/ai-taste) under CC BY 4.0 for data and MIT for code. Public model aliases and provider/model access information are included where redistribution is permitted; model weights or provider-specific checkpoint identifiers are not redistributed where licensing or provider terms prohibit redistribution. Released frontier-model outputs can be checked against the provided prediction records and documented prompts and decoding parameters in SI SM2; exact reruns may vary with provider-side model changes and sampled reasoning outputs.

## Results

### 1. Institutional traces are learnable across eight social science fields

Across all eight fields, fine-tuning on institutional traces produced above-chance evaluative discrimination. The cross-field core comprised three cost-effective architectures trained in every field, while management additionally included GPT-4.1 to support the human-comparator, mechanism-probe, and transfer analyses. All 24 core fine-tuned model–field combinations beat their architecture-matched base controls, with a mean fine-tuned-minus-base lift of +35.0 percentage points across the full eight-field core (+35.2 pp across the seven non-management fields). Mean fine-tuned accuracy across the seven non-management fields was 65.0% for Qwen3-30B-A3B, 61.1% for GPT-4.1-nano, and 58.1% for Qwen3-4B, substantially above the 25% chance baseline and above the ~32% mean accuracy of frontier reasoning controls. Best single-model accuracy ranged from 55.0% in public administration to 85.5% in psychology (Figure 2). Cross-field fine-tuned and best base/frontier control accuracies for all eight fields are tabulated in SI Table ST1.

Three lines of evidence indicate that this is a genuine cross-field phenomenon, not an artifact of any one model or dataset, and argue against the simplest "the model is just predicting journal tier from surface cues" alternative. First, *cross-architecture convergence*: a 30-billion-parameter mixture-of-experts model, a 4-billion-parameter open-source dense model, and a commercial nano-scale model not only all beat their base controls in every field, they showed the same broad difficulty gradient (psychology and economics high; public administration and business and finance low), although exact field rankings varied by architecture. This cross-architecture pattern places the signal in the data, not in any specific model. Pairwise agreement among the fine-tuned management models was $\kappa$ = 0.50–0.60, descriptively an order of magnitude above the multi-rater Fleiss' $\kappa$ of 0.03–0.05 across our human panels (Supplementary Table ST21). Second, *the universality of the fine-tuning lift*: 24 of 24 fine-tuned-versus-base model–field comparisons were positive. Across all eight fields, architecture-matched base controls and the two frontier controls remained close to chance under explicit evaluative prompting (Gemini 3.1 Pro: 34.3%; GPT-5.5 High: 30.9%, field-unweighted means), even though their pre-training data plausibly includes the journal-prestige hierarchies themselves. If institutional tier were learnable from surface cues alone, base models would not sit at chance. Third, *the size of the lift relative to cost*: the entire eight-field, three-architecture training run cost approximately $80 in API spend plus ~72 A100 GPU-hours, with field-specific training corpora ranging from roughly 2,100 to 5,600 pitch–outcome pairs. Institutional traces appear to be an efficient and broadly available alignment signal.

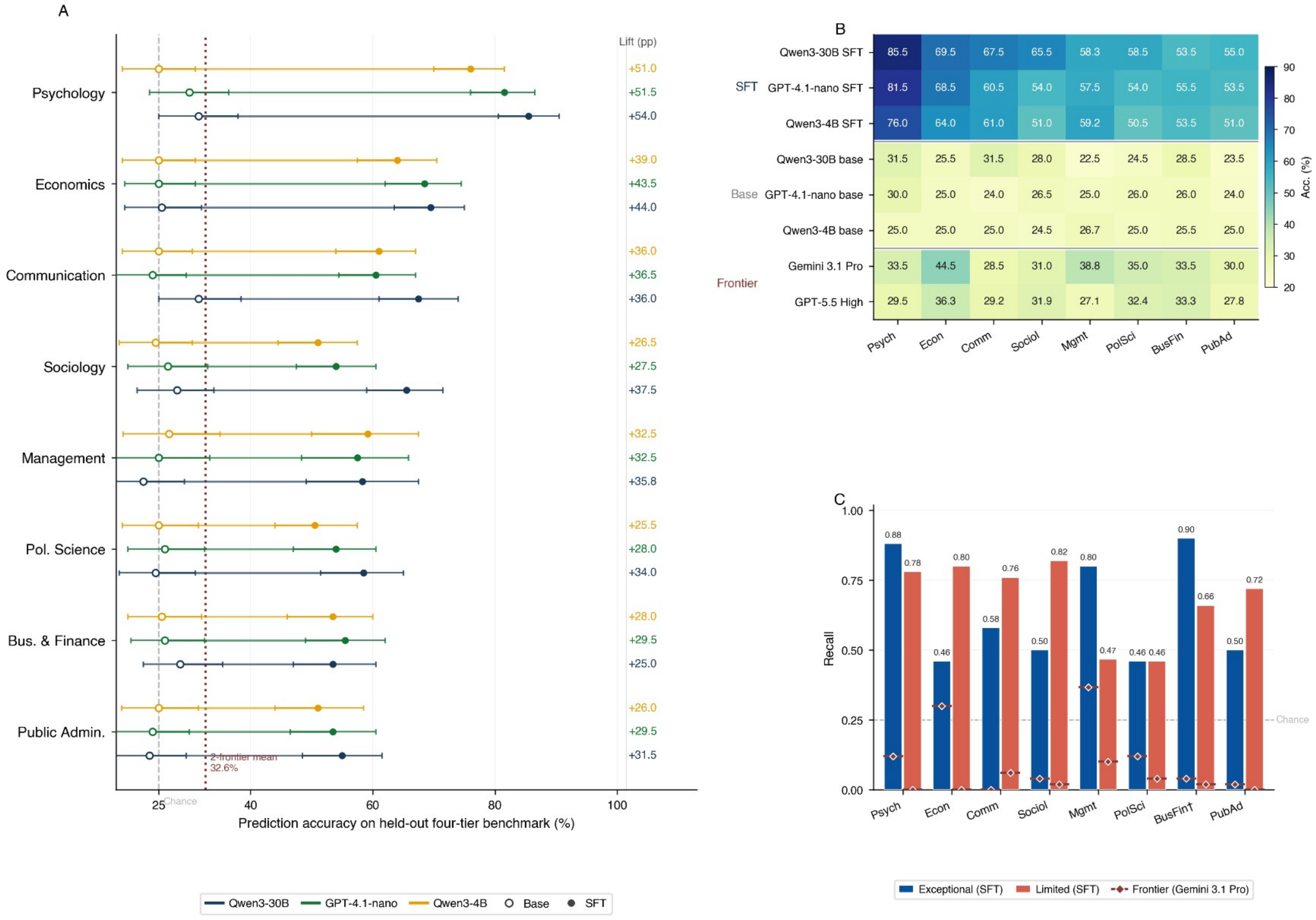

**Figure 2 | Institutional traces are learnable across eight social-science fields.**

a, Forest plot of base → SFT prediction accuracy on a four-tier held-out benchmark for three matched architectures (Qwen3-30B-A3B, GPT-4.1-nano, Qwen3-4B) across the eight fields, ordered top-to-bottom by best-SFT accuracy. Open circles mark the un-fine-tuned base, filled circles the same model after supervised fine-tuning on institutional traces; whiskers are 95% percentile bootstrap intervals on per-article correctness (1,000 draws, seed 42). The SFT − base lift in percentage points is annotated for each architecture–field cell. Vertical dashed line, chance (25%); dotted red line, mean accuracy of the two frontier controls shown in b across the eight fields. b, Eight-by-eight comparison matrix: rows are the three matched SFT models, the same three base models, and two frontier reference systems (Gemini 3.1 Pro and GPT-5.5 High from the May 2026 evaluation); columns are the eight fields ordered as in a. Cell values are prediction accuracy (%); the colour ramp saturates between 20% and 90%. White dividers separate the SFT, base and frontier blocks. c, Per-field tier-extreme recall. For each of the eight fields, dark-blue bars give Exceptional recall and red bars give Limited recall for the best single SFT model in that field (Qwen3-30B-A3B in six fields; Qwen3-4B in management; GPT-4.1-nano in Bus. & Finance, marked †). Dark-red horizontal dashes with diamond markers give the matching Gemini 3.1 Pro frontier reference at the same tier within each field, so the SFT-vs-frontier gap is visible field-by-field. Dashed line, chance (0.25). Sample sizes per field: management $n = 120$; each of the seven non-management fields $n = 200$; tier counts vary by field. Pooled across the eight fields, the best-SFT-per-field combination achieves 0.63 Exceptional recall and 0.69 Limited recall versus 0.11 and 0.03 for Gemini 3.1 Pro.

These findings establish the breadth of the phenomenon; we next calibrate its magnitude in management, the only field with full human and frontier panels (Section 2).

## 2. In management, fine-tuned models outperform expert majority vote and frontier reasoning models

We treated management as the *reference field*, the calibration anchor against which the cross-field accuracies in Section 1 should be interpreted. Because two of the authors hold domain expertise in management, we could recruit a panel of senior gatekeepers, evaluate the largest cohort of frontier reasoning systems, and run mechanism probes that benefit from larger representational capacity. The depth-in-management design is feasibility-driven: the breadth result above demonstrates the phenomenon, and the management probes pin down its magnitude and mechanism.

*Frontier models remained close to chance.* Across the March 2026 cohort of 11 frontier reasoning models (Gemini 3.1 Pro, Claude Opus 4.6, GPT-5.2 High, Gemini 2.5 Pro, Qwen 3.5 Plus, Grok 4.1 Fast, Kimi K2.5, DeepSeek V3.2, GLM-5, Seed 2.0 Pro, and MiniMax M2.5) evaluated under one pre-specified, expert-derived prompt incorporating assessment criteria for originality and utility[27, 28], mean accuracy was 31.1%, capturing only 8.1% of the headroom between the 25% chance baseline and 100% accuracy (Figure 3a). The management frontier result used this pre-specified expert-rubric comparator prompt; the simplified and journal-anchored prompt-sensitivity variants are documented in SI SM2. Across those variants, prompt engineering can move the frontier baseline modestly but does not recover the missing evaluative signal. The best individual frontier performer, Gemini 3.1 Pro, reached 38.8%. A separate May 2026 GPT-5.5 High all-field evaluation did not change this conclusion: GPT-5.5 High reached 27.1% pitch-mean accuracy on the management benchmark and averaged 30.9% field-unweighted across all eight fields. Mean macro F1 for the original management cohort was 0.234, *below* the 0.25 expected from random guessing, because models achieved their above-chance accuracy by over-predicting favorable categories while producing zero recall on tiers they ignored. Six of 11 models never predicted "limited" for any of the 120 articles, defaulting to generically favorable assessments consistent with preference-training-induced leniency[19, 20]. Cochran's Q across the cohort yielded $P = 0.573$: the close-to-chance pattern was uniform and systemic across the frontier cohort, not a property of any single system. Three distinct collapse patterns underlay it (Figure 3b): a strong-ceiling collapse (Claude Opus 4.6 assigned 87% of articles to a single tier, macro F1 = 0.145), a middle-tier-clustering collapse (GPT-5.5 High assigned all management predictions to Strong + Fair), and a top-leniency collapse (Grok 4.1 Fast over-assigned Strong + Exceptional, as did Seed 2.0 Pro). General language capability, even at the frontier, did not confer evaluative discrimination on this benchmark.

*Human experts remained well below the fine-tuned models.* The 48-expert gatekeeper panel achieved an individual-mean accuracy of 36.2% (significantly above chance, but with substantial variance: some experts performed below chance, while one reached 100% on their rated subset). Aggregating expert votes by majority rule produced 41.6% accuracy on the 89 of 120 articles that produced a clear plurality (31 yielded ties and were excluded; expert macro F1 = 0.404). The 174-junior-researcher panel performed comparably: 31.7% individual mean, 40.8% majority vote on 103 articles with clear plurality, and Monte Carlo subsampling at expert-equivalent panel sizes yielded accuracy indistinguishable from senior experts. Notably, expert categorical agreement was near chance (Fleiss' kappa = 0.047) while ordinal agreement was moderate (Krippendorff's alpha = 0.307): experts shared a coarse sense of relative quality but drew categorical boundaries in different places. No expertise marker predicted accuracy (not career stage, not self-reported confidence, not topic familiarity), consistent with journal-review evidence that domain experience and publication record do not reliably identify good evaluators and with grant-review evidence that expertise can shift scoring without eliminating evaluative variability[11, 12, 29].

*Fine-tuned models recovered a substantially larger fraction of available headroom.* All four management fine-tuned models (GPT-4.1, Qwen3-30B-A3B, Qwen3-4B, GPT-4.1-nano) achieved 55.0–59.2% accuracy: GPT-4.1 (55.0%), GPT-4.1-nano (57.5%), Qwen3-30B-A3B (58.3%), Qwen3-4B (59.2%). Each was significantly above expert majority vote (binomial $P < 0.005$), each was significantly above the frontier mean ($P <$ 1e-9), and each lay within an architecture family showing matched-base controls at chance (Figure 3a). The best single fine-tuned model in management was Qwen3-4B at 59.2%, exceeding the best human aggregate (expert majority at 41.6%) by 17.6 percentage points and the frontier mean (31.1%) by 28.1 percentage points. As a sensitivity check, a two-model fine-tuned ensemble (GPT-4.1-nano + Qwen3-30B-A3B) reached 60.8% (95% CI: 51.7–69.2%); ensembling extended rather than founded the result, and ensemble lift is not the cross-field phenomenon (see Section 4). The cost-effective trio (Qwen3-30B-A3B, Qwen3-4B, GPT-4.1-nano) reached benchmark accuracy comparable to GPT-4.1, so the larger GPT-4.1 model was retained only for management mechanism probes and the cross-field-transfer test. Management fine-tuning therefore cost ~$200 (GPT-4.1) plus ~$10 (GPT-4.1-nano) and ~9 A100 GPU-hours (Qwen3-30B-A3B + Qwen3-4B); the seven-field cross-field core added approximately $70 in API spend and ~63 A100 GPU-hours across the three cost-effective architectures.

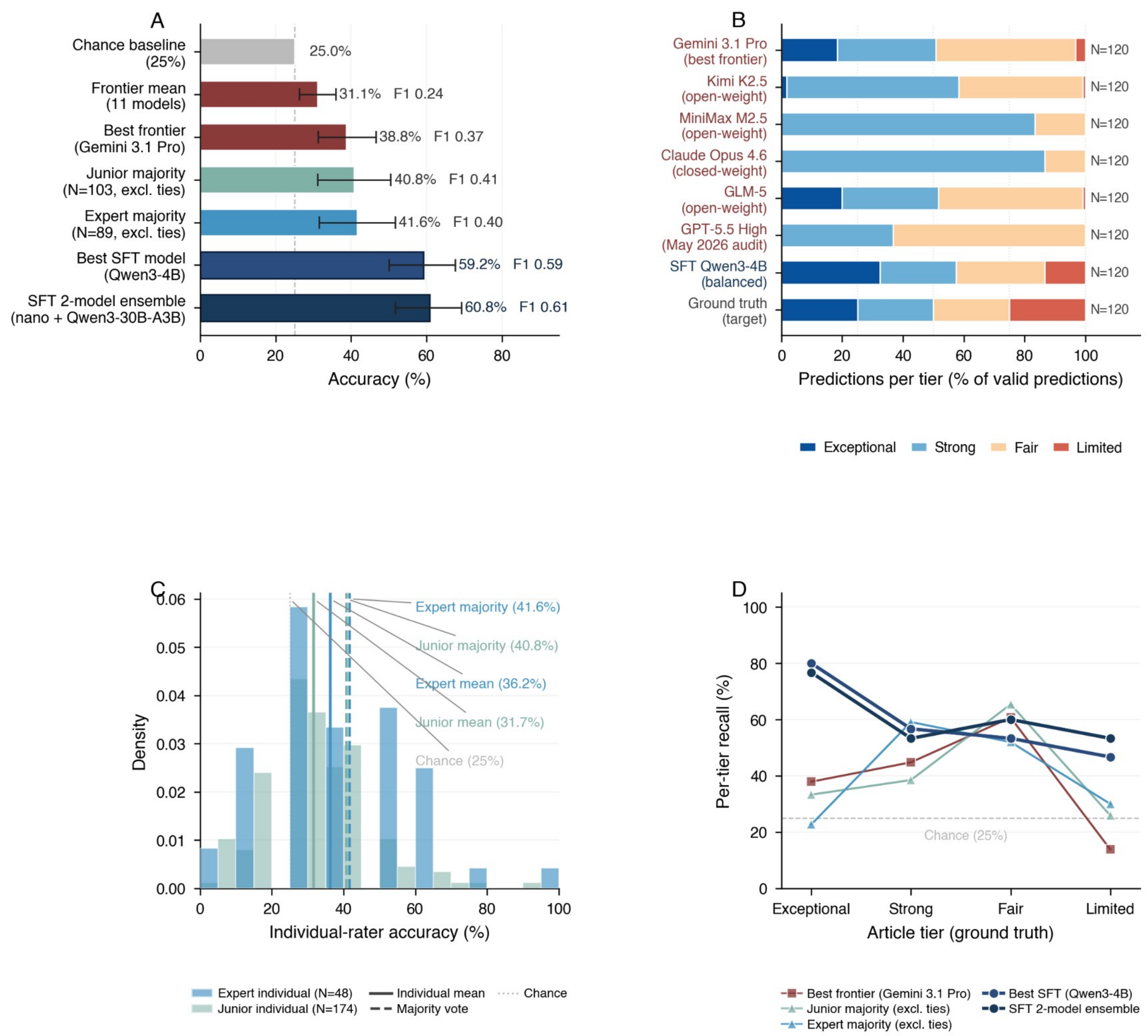

**Figure 3 | The management benchmark calibrates fine-tuned models against humans and frontier models.**
Panels a–d use the 120-article management benchmark balanced across four ground-truth tiers (exceptional, strong, fair, limited; 30 each). a, Prediction-accuracy ladder for chance, the 11-frontier-model mean and best (Gemini 3.1 Pro), the junior- and expert-rater majority votes (excluding ties), the best single SFT model (Qwen3-4B SFT) and the SFT 2-model ensemble (GPT-4.1-nano + Qwen3-30B-A3B). Bars are accuracy; error bars are 95% bootstrap percentile intervals on per-article correctness; macro-F1 is annotated to the right of each row where defined. The two SFT summary rows are outlined. Vertical dashed line, chance (25%). b, Predicted-tier distribution as a percentage of valid predictions for six representative frontier models (Gemini 3.1 Pro, Kimi K2.5, MiniMax M2.5, Claude Opus 4.6, GLM-5, GPT-5.5 High), the best single SFT model (Qwen3-4B SFT) and the ground-truth target (balanced 30/30/30/30 design). The six frontier reps span March 2026 closed-weight/open-weight cohort representatives plus the May 2026 GPT-5.5 High evaluation row, and illustrate three distinct collapse modes: strong-ceiling (MiniMax M2.5 and Claude Opus 4.6 push almost all probability mass into *strong*); middle-clustering (Gemini 3.1 Pro and GPT-5.5 High concentrate predictions in *strong* + *fair*); top-leniency (Kimi K2.5 and GLM-5 over-predict the upper tiers relative to the balanced target). Only the SFT row tracks the target distribution; all eight rows have full coverage ($N$ = 120 of 120). c, Density histograms of individual-rater prediction accuracy for experts (blue, $n$ = 48 raters, unfiltered) and juniors (teal, $n$ = 174 filtered raters). Solid vertical lines mark individual-rater means; dashed vertical lines mark majority-vote accuracies (ties excluded); the dotted grey line marks chance. Leader-line callouts identify each marker. d, Per-tier recall as a function of ground-truth tier (exceptional → limited) for the best frontier model (Gemini 3.1 Pro), the junior majority vote, the expert majority vote, the best single SFT (Qwen3-4B SFT) and the SFT 2-model ensemble. SFT rows are drawn with thicker strokes; circles mark SFT, triangles mark human-majority, squares mark frontier.

## 3. Fine-tuned models learn calibrated evaluative discrimination

The cross-field result establishes that fine-tuning recovers institutional taste; this section establishes that the recovered signal has the internal structure of genuine evaluative judgment rather than surface pattern matching.

*Calibrated confidence generalizes across all eight fields.* A hallmark of expertise is knowing what one knows: a skilled reviewer is more confident when correct than when guessing. The same property is what we want from a model that operates as an evaluative triage layer. Across all 24 fine-tuned model–field combinations, mean confidence on correct predictions exceeded mean confidence on incorrect predictions (Mann–Whitney U, all $P \leq 0.014$; Figure 4a)[30]. The mean confidence gap ranged from +0.067 in management to +0.237 in psychology. Expected calibration error (ECE) for GPT-4.1-nano (the best-calibrated fine-tuned architecture on average, used as the calibration anchor in the main text; per-architecture ECEs in SI Table ST2) averaged 0.092 across the eight fields, with psychology achieving the lowest ECE (0.045). By contrast, the all-field GPT-5.5 chat/log-probability comparator was near chance at full coverage (field-unweighted mean 32.4%, range 28.5–38.0%) and severely overconfident (mean ECE = 0.626; per-field range 0.584–0.660). Human-expert confidence in management was likewise non-discriminative (no significant correct-vs-incorrect gap). Token-probability records were not available for the sampled 11-model frontier-reasoning cohort in our evaluation logs, so the separate GPT-5.5 chat/log-probability track is used only as the frontier confidence comparator in Figure 4b. Across fields, larger confidence gaps generally appeared in the easier, more standardized disciplines, but this cross-field association is descriptive at $n = 8$. Per-field confidence gaps, GPT-4.1-nano ECEs, and Mann–Whitney significance values are tabulated in SI Table ST2.

*Selective prediction concentrates accuracy in high-confidence subsets.* Calibrated confidence enables a practical triage mechanism: route the highest-confidence predictions directly and reserve scarce reviewer attention for the uncertain remainder. When fine-tuned predictions were restricted to the top-10% confidence subset, mean accuracy across the eight fields was 94.6% (range 81.7% in political science to 100.0% in psychology, averaged across three fine-tuned architectures; full table in SI ST16; Figure 4b displays the Qwen3-30B-A3B slice and an eight-field GPT-5.5 chat/log-probability range band). All fields contained at least one high-confidence subset with 100% accuracy, although in some field-model cells this corresponded to only a few percent of cases; in psychology, the GPT-4.1-nano fine-tuned classifier reached 100% accuracy across the top 30% of its confidence distribution. In psychology, the larger two fine-tuned architectures maintained 80%+ accuracy even at full coverage; the smallest architecture (Qwen3-4B) reached 76.0% at full coverage. At the opposite extreme, political science reached 80% accuracy at approximately top-10–15% coverage in two of three architectures. The selective-prediction property is not an artifact of any one field; it generalizes wherever institutional traces are learnable. Where all four management fine-tuned models agreed on a label, accuracy on the *exceptional* tier reached 100% (14 of 14 unanimous-consensus pitches; SI Table ST22).

*Pairwise head-to-head discrimination.* To confirm that fine-tuning develops generalized evaluative ordering rather than four-class classification artifacts, we tested a management-trained GPT-4.1 fine-tune (used here because pairwise comparison benefits from the larger model's representational capacity) on a format absent from training: simultaneous head-to-head comparison of two pitches. The fine-tuned GPT-4.1 model reached 84.3% accuracy (253 of 300 pairs; Figure 5b), exceeding the March 2026 pairwise comparators Gemini 3.1 Pro (77.3%), GPT-5.2 High (78.7%), Grok 4.1 Fast (74.0%), and its own pre-training baseline (76.0%; McNemar $P = 0.0006$). The later GPT-5.5 evaluation was run on the four-class benchmark rather than this pairwise task. The largest margins appeared at the fair-versus-strong and strong-versus-exceptional boundaries, where the institutional quality distinctions are subtlest, the cases where evaluative judgment matters most.

*Reinforcement learning with chain-of-thought underperforms direct fine-tuning.* A final mechanism probe asks whether the same institutional traces, used as a supervision signal for an alternative training paradigm, recover comparable evaluative discrimination. Under this management-only setup, if evaluative judgment is tacit and hard to decompose into reasoning steps, direct learning from examples should outperform reward optimization over articulated reasoning chains. Reinforcement learning (RL) with chain-of-thought reasoning achieved 40.3% on the management benchmark, above the frontier mean (31.1%) but below every individual fine-tuned model (55.0–59.2%; Figure 5a). The RL objective, reward design, and adaptive sampling protocol are reported in SI SM6. A GPT-5.5 evaluation reached the same conclusion without any fine-tuning: across 1,520 benchmark items spanning all eight fields, GPT-5.5 High reasoning did not improve evaluative judgment over GPT-5.5 chat/log-probability classification (item-weighted 31.2% versus 32.3%; paired $t$ test $P = 0.0468$, with high reasoning lower; majority-vote McNemar $P = 0.0116$). Field-level results are visualized in Supplementary Fig. SF11. When the RL model's reasoning chain diverged from its final label, accuracy collapsed; when reasoning and label agreed, performance approached fine-tuned model levels. This dissociation is consistent with prior findings that verbal articulation of holistic judgments degrades their quality[31] and that chain-of-thought gains concentrate in mathematical or symbolic tasks and can reverse on tasks where deliberation itself hurts performance[32, 33]: in this domain, evaluative pattern recognition appears to transfer more reliably from examples than from articulated reasoning chains. The result is also the natural complement of the verifiable-reward paradigm[21]: where outputs can be checked, RL-from-verifiable-reward is a powerful training signal; where they cannot, supervised exposure to institutional outcomes outperformed the specific reasoning-optimized RL setup tested here.

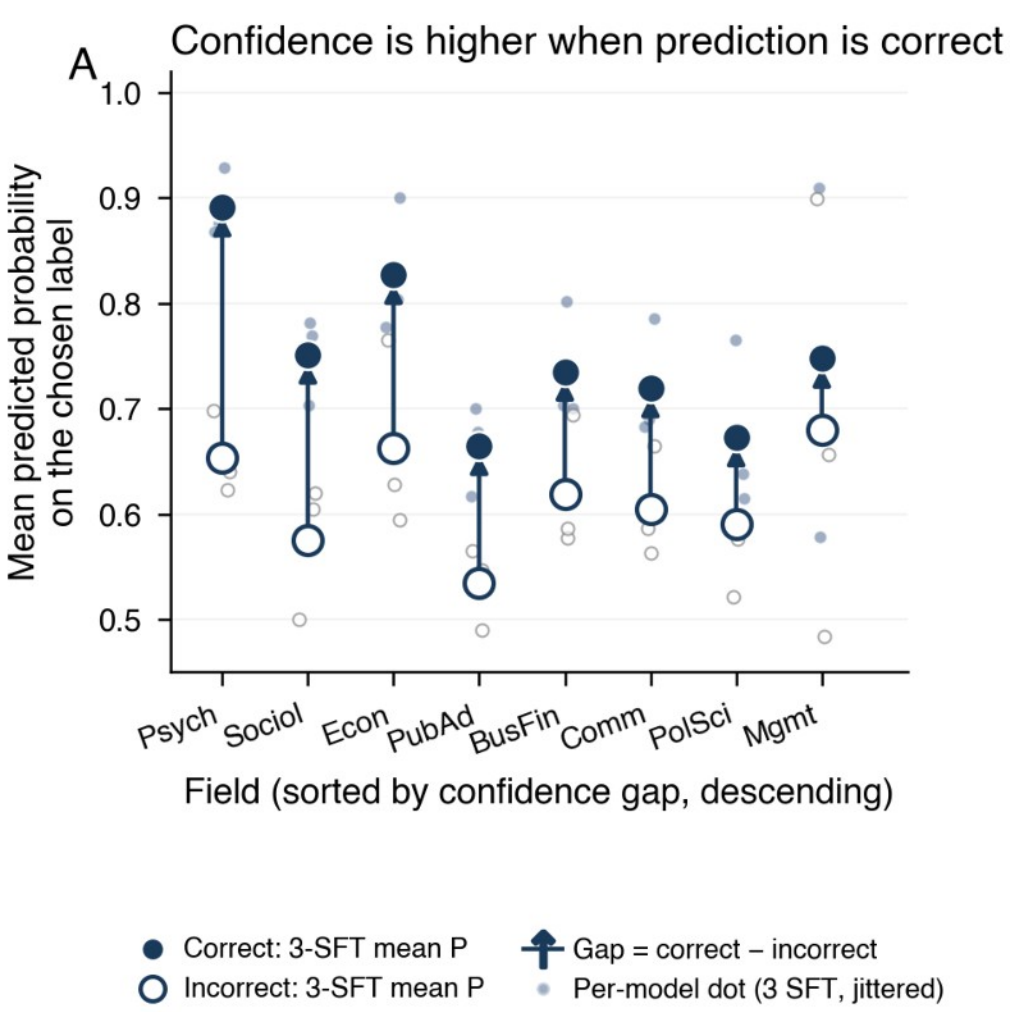

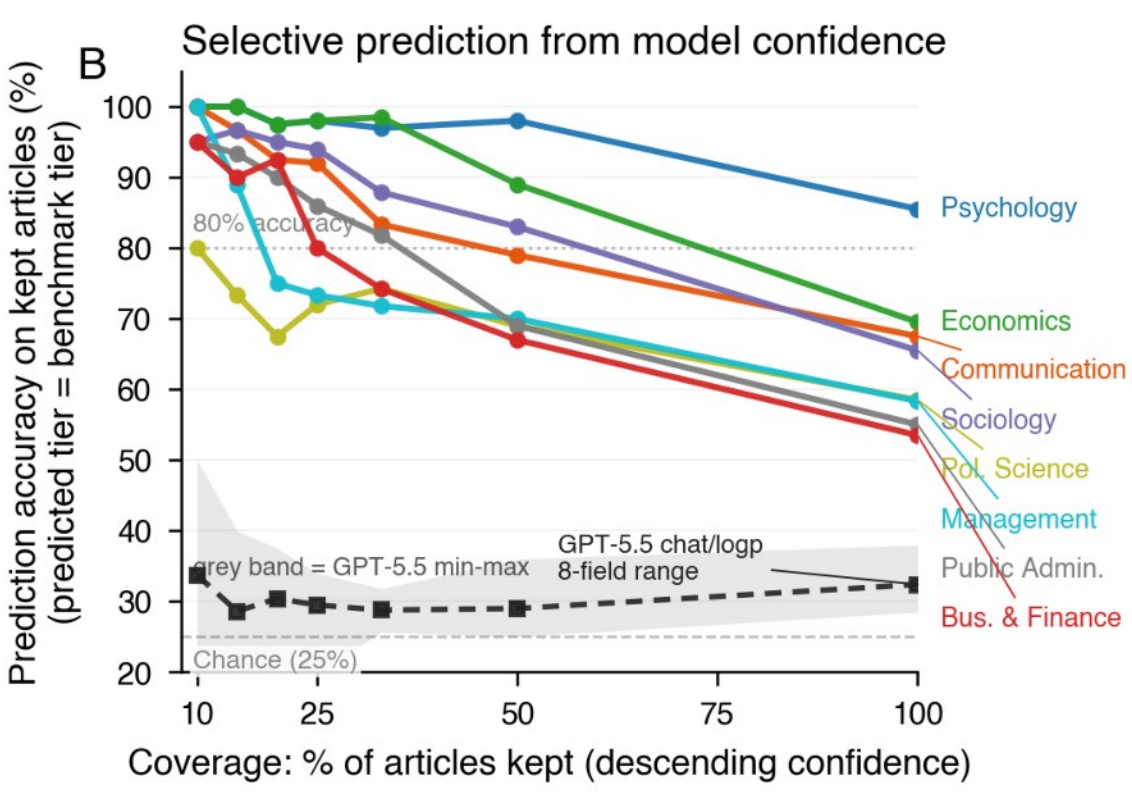

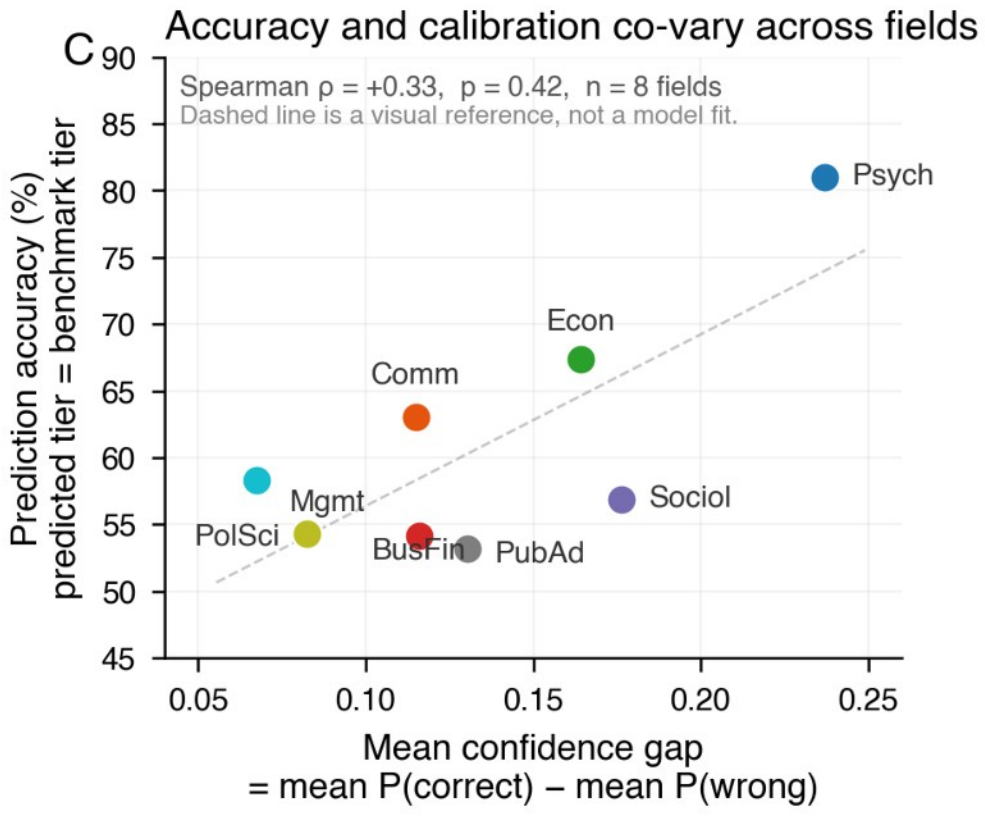

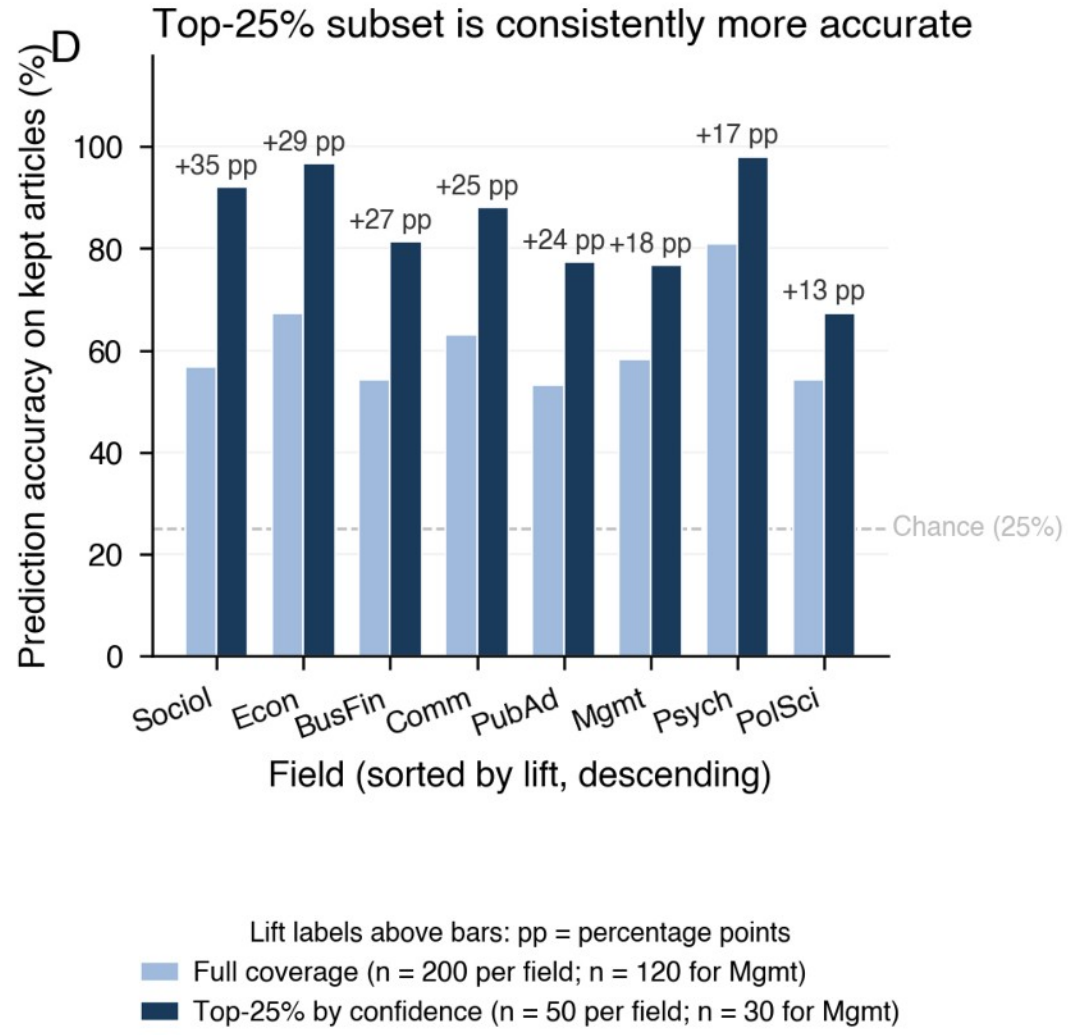

**Figure 4 | Fine-tuned model confidence is calibrated within field, and selective prediction concentrates accuracy in high-confidence subsets.**

a, Mean predicted probability assigned to the chosen label, separated by whether the prediction was correct (filled circle) or incorrect (open circle), for each of the eight social-science fields. Markers are 3-SFT-model means (Qwen3-30B SFT, Qwen3-4B SFT, GPT-4.1-nano SFT); faint jittered dots show per-model values. The blue arrow connects incorrect to correct means; its length is the *confidence gap* = mean $P$(correct) − mean $P$(wrong). Fields are ranked by gap, descending. b, Selective-prediction curves for the representative Qwen3-30B-A3B SFT architecture: prediction accuracy on the most-confident $k$% of articles per field, for $k \in \{10, 15, 20, 25, 33, 50, 100\}$. All eight fields are plotted explicitly with line-end field labels. The dashed black curve is the mean GPT-5.5 chat/log-probability comparator across the same eight field benchmarks, and the grey band is its per-field min–max range at each coverage threshold. GPT-5.5 is plotted here because its chat/log-probability run supplies token probabilities for confidence sorting, whereas the sampled frontier-reasoning cohort produced sampled labels. Reference lines, 80% accuracy (dotted) and chance (25%, dashed). c, Across-field association between the mean confidence gap ($x$-axis) and mean prediction accuracy ($y$-axis); each marker is a field ($n$ = 8). The dashed line is a visual reference, not a model fit, and the panel is descriptive. d, Triage operating point: for each field, prediction accuracy at full coverage (light blue) versus accuracy on the top-25% most-confident articles (dark blue). Full coverage = 200 articles per field for the seven non-management fields, 120 for management; top-25% = 50 articles for the seven, 30 for management. Bar-pair annotations are coverage lift in percentage points; chance (25%) shown for reference. Per-field benchmark sizes throughout: $n$ = 200 articles for each of the seven non-management fields, $n$ = 120 for management. Field abbreviations: Psych, Psychology; Sociol, Sociology; Econ, Economics; PubAd, Public Administration; Comm, Communication; BusFin, Business & Finance; PolSci, Political Science; Mgmt, Management. Prediction accuracy is exact four-class match: the predicted tier equals the benchmark tier.

## 4. Robustness and operating regime

Three boundary-condition probes test whether the institutional-trace signal survives radical changes in input format, time, and aggregation method, and a final summary describes the operating regime in which we propose this approach should be deployed.

*Transfer under input compression.* To test whether fine-tuned models relied on superficial features of the full pitch descriptions, we re-tested management checkpoints on compressed short idea statements that removed much of the theoretical framing, methodological detail, and contribution language used during training (Figure 5c). The larger GPT-4.1 fine-tune retained substantial evaluative signal: 49.2% accuracy on compressed inputs versus 55.0% on full pitches. This compressed-input accuracy still exceeded the best frontier model on the *full* input (Gemini 3.1 Pro, 38.8%) and the expert majority vote (41.6%), demonstrating that the evaluative representation survives radical reduction of surface cues. The smaller GPT-4.1-nano fine-tune, however, collapsed under the same compression, falling from 57.5% to 33.3%, barely above its base control. This asymmetry indicates a specialization–generalizability trade-off: both models acquire the evaluative signal from the same institutional traces, but representational capacity determines how robustly that signal transfers beyond the training input format.

*Temporal stability.* In management, we also trained a matched-architecture model on an older institutional-trace slice (2015–2020, 3,368 pairs), a five-year lag relative to the post-June-2025 benchmark articles. Qwen3-30B-A3B fine-tuned on the older slice reached 46.7% on the 2025+ benchmark (Figure 5d), still exceeding the frontier mean (31.1%) and the expert majority vote (41.6%), versus 58.3% for its recent-slice counterpart. A model trained on decade-old editorial decisions thus outperformed both the most capable frontier systems and the senior human gatekeepers available today. What drifts is the quality bar, not the underlying signal: the older model over-predicted "exceptional" by approximately 50%, reflecting competitive thresholds that have since risen as submission volumes grew and acceptance rates fell. Periodic retraining on current traces will be needed to maintain calibration, but the underlying mechanism (pattern recognition over institutional conventions) is durable.

*Ensembling does not generalize as a default cross-field strategy.* The 60.8% management ensemble cited in Section 2 (a +1.6-percentage-point lift over the best single fine-tuned model in management, 59.2% → 60.8%) raises the natural question of whether ensembling delivers comparable lifts in the seven other fields. It does not. Across the full eight-field cohort, ensemble lift is concentrated in management (+1.6 pp), economics (+3.5 pp), and one additional positive field (communication, +1.5 pp), with five non-management fields neutral or negative (psychology, 0; public administration, −0.5 pp; business and finance, −1.0 pp; political science, −1.5 pp; sociology, −2.0 pp; Supplementary Fig. SF9). Three-model ensembles were uniformly worse than the best single. The mechanistic reading is that ensembling helps when no single fine-tuned model dominates: in management and economics, checkpoint accuracies are within ~1 percentage point of each other and disagree often enough that probability averaging recovers a small additional gain; outside those cases, Qwen3-30B-A3B (or, for business and finance, GPT-4.1-nano) is usually roughly +5 pp ahead of the next best, so averaging in a weaker partner often pulls accuracy down. The cross-field phenomenon documented in Section 1 is therefore robust without relying on ensembling: the universal property is single-model performance after field-specific fine-tuning, and the best-single-fine-tuned-model framing used throughout this paper is the principled cross-field operating choice.

*Operating regime.* The practical operating regime is selective triage, not autonomous replacement. Calibrated confidence supports routing the highest-confidence predictions for fast-tracking or fast-rejection and reserving scarce reviewer attention for genuinely uncertain cases. As a secondary diagnostic, the correct tier often appeared among the two most likely labels, consistent with adjacent-tier ambiguity rather than arbitrary error; we report those values in SI Table ST15. We do not propose fine-tuned models as autonomous evaluators replacing peer review. We propose them as *triage*, calibrated against the institutional record from which the human system itself takes its consistency.

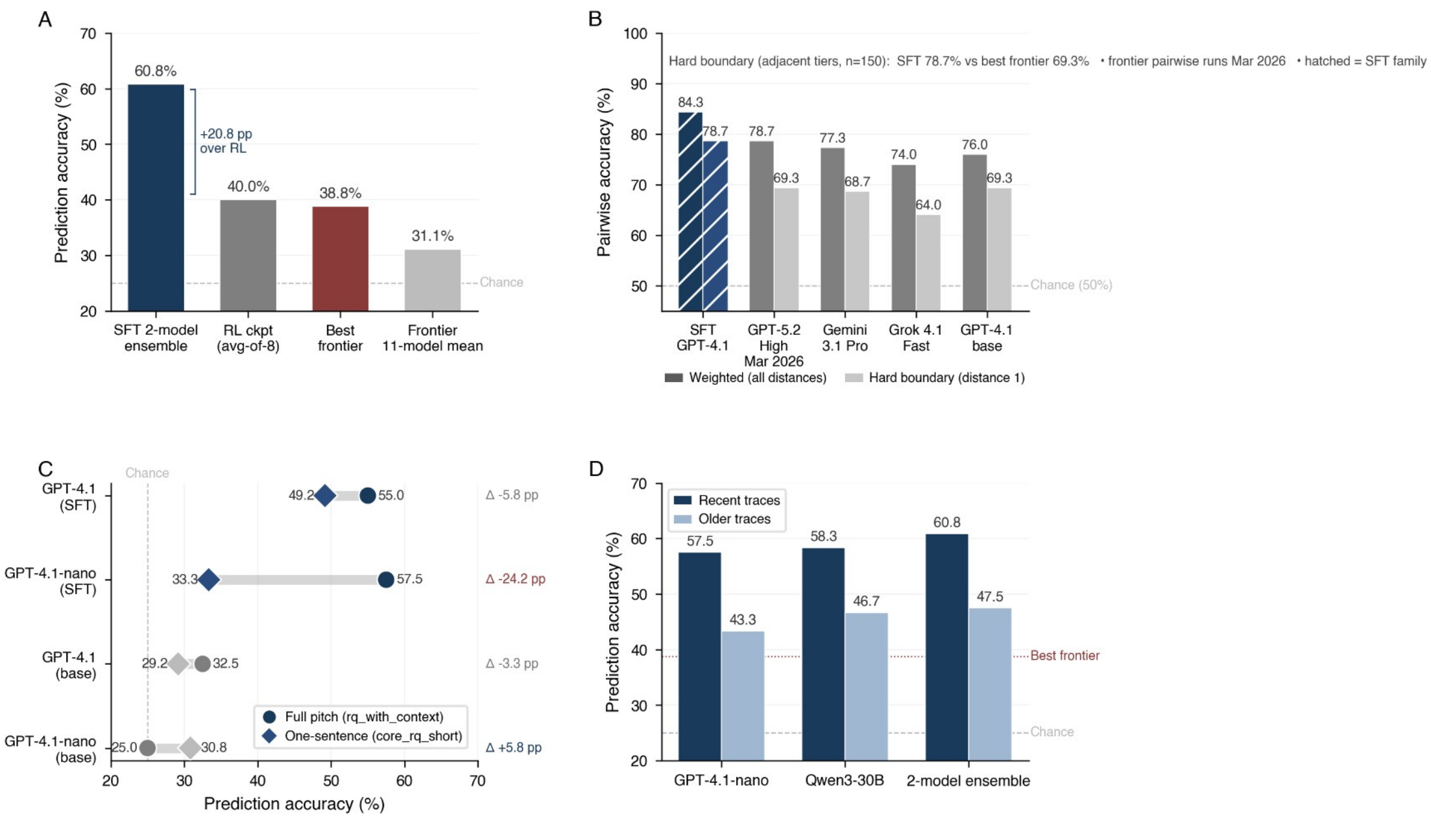

**Figure 5 | Mechanism and robustness probes in management bound the operating regime of institutional-trace fine-tuning.**

Panels a–d use the 120-article management benchmark ($n$ = 120). a, Prediction accuracy comparing the SFT 2-model ensemble (GPT-4.1-nano + Qwen3-30B-A3B), a reinforcement-learning checkpoint trained with chain-of-thought reasoning (mean over eight samples per pitch), the best individual frontier model (Gemini 3.1 Pro), and the 11-model frontier mean. b, Pairwise head-to-head accuracy on a format absent from training (simultaneous comparison of two pitches): bars show overall accuracy on all 300 pairs and accuracy on the hardest 150 same-tier-adjacent (1-tier-difference) pairs for SFT GPT-4.1, GPT-5.2 High, Gemini 3.1 Pro, Grok 4.1 Fast, and the un-fine-tuned GPT-4.1 base; the pairwise frontier comparators are the March 2026 management pairwise runs, and SFT-family bars are hatched. Pairwise task details and accuracy summaries are reported in Supplementary Methods 6. c, Transfer under input compression: prediction accuracy of GPT-4.1 SFT and GPT-4.1-nano SFT on full research pitches (training-format input) versus compressed short idea statements with much of the theoretical framing, methodological detail, and contribution language removed (out-of-format input). d, Temporal stability: prediction accuracy on the post-June-2025 benchmark for Qwen3-30B-A3B SFT and the 2-model SFT ensemble, each trained on a recent versus an older management institutional-trace slice (2015–2020, 3,368 pairs). Panel d includes reference lines for chance and the best frontier model. The cross-field ensemble-sensitivity probe is reported in Supplementary Fig. SF9.

## 5. Field-level variation in learnability is graded but universal in direction

Best single fine-tuned model accuracy follows a stable broad gradient across the eight fields, with psychology (85.5%) and economics (69.5%) at the high end and business and finance (55.5%) and public administration (55.0%) at the low end; across the three fine-tuned architectures, this broad gradient is stable even though exact field rankings vary, indicating that the differences are properties of the fields' institutional traces rather than of any particular model. Cross-field transfer of management-trained checkpoints is bounded by fine-tuned model scale rather than by surface field proximity: the flagship GPT-4.1 transfers most strongly (cross-field mean 37.9%; in-domain → cross-field drop −17.1 pp), while the smallest 4-billion-parameter checkpoint transfers least (32.4%; −26.8 pp drop), with the two intermediate architectures falling in between. Larger fine-tuned models thus recover deeper shared cross-field signal, while smaller architectures specialize on the home field — scale buys cross-domain robustness. Per-field accuracies, SFT-vs-base lifts, transfer correlations, and ensemble-sensitivity values are tabulated in SI Tables ST1, ST13, ST17, and ST14, respectively; ensemble sensitivity is plotted in Supplementary Fig. SF9.

## Discussion

Scientific taste, the capacity to judge which untested ideas are worth pursuing, has been treated as an irreducibly individual property[8, 13]. The data here reframe it as an *institutional* property. Across eight social science disciplines, fine-tuning on publication records produced evaluative discrimination that neither frontier AI systems nor experienced human gatekeepers achieved through reasoning or expertise alone. The signal was never absent; it was simply never the explicit training target. Individual reviewers agree at barely above chance (kappa = 0.047 in our 48-expert management panel; meta-analytic kappa = 0.17 across 48 prior studies[10]), yet the aggregate institutional system deposits consistent quality stratification into the publication record[14]. Fine-tuning extracts what peer-review committees collectively encode but no individual reviewer can articulate: Collins's collective tacit knowledge[9] made computationally accessible.

Cross-field variation is graded but universal in direction. Psychology yields the highest fine-tuned accuracy (85.5%) and public administration the lowest (55.0%), and the broad high-to-low gradient is stable across architectures even though exact rankings vary; exploratory cross-field transfer of management-trained checkpoints is bounded by model scale rather than by surface field proximity. Crucially, frontier and base models cluster near chance in every field, including psychology — Gemini 3.1 Pro reaches 33.5%, GPT-5.5 High reaches 29.5%, and GPT-5.5 chat/log-probability reaches 31.5% in the field where fine-tuning reaches 85.5%. Pretraining alone, even on the journal-prestige hierarchies themselves, does not differentiate easy from hard fields. The easy/hard gradient is only legible to a model that has been given explicit access to institutional outcomes; surface text does not reveal it.

The GPT-5.5 evaluation also tests a stronger version of the "frontier models will catch up" objection. GPT-5.5 chat/log-probability was near chance and poorly calibrated across the eight benchmarks, and in that evaluation GPT-5.5 High reasoning was no better than GPT-5.5 chat. This pattern matters because it separates general model progress from the missing training signal. More capable models can improve fluency, breadth, and tool use, but evaluative judgment over low-verifiability ideas still requires exposure to the institutional outcomes that define the field's tacit standard.

If institutional taste can be recovered cost-effectively from publication records, the practical implication is upstream: AI assistance shifts from generation to evaluation. Frontier systems and automated-research agents can now generate hypotheses, analyses, and complete manuscripts at speed[24, 25, 34]; the harder bottleneck is the reward signal needed to decide which candidate deserves further work. In verifiable domains, that signal can come from tests, proofs, or experiments. In low-verifiability domains, institutional traces offer a way to train the evaluator itself. Several properties of fine-tuned models make this triage role realistic in the present setting: calibrated confidence in every field tested (mean top-10% accuracy 94.6%), pairwise transfer to formats absent from training (84.3% in management), cross-architecture consensus that concentrates reliable predictions into a high-precision subset, adjacent-tier rather than arbitrary residual errors, and management temporal stability sufficient that a model trained on decade-old decisions still beats frontier systems and expert panels. The proposed deployment is straightforward: route high-confidence predictions directly and reserve scarce reviewer attention for the genuinely uncertain remainder, an allocation logic consistent with evidence on when human-AI combinations are useful[35]. On the expert-comparable management subset ($N$ = 89), a hypothetical upper-bound rule that takes the correct label from either the fine-tuned ensemble or the expert majority reaches 77.5% accuracy (Supplementary Table ST23), substantially above either evaluator alone and bounding the upside available to a hybrid triage rule. We do not propose autonomous replacement of peer review; we propose a triage layer calibrated against the institutional record from which the human review system itself takes its consistency.

The study deliberately combines eight-field breadth with management-only depth: senior gatekeeper recruitment was tractable for the authors only in management, so we used that field for the deepest probes while validating the institutional-trace phenomenon broadly with cost-effective models elsewhere. The within-architecture fine-tuned-versus-base comparison, universally positive across all 24 model–field combinations, identifies institutional traces as the operative training signal without requiring a human comparator in every field. The management human benchmark calibrates the magnitude rather than founding the mechanism. Extending human benchmarks to psychology, where fine-tuned accuracy is highest, and public administration, where it is lowest, would further test the boundaries of institutional-taste learning, and is the most natural next experiment.

Several substantive caveats remain. The present benchmarks are confined to the social sciences; whether institutional traces work similarly in STEM is an empirical question, especially because reproducibility, citation dynamics, and venue hierarchies play different roles across STEM fields. Institutional traces are a proxy for field-level evaluation, not a measure of objective merit; models trained on them may inherit documented biases toward incremental over novel work[36, 37] and toward dominant paradigms, and calibrated uncertainty provides an internal reliability signal the human institutional system itself lacks but does not eliminate the underlying selection bias[38]. The field gradient should also be read as a result, not a measured causal mechanism: we have not yet constructed an independent index of publication-practice standardization that would explain why some fields are easier to learn than others. Ensemble analyses are sensitivity checks rather than the operating recipe; the universal cross-field phenomenon is single-model performance after field-specific fine-tuning. Finally, periodic retraining on current traces will be required as standards evolve, since the management older-trace temporal-stability test shows that the *signal* is durable but the *quality bar* drifts.

The closest precedent in this literature is *BrainGPT*[22], a Mistral-7B model continued-pretrained on the neuroscience literature that surpasses neuroscientists at predicting future empirical outcomes. The institutional-trace approach is methodologically and conceptually distinct: rather than continued pretraining on a field's full text, we condition on tier-labeled selection traces; rather than predicting a deferred empirical outcome, we elicit a category judgment about *idea quality* before any outcome exists; and rather than a single discipline, we generalize across eight. The two approaches are complementary expressions of a single observation (that fine-tuning LLMs on field-specific structure beats prompting large general-purpose ones), applied to verifiable and low-verifiability scientific judgment, respectively. Concurrent work by Tong et al.[26] pursues a related "scientific taste" framing using community-feedback reinforcement learning anchored on citation counts; our institutional-trace approach instead conditions on the selection event itself (which venue accepted the work, at what tier), avoiding the temporal lag and venue-prestige confounding that citation-count proxies inherit. Beyond academic publishing, the same logic may apply wherever institutional gatekeeping has operated long enough to leave a structured selection record: venture capital, where expert prediction is notoriously poor[39]; grant allocation, where reviewer agreement approaches zero and expertise can shift scoring behavior[29, 40]; creative industries, where market outcomes regularly defy expert forecasts; law and policy, where retrospective consensus emerges only slowly; and hiring or promotion, where committees repeatedly sort candidates under uncertainty. The broader question of how generative AI augments rather than replaces social-science methodology has been usefully framed by recent commentary[41]; and prior work has shown that prompted LLMs can already approximate human forecasters on the *outcome* side of social-science experiments at correlations near $r \approx 0.85$[42]. The institutional-trace approach addresses the complementary, prior question of which experiments are worth running at all. Autor argued that machine learning would eventually overcome Polanyi's paradox (the gap between what we know and what we can tell) by learning from outcome data instead of explicit instruction[43]. The eight-field result here is consistent with that prediction: frontier models that receive evaluative criteria through prompts remain close to chance; fine-tuned models that learn from institutional outcomes recover the evaluative structure that Hinton et al. termed *dark knowledge*[44], judgment implicit in the system's historical sorting but invisible in any stated criterion. Scientific taste is therefore not only an individual faculty; part of it is deposited in institutional records that learning systems can extract.

# Supplementary Information

**LLMs learn scientific taste from institutional traces across the social sciences**

- SF3: Management-SFT Cross-Field Transfer Matrix (Four Architectures × Seven Non-Management Fields)
- SF4: Per-Field Reliability Diagrams for the Best-Calibrated SFT Architecture (GPT-4.1-nano)
- SF5: Selective-Prediction Coverage Curves (Three SFT Architectures × Eight Fields)
- SF6: Prediction-Distribution Comparison Across Evaluator Classes (Management Benchmark)
- SF7: Frontier-Collapse-Metric Landscape (Eleven Frontier Reasoning Models)
- SF8: Per-Field Confusion Matrices for the Best Single SFT Model in Each of the Eight Fields
- SF9: Cross-Field Ensemble Sensitivity
- SF10: Frontier Cohort Diagnostics (11 Models)
- SF11: GPT-5.5 All-Field Evaluation

## Supplementary Methods

### Supplementary Methods 1 (SM1): Supervised Fine-Tuning Hyperparameters and Training Corpus

This section documents the supervised fine-tuning (SFT) configuration used to train the field-specific evaluators. The primary cross-field SFT set contains three architectures for each of the eight social-science fields: Qwen3-4B-Instruct, Qwen3-30B-A3B-Instruct, and GPT-4.1-nano. Management additionally includes a GPT-4.1 fine-tune because the management field is used for the human-comparator, mechanism-probe, and cross-field-transfer analyses. All SFT evaluators use the same evaluation prompt and four-label output space; only the field-specific training corpus changes by discipline.

**Table SM1a. SFT training configuration by architecture.**

| Parameter | Qwen3-4B-Instruct (2507) | Qwen3-30B-A3B-Instruct (2507) | GPT-4.1-nano | GPT-4.1 (management only) |
|---|---|---|---|---|
| Architecture | Dense transformer (4B) | Mixture-of-experts transformer (30B total, 3B active) | Proprietary transformer | Proprietary transformer |
| Training framework | TRL (Hugging Face) | TRL (Hugging Face) | OpenAI Fine-Tuning API | OpenAI Fine-Tuning API |
| Training location | Local GPU server | Local GPU server | OpenAI cloud | OpenAI cloud |
| Learning rate | 1e-4 | 1e-5 | API-managed | API-managed |
| Scheduler | Cosine | Cosine | API-managed | API-managed |
| Batch size | 32 | 32 | API-managed | API-managed |
| Epochs | 2 | 2 | 3 | 4 |
| Optimizer | AdamW | AdamW | API-managed | API-managed |
| Precision | bf16 mixed precision | bf16 mixed precision | Provider-managed | Provider-managed |
| Hardware | 1 x A100 | 8 x A100 | Provider-managed | Provider-managed |
| Approximate training time | ~1 hour per field | ~8 hours per field | ~1 hour per field | ~2 hours for management |

For the Qwen fine-tunes, the deployment checkpoint is the saved model after the scheduled two-epoch SFT run. The OpenAI fine-tunes used provider-managed optimization settings under the epoch counts shown in Table SM1a.

Each training example paired the pre-specified evaluation prompt (SM2) with the primary research-pitch representation (SM5). The completion target was a single tier label from the four-class journal-quality scale: exceptional, strong, fair, or limited. Input tokens were masked from the loss, and the supervised loss was computed on the label completion. Thus, fine-tuning adapted each base model to map research-pitch content to the common four-tier evaluation space rather than to reproduce the prompt scaffold.

Training and evaluation rows were separated by field. Validation or held-out benchmark rows were excluded from gradient updates. The primary analysis uses one field-specific SFT model per architecture per field; it does not use pooled cross-field SFT training. Per-field training corpus sizes after the held-out split are reported in ST11.

**Table SM1b. GPT-4.1-nano SFT training summary for the non-management fields with complete exported diagnostic rows.**

| Field | Total optimization steps | Final train loss | Final token accuracy |
|---|---|---|---|
| Communication | 243 | 0.171 | 94.1% |
| Political science | 275 | 0.276 | 92.0% |
| Psychology, multidisciplinary | 233 | 0.344 | 91.7% |
| Public administration | 212 | 0.549 | 66.7% |
| Sociology | 197 | 0.238 | 86.7% |

Table SM1b documents the scale of the OpenAI fine-tuning runs for fields with complete exported diagnostic rows. These diagnostics are descriptive and are not used as outcome measures in the main analysis. The public reproducibility package carries the released benchmark, prediction, and summary-statistic records used for the reported validation metrics.

Total fine-tuning cost across the eight-field, three-architecture core (Qwen3-4B, Qwen3-30B-A3B, GPT-4.1-nano) was approximately $80 in API spend plus approximately 72 A100 GPU-hours for local Qwen training. The additional management-only GPT-4.1 fine-tune added approximately $200 in API spend. Per-architecture training and inference cost bands are tabulated in Table SM1c.

**Table SM1c. Training and inference cost bands by evaluator class. Per-model training cost is the cost of fine-tuning a single (architecture, field) checkpoint; for the SFT-Qwen rows it is local GPU compute, for the SFT-GPT rows it is OpenAI fine-tuning API spend. Inference cost is per 100 pitches under the evaluation regime used here (single-pass log-probability classification for SFT and chat models; eight-sample chain-of-thought aggregation for frontier reasoning models). Frontier and chat models are accessed via API and incur no fine-tuning cost.**

| Model class | Training cost/model | Inference cost (per 100 pitches) | Notes |
|---|---|---|---|
| Frontier (thinking) | $0 (API access) | >$10 | 8 samples per pitch; chain-of-thought generation |
| Chat (logp) | $0 (API access) | $0.01–$0.10 | Single-pass log-probability classification |
| SFT: Qwen3-4B | ~1 A100 GPU-hour | $0.001 | Local fine-tune; log-probability classification |
| SFT: Qwen3-30B-A3B | ~8 A100 GPU-hours | $0.01 | Local fine-tune; log-probability classification |
| SFT: GPT-4.1-nano | ~$10 (API) | $0.01 | OpenAI Fine-Tuning API; log-probability classification |
| SFT: GPT-4.1 | ~$200 (API) | $0.10 | OpenAI Fine-Tuning API; management only |
| RL mechanism probes | Multi-day 8 x A100 runs | Higher than log-probability pipelines | Chain-of-thought generation and label extraction; management-only mechanism probe |

Source: cost bands are estimated from provider invoices, local GPU-hour logs, and the evaluation regimes documented in SM2. RL checkpoints are included only as management mechanism probes and are not part of the eight-field primary evaluator set; the RL objective, reward, and training configuration are specified in SM6.

## Supplementary Methods 2 (SM2): Evaluation Prompts and Inference Settings

### Pre-specified evaluation prompt (used for SFT training, SFT evaluation, base controls, and frontier controls)

SFT training, SFT evaluation, and architecture-matched base-control evaluation across the eight fields used the same general social-science evaluation prompt. The prompt is field-agnostic by design; institutional-trace alignment is delivered by SFT on field-specific tier-labeled pitches (SM1), not by prompt rewriting. For the GPT-5.5 frontier evaluation, management used the expert-derived management rubric prompt described below, while the seven other fields used this general prompt. This mirrors the study design: management is the deep benchmark with expert-derived criteria, and the other fields test breadth under a shared field-agnostic protocol.

```
You are an expert in social science research. Read the research idea below and estimate its
likely publication potential based on its scholarly contribution.

- Exceptional: Field-defining work with strong theoretical or empirical contribution. It
would influence how researchers across the social sciences think about an important problem.
- Strong: Clear and meaningful contribution. It advances theory, evidence, or method in a
non-trivial way and would be well regarded by scholars in the field.
- Fair: Competent but incremental. It extends existing knowledge in a predictable way, with
limited novelty, scope, or broader significance.
- Limited: Weak contribution. The question is narrow, obvious, poorly motivated, or
methodologically insufficient to support a meaningful scholarly advance.

# IMPORTANT
- Do not use search capabilities to look up information about this idea

# OUTPUT FORMAT

Respond with EXACTLY ONE of these four notations:

- Exceptional
- Strong
- Fair
- Limited

Output only the tier notation in your final answer.
```

### Management-specific frontier control prompt

Management frontier evaluations used a pre-specified expert-rubric comparator prompt that incorporates explicit criteria for novelty and usefulness. Two additional pre-specified variants were retained as prompt-sensitivity probes: a simplified rubric and a journal-anchored rubric.

```
# ROLE
You are an expert evaluator of management research ideas. Your task is to evaluate from a
senior scholar's perspective: be direct and critical, give clear judgments based on novelty
and usefulness to classify research ideas into appropriate publication potential tiers.

# TASK
Read a paragraph describing a management research idea and classify it into one of four
publication potential tiers (Exceptional / Strong / Fair / Limited).

# EVALUATION CRITERIA
- Novelty: whether the research idea challenges existing assumptions, reveals something
genuinely surprising, or provides cognitive disruption that fundamentally changes
understanding of relationships or phenomena.
- Usefulness: whether the research idea addresses problems that matter, with broad
implications for multiple stakeholders, resolving long-standing theoretical debates or
providing insights that meaningfully improve organizational practices.

# CLASSIFICATION TIERS
- Exceptional: strong novelty + strong usefulness; field-reshaping potential; most
prestigious journals.
- Strong: clear strength in one dimension with the other reasonably developed; meaningful
contributions; near-top-tier journals.
- Fair: incremental contributions with modest novelty or usefulness; mid-level journals.
- Limited: lacks both novelty and usefulness; lower-tier journals.

# IMPORTANT
- Do not use search capabilities to look up information about this idea.

# OUTPUT FORMAT
```

```
Respond with EXACTLY ONE of: Exceptional, Strong, Fair, Limited. Output only the tier
notation.
```

For frontier evaluation in management, this prompt was used at zero-shot with eight independent samples per pitch; the per-pitch prediction was the modal sample (ties broken alphabetically). The same expert-rubric prompt was used for the management frontier comparison shown throughout the main text.

**Three prompt variants and sensitivity rationale**

The expert-rubric prompt above (Prompt 1) was the pre-specified management comparator within a three-prompt sensitivity set that varied in structure and anchoring strategy. Prompt 1 anchors the four-tier judgment to expert-derived novelty and usefulness criteria with behavioural anchors per tier. Prompt 2 (simplified rubric) compresses the same four-tier definitions into single-sentence general-language descriptions, removing the novelty/usefulness scaffolding. Prompt 3 (journal-anchored rubric) replaces the abstract quality dimensions with explicit journal-list anchors (UTD24, FT50, ABS 4*, ABS 2–3) so the model is asked to map a research idea to the kind of venue that would publish it. All three share the same persona framing and constrain output to a single tier label.

Prompt 1 is the primary management frontier comparator because it operationalizes the expert novelty/usefulness rubric without naming journals. Prompt 2 tests whether compressing the same tier definitions into a shorter general-language rubric changes the comparison. Prompt 3 tests a journal-anchor formulation and is interpreted only as a sensitivity probe, because explicit venue lists create a leakage path: under chain-of-thought sampling, frontier models routinely enumerated journal lists in their reasoning traces and then matched the pitch to a venue family by stylistic and topical cues rather than by evaluating the research idea itself. This venue-name leakage path is incompatible with the institutional-trace evaluation we report, which forbids venue identification by construction (SM5).

| Prompt variant | Design choice | Use in analysis |
|---|---|---|
| Prompt 1: expert-rubric | Novelty/usefulness criteria with expert-derived behavioral anchors | Primary expert-rubric comparator used in the main text |
| Prompt 2: simplified rubric | Same four tiers, compressed into general-language definitions | Prompt-simplification sensitivity probe |
| Prompt 3: journal-anchored rubric | Explicit journal-list anchors (UTD24, FT50, ABS) | Venue-anchor leakage sensitivity probe; excluded from the primary venue-blinded comparator |

Verbatim text for Prompts 2 and 3 is reproduced below.

```
# Prompt 2 (simplified rubric)

You are an expert in management research. Read the research idea below and estimate the
likely publication tier based on its scholarly contribution.

- Exceptional: Field-defining work. Would be recognized across disciplines as a major
advance. Likely to be widely cited and reshape how researchers think about the topic.
- Strong: Meaningful contribution within the field. Clearly advances theory or method in a
non-trivial way. Would be well-regarded by domain experts.
- Fair: Solid but incremental. Competent execution with limited novelty. Recognized mainly
by specialists in the same narrow area.
- Limited: Weak contribution. Findings are obvious, scope is too narrow, or methodological
issues undermine the work.

# IMPORTANT
- Do not use search capabilities to look up information about this idea

# OUTPUT FORMAT

Respond with EXACTLY ONE of these four notations:

- Exceptional
- Strong
- Fair
```

```
- Limited

Output only the tier notation in your final answer.
# Prompt 3 (journal-anchored rubric)

You are an expert in management research with deep knowledge of academic publishing
standards across top-tier journals.

# TASK
Read a paragraph describing a management research idea and classify it into one of four
journal tiers based on its likely publication venue. Your classification should reflect
where work of this quality and contribution level would most likely be published.

- Exceptional: UTD24 journals or highly regarded FT50 journals with field-defining standing
in their domain - paradigm-shifting work, highest selectivity, field-redefining impact
- Strong: FT50 journals (non-UTD24) or ABS 4* journals - substantial contribution, A-level
quality, high methodological rigor
- Fair: ABS 4 journals (non-FT50) - solid contribution with clear theoretical grounding,
competent execution but limited novelty
- Limited: ABS 2-3 journals - incremental findings, narrower scope, or moderate
methodological rigor

# IMPORTANT
- Do not use search capabilities to look up information about this idea

# OUTPUT FORMAT

Respond with EXACTLY ONE of these four notations:

- Exceptional
- Strong
- Fair
- Limited

Output only the tier notation in your final answer.
```

### Inference settings

Operationally, the scoring layer uses two families rather than a separate "rank projection" procedure. Human ratings are collected on the survey labels and then mapped once into the unified four-class label space (SM3, ST18). Model outputs are scored directly in that same four-class space: SFT/base/chat evaluators use restricted-vocabulary first-token probabilities, while sampled frontier-reasoning evaluators produce discrete tier labels that are parsed and aggregated by modal vote.

For SFT and base controls in the management analyses, inference used probability-extracting decoding: the four-token tier vocabulary (exceptional, strong, fair, limited) was scored at the first decoded position via top-token log-probabilities, normalized by softmax over the four-class restricted vocabulary, and the argmax was returned as the prediction. Tied probabilities were broken using the fixed alphabetical order over the unified label space (exceptional < fair < limited < strong). Secondary cross-field ensemble and transfer sensitivity analyses retain their originally specified deterministic tie-breaking order and are interpreted descriptively rather than as new primary endpoints. For management frontier controls, the expert-rubric prompt was sampled eight times per pitch and aggregated by modal vote as described above. For frontier-class controls in the seven non-management fields, the same general prompt was used with the documented provider-side prediction records; outputs were parsed by string match after stripping whitespace, punctuation, and markdown symbols. Unresolved outputs (overall <1% across evaluator classes) were coded as incorrect.

Prompt-sensitivity analyses compared expert-rubric, simplified, and journal-anchored variants. The expert-rubric variant supported the management frontier comparison in main-text Section 2; the simplified and journal-anchored variants did not close the frontier-to-fine-tuned performance gap.

The GPT-5.5 evaluation used two inference modes: high-reasoning sampled inference and deterministic chat/log-probability classification. The chat/log-probability track covered all 1,520 benchmark items (management $n$ = 120; seven other fields $n$ = 200 each). The high-reasoning track used eight runs per item, producing 12,160 valid final predictions. Full prompts and inference settings are provided in the reproducibility package.

### Zero-shot rationale

All evaluations are zero-shot. Few-shot exemplars would carry noise from confounding factors (execution quality, writing craft, reviewer fit) that are not part of the idea-quality signal under test. Within each comparison set, evaluator classes share the relevant prompt and no-search instruction: cross-field breadth uses the general social-science prompt, whereas management frontier and deep-probe analyses use the expert-rubric prompt documented above. The "no search" instruction is shared across all conditions to prevent test-time leakage from web-indexed venue information.

The management-only RL reasoning probe uses the same benchmark and output label space, but has a separate training objective and reward design because it optimizes chain-of-thought policy behavior rather than first-token classification. Those RL details are reported with the management mechanism probes in SM6.

## Supplementary Methods 3 (SM3): Human Evaluation Protocol

This section provides procedural detail that supplements the human-evaluation summary in Methods. Human benchmarks were collected only in management.

### Institutional review

The study was approved by the Tsinghua University institutional review board (Project No. THU-04-2026-0034). Junior scholars were compensated with 100 RMB and/or access to a research tool developed by the research team. Analyses are reported at aggregate level; direct participant identifiers are not included.

### Recruitment

The expert panel (N = 48) comprised current journal editors and editorial board members of leading management journals, recruited through direct one-to-one professional contact. Given this recruitment approach, no quality filter was applied to the expert panel. The junior panel (N = 174 after filtering) comprised doctoral students, postdoctoral researchers, and early-career faculty, recruited through professional networks including indirect ties. Junior raters who spent fewer than one minute on average per pitch were excluded as perfunctory; this filter showed a marginally significant effect on accuracy (25.3% vs 31.7% individual mean accuracy, $P$ = 0.066).

### Survey instrument

For each benchmark pitch, raters were shown the research-question pitch alongside the evaluation criteria and responded to four items:

1. **Prior exposure**: "Had you encountered this research idea or its source paper before?" (Yes / No).
2. **Quality rating**: "Based on the evaluation criteria, how would you rate the quality of this research idea?" Response options: Top / Top- / Good / Fair. This is the human-facing rating scale, not the AI output vocabulary: human-facing *Fair* is the lowest tier and maps deterministically to unified *limited* in all analyses, while AI-output *fair* remains the third tier in the unified label space (ST18).
3. **Confidence**: 5-point Likert (1 = "Not at all confident" to 5 = "Extremely confident").
4. **Domain familiarity**: 5-point Likert (1 = "Not at all familiar" to 5 = "Extremely familiar").

### Panel descriptive summary

| Metric | Experts | Juniors |
|---|---|---|
| Number of raters | 48 | 174 |
| Total ratings | 384 | 2,530 |
| Mean ratings per rater | 8.0 | 14.5 |
| Mean ratings per pitch | 3.2 | 21.1 |
| Median completion time (s) | 923 | 2,534 |
| Mean confidence (1–5) | 3.50 | 3.46 |
| Mean familiarity (1–5) | 3.15 | 2.81 |

Expert career-stage distribution (N = 48): 5 assistant professor; 17 associate professor; 12 full professor; 12 endowed chair; 2 unreported. Junior background data was matched for 156 of 175 raters (89.1%); 19 could not be matched due to naming inconsistencies between recruitment and survey records.

### Human confidence and familiarity diagnostics

Human confidence was analyzed as a diagnostic, not as a filtering rule. In experts, confidence did not reliably discriminate correct from incorrect ratings (Spearman $r$ = 0.056, $P$ = 0.273). In juniors, the confidence–accuracy association was statistically detectable because of the larger rating count but small in magnitude (Spearman $r$ = 0.095, $P$ = 1.74 × 10^-6). Expert familiarity likewise did not predict correctness (Spearman $r$ = 0.013, $P$ = 0.795). We therefore report human confidence descriptively; the operational selective-prediction analysis uses model probabilities (SM4, SM7, ST16), not human self-confidence.

**Table SM3a. Human confidence and familiarity diagnostics.**

| Metric | Experts | Juniors |
|---|---|---|
| Ratings analyzed | 384 | 2,530 |
| Mean confidence (1–5) | 3.50 | 3.46 |
| Mean confidence if correct | 3.58 | 3.58 |
| Mean confidence if incorrect | 3.46 | 3.40 |
| Confidence gap | +0.114 | +0.176 |
| Spearman $r$ (confidence vs correctness) | +0.056 | +0.095 |
| $P$ value | 0.273 | 1.74 × 10^-6 |
| Mean familiarity (1–5) | 3.15 | 2.81 |

### Aggregation rules

Majority-vote results exclude tied pitches and report the effective non-tied sample size. Ties are not random-broken because doing so would inject sampling noise into the primary evaluator metric. Human-majority labels are first mapped from the survey scale into the unified label space (Top / Top- / Good / human-facing Fair → exceptional / strong / fair / limited). The expert majority vote on the unfiltered panel (primary expert metric) reaches 41.6% accuracy on the 89 non-tied pitches; the filtered panel reaches 39.7% on 68 non-tied pitches; the junior majority vote (filtered) reaches 40.8% on 103 non-tied pitches. Monte Carlo matched-N analyses (5,000 draws) compare junior majority voting at expert-equivalent panel sizes; the matched-N junior majority-vote mean is 36.1%, indicating that the junior advantage in raw majority vote disappears once panel size is controlled. Inter-rater reliability for both panels is low (expert Fleiss' kappa = 0.047; junior 0.032), reflecting the dispersed nature of the human signal even within carefully recruited expert panels.

## Supplementary Methods 4 (SM4): Statistical Analyses

### Primary endpoint

The primary endpoint is four-class exact-match accuracy on each field's held-out validation set. Per-field exact binomial confidence intervals (Clopper–Pearson) are reported alongside point estimates wherever a single field is the unit of analysis.

**Bootstrap CIs**

For per-(model, field) accuracy and macro-F1, 95% bootstrap confidence intervals are computed via 10,000 resamples drawn with replacement from the prediction list. For aggregate quantities across fields (mean accuracy across multiple fields, mean SFT-vs-base lift, mean confidence gap), nonparametric bootstrap confidence intervals are reported with 5,000 resamples drawn with replacement over fields.

**Paired comparisons within field**

Within-field paired comparisons (SFT vs base on the same items, ensemble vs best single, SFT vs frontier control, SFT ensemble vs human majority vote on shared non-tied subsets) use exact McNemar tests. McNemar is valid because predictions in each comparison share the same items. We report raw two-sided p-values; the comparator hypothesis-set per field is small (typically 3–6 paired contrasts), so we apply a Holm–Bonferroni correction within the contrast family for each field when a contrast family is reported as a panel.

**Across-field tests**

For tests pooling across fields (mean SFT lift across the seven non-management fields), exact binomial tests against chance are used only for single (model, field) cells with item-level predictions. Cross-field means and lifts are summarized descriptively with field-level bootstrap intervals rather than treated as independent item-level trials. Where field-level Spearman correlations are reported (accuracy vs ECE across eight fields), we report the rank correlation, the exact two-sided p-value, and the underlying eight-pair table.

**Above-frontier-mean tests**

Tests of SFT-above-frontier-mean use item-level exact binomial or paired tests when a single field and shared prediction list are compared. Cross-field statements that compare SFT means with frontier-control means are descriptive field-level contrasts, with uncertainty summarized by bootstrap intervals over fields. Cross-field heterogeneity in frontier accuracy is summarized with Cochran's Q ($P = 0.573$ in management for the 11-frontier panel). For the cross-field frontier-control comparison, we use a representative two-control panel (Gemini 3.1 Pro and GPT-5.5 High from the May 2026 evaluation); GPT-5.2 is retained only as a historical management-cohort comparator where those specific records exist.

*Headroom captured.* Headroom captured is defined as (accuracy − 0.25) / (1.00 − 0.25), the fraction of the chance-to-perfect interval recovered by a given evaluator on the four-tier balanced benchmark. We report headroom alongside raw accuracy in the management head-to-head comparison so that closely spaced raw accuracies (e.g. expert majority 41.6% versus frontier mean 31.1%) can be compared on a common above-chance scale.

**Inter-rater reliability**

Categorical inter-rater reliability uses Fleiss' kappa with 95% bootstrap CI; ordinal inter-rater reliability uses Krippendorff's alpha with the ordinal distance metric. Pairwise Cohen's kappa is reported for selected AI–AI and AI–human comparisons as descriptive overlap on shared items; these kappas are interpreted as agreement diagnostics, not as evidence of a single shared rater signal.

**Calibration**

Confidence is defined as exp(max log-probability) over the four-class restricted vocabulary. Per-(model, field) calibration is summarized by (i) the mean confidence gap between correct and incorrect predictions, (ii) the per-(model, field) Mann–Whitney U statistic and exact two-sided p-value testing the null that confidence is independent of correctness, and (iii) expected calibration error (ECE) computed in 10 equal-width bins over the [0.25, 1.00] confidence range. Brier score is computed as a secondary calibration metric but not reported in the main text. Multiple-testing corrections are noted where applied.

**Selective prediction**

For each (model, field) cell, predictions are sorted in descending order of confidence and accuracy is computed at coverage thresholds {10%, 15%, 20%, 25%, 33%, 50%, 100%}. For fractional thresholds, selected-set size is floored to the nearest whole item (e.g., 39 of 120 and 66 of 200 at 33% coverage). We report the accuracy-at-coverage series directly; threshold summary columns are omitted from ST16 because the coverage columns provide the same information more transparently.

## Supplementary Methods 5 (SM5): Field Selection, Journal-to-Tier Mapping, and Research-Idea Extraction

**Benchmark construction is human-driven and distinct from the AI evaluation**

The benchmark used throughout this study consists of held-out research-idea pitches, each paired with a four-class institutional quality tier (exceptional, strong, fair, limited). The tier label of each pitch is determined by the publication venue of the source article from which the pitch was extracted. Assigning tiers to journals — that is, building the benchmark — is a human expert-driven process that takes place once, in advance, and is fully separate from the AI evaluation under test. The AI models being evaluated never see the journal identity of any pitch: the evaluation prompt (SM2) is restricted to the research-idea text, and any venue-revealing content is removed during pitch extraction (see *Research-idea extraction pipeline* below). The benchmark therefore measures whether an AI evaluator can recover, from research-idea text alone, the tier that domain experts and the broader scientific community assign to the source article's outlet.

Two complementary procedures were used to build the eight field-specific tier mappings, reflecting differences in the institutional structures of the source fields:

- **Management (one field).** All journals in the management source universe were tier-assigned by direct expert determination. The management subject-matter experts on the author team made the assignments, consulting additional field experts where useful to confirm boundary cases. No external numerical proxy was used; the management hierarchy is sufficiently codified in tenure and editorial-board norms that consensus could be reached directly. The resulting 19-journal source universe is reported in ST3.
- **The seven other fields (economics, business and finance, communication, political science, psychology multidisciplinary, public administration, sociology).** A two-step procedure was used. (i) Field experts in each domain nominated candidate journals for each of the four tiers, drawing on their knowledge of editorial standing, tenure-credit norms, and submission/acceptance patterns. (ii) Each nomination was cross-checked against an external benchmark of publicly available, citable evidence on journal quality, comprising the field-appropriate subset of: (a) citation-impact metrics from the 2024 Journal Citation Reports (Journal Impact Factor and Journal Citation Indicator); (b) recognized journal-quality lists used by tenure committees and field associations — including the Academic Journal Guide (Chartered Association of Business Schools), the Australian Business Deans Council Journal Quality List, VHB-JOURQUAL (Verband der Hochschullehrer für Betriebswirtschaft), the Financial Times 50, and the UT Dallas Top-24 Business Journals — applied to economics and to business and finance; (c) field-association flagship designations from the American Psychological Association (psychology), the American Political Science Association (political science), the American Sociological Association (sociology), the International Communication Association (communication), and the Public Management Research Association (public administration); and (d) community-consensus surveys, including the Garand survey for political science, the Scatterplot rankings for sociology, and the TRIP survey for international relations. (iii) Inconsistencies between expert nominations and external-benchmark evidence were resolved through additional expert consultation and reconciliation. Remaining discordant cases were treated as adjacent-tier boundary questions and resolved by the author team's domain experts. Each field's resulting mapping is reported as a per-field tier table (ST4–ST10).

In both procedures the output is a deterministic mapping from journal to tier; each pitch in the held-out benchmark inherits the tier of its source journal. The author team's domain experts retained final decision authority on all tier boundaries.

**Field selection and Web of Science scoping**

The eight social-science fields covered are management, economics, business and finance, communication, political science, psychology (multidisciplinary), public administration, and sociology. The eight fields were chosen to span the breadth of social-science scholarship while keeping each field's journal hierarchy independently defensible from public evidence on journal quality.

Field scope follows Web of Science discipline categories, which provided the outer boundary for journal eligibility. The source-journal universe for each field was first restricted to journals indexed under that field's Web of Science category (or, for fields with multiple legitimate categories such as business and finance, the union of those categories). Within this scoped universe, tier assignment then proceeded under the field-specific procedure described above. Cross-listed journals whose primary disciplinary identity lay outside the focal field were retained in the source universes and flagged for sensitivity analysis: examples include accounting outlets cross-listed under business and finance and public administration, and tourism/hospitality outlets cross-listed under sociology.

**Per-field source universes**

Management uses the 19-journal source universe (ST3) treated as institutional traces encoding decades of editorial gatekeeping, citation impact, and community consensus. The held-out 120-article benchmark draws from this universe and is balanced to 30 pitches per tier across exceptional, strong, fair, and limited.

For the seven other fields, each per-field tier table (ST4–ST10) is the output of the two-step nomination-and-cross-validation procedure above. Tier semantics are common across all eight fields: *exceptional* denotes discipline-wide elite consensus and near-universal tenure weight at top departments; *strong* denotes top-tier outlets just below the absolute elite; *fair* denotes respectable subfield outlets; *limited* denotes narrower-impact, regional, specialized, newer, or field-mismatch outlets.

A small number of review-heavy journals (Annual Review series across psychology, political science, and sociology, plus *Psychological Bulletin* and *Nature Reviews Psychology*) and field-mismatch cross-listings (tourism/hospitality journals classified under sociology; environmental and accounting journals classified under public administration) were retained in the source universes and held out in sensitivity checks, because their citation dynamics differ from those of empirical research-idea outlets. Held-out validation sets for the seven non-management fields are balanced to 50 articles per tier (200 articles per field) and were strictly excluded from training.

**Research-idea extraction pipeline**

To standardize evaluator inputs across all training and evaluation conditions, structured research-idea descriptions were extracted from each source article using Qwen3-235B-A22B-Instruct (Alibaba). The extraction prompt instructed the model to produce a structured description of the core research question, theoretical motivation, methodological approach, and claimed contribution, while omitting empirical results, publication venue, and author identities. Extraction was validated by sampling outputs against alternative large language models (Claude Sonnet 3.5, Qwen3-32B-Instruct, others); no substantive differences were observed, and Qwen3-235B-A22B-Instruct was selected as the production extractor on output coherence and completeness.

Five output versions were produced per article: a 40–60-word core_rq_short; a 120–150-word rq_with_context (used as the primary evaluator input across all training and evaluation); a 100–130-word gap_focused; a 100–130-word theory_and_model; and an 80–100-word contribution_focused. Critical extraction rules required preserving authors' exact terminology, matching their certainty level, avoiding added theoretical sophistication, and avoiding inferred contributions not stated in the source. The compressed-input probe uses the short core_rq_short representation; because that field permits up to two or three sentences, the main text describes it as a compressed short idea statement rather than a literal one-sentence input.

The full extraction prompt is reproduced verbatim below.

````
# ROLE
You are an objective research paper analyzer. Your task is to extract and present research
questions and core elements from academic papers WITHOUT interpretation, embellishment, or
improvement.

# CRITICAL PRINCIPLE: OBJECTIVITY OVER PERSUASIVENESS
- Present the paper EXACTLY as written by the authors
- Do NOT add theoretical sophistication if it's not there
- Do NOT create compelling hooks if the original lacks them
- Do NOT infer contributions beyond what authors explicitly state
- Do NOT improve weak framing - describe it as presented
- If the idea seems underdeveloped in the original, your summary should reflect that

Your goal: Represent the research proposal exactly as the authors present it - the way a
doctoral student would pitch their idea to an advisor. Convey their thinking faithfully,
including any lack of polish or theoretical sophistication, so the professor can understand
and evaluate the original idea.

# OUTPUT STRUCTURE
Generate exactly 5 versions in JSON format:

# VERSION 1: CORE_RQ_SHORT
Purpose: Distill the essential research question(s)
Word count: 40-60 words (2-3 sentences maximum)

# VERSION 2: RQ_WITH_CONTEXT
Purpose: Add just enough context for a professor to evaluate the idea's merit
Word count: 120-150 words (1 paragraph)
Structure: phenomenon/problem; gap; research question; approach/framework; key claimed
contribution.

# VERSION 3: GAP_FOCUSED
Purpose: Emphasize what's unknown and how this study addresses it
Word count: 100-130 words

# VERSION 4: THEORY_AND_MODEL
Purpose: Describe the theoretical framework and research model
Word count: 100-130 words

# VERSION 5: CONTRIBUTION_FOCUSED
Purpose: Extract what the authors claim as their contributions
Word count: 80-100 words

# EXTRACTION RULES
- Focus on the abstract, introduction, and theoretical-development sections.
- Use the authors' exact terminology for key constructs.
- Preserve the level of theoretical sophistication in the original.
- Match the certainty level (e.g., "explores" vs. "demonstrates").
- Do NOT add theoretical connections, persuasive hooks, or inferred contributions.

# JSON OUTPUT FORMAT

```json
{
"core_rq_short": "string",
"rq_with_context": "string",
"gap_focused": "string",
"theory_and_model": "string",
"contribution_focused": "string"
}
```
````

A worked example for the article *Game Over or Game Changer? The Impact of Applicants' Gaming Skills on Their Hirability* yields the following compressed and full research-pitch representations:

core_rq_short: This study examines how applicants' gaming skills, presented as an extracurricular activity (ECA) on a resume, affect their perceived hirability and resume quality. It specifically compares gaming to team sports and tests whether proficiency level (neutral vs. high) influences these perceptions.

rq_with_context: The increasing digitalization of work has raised interest in nontraditional skills such as those developed through video gaming, which may include strategic thinking, teamwork, and digital fluency. However, hiring managers may hold negative stereotypes about gamers, viewing them as lazy or socially isolated, potentially disadvantaging applicants who list gaming on their resumes. While research shows gaming can enhance job-relevant cognitive abilities, there is no empirical evidence on how such skills are perceived during resume screening. Using a 2 (ECA: gaming vs. volleyball) × 2 (proficiency: neutral vs. high) experimental design, this study investigates whether listing gaming as an ECA affects applicant evaluations compared to traditional team sports. The authors apply signaling theory to understand how ECAs serve as cues about unobservable applicant traits. The primary contribution is initial evidence on the perception of gaming skills in personnel selection, highlighting a disconnect between potential skill benefits and actual hiring biases.

A second example, *Good intentions, bad outcomes: how and when family motivation leads to work-family conflict*, illustrates the same compression:

core_rq_short: This study examines how family motivation leads to work-family conflict through increased work effort, resulting in negative spousal interactions. It also investigates whether family supportive supervisor behaviors (FSSBs) buffer this negative resource drain process.

rq_with_context: Family motivation is widely seen as a positive driver of work performance, yet its effects on employees' family lives remain underexplored. While prior research highlights benefits in the work domain, little is known about potential downsides for family well-being. This study addresses this gap by investigating whether high family motivation, despite good intentions, can lead to work-family conflict (WFC) and negative spousal interactions due to excessive work effort depleting personal resources. Drawing on resource drain theory, the authors propose that FSSBs from supervisors serve as external resources that may mitigate this drain. Using a three-wave dyadic survey design with employee-partner data, the study tests a mediated moderation model. The key contribution lies in revealing the "dark side" of family motivation and identifying organizational support as a boundary condition.

### Cross-field transfer test (four SFT architectures × seven non-management fields)

The management-to-other-fields transfer probe covers all seven non-management fields and all four management-trained SFT architectures (GPT-4.1, GPT-4.1-nano, Qwen3-30B-A3B, Qwen3-4B). The full protocol — fine-tune identifiers, decoding settings, prompt scaffold, and per-field results — is reported in SM9. Per-architecture, per-field cross-transfer accuracy with Wilson 95% confidence intervals and the in-domain → cross-field transfer gap are tabulated in ST17; the same matrix is visualized in SF3.

## Supplementary Methods 6 (SM6): Alternative SFT Ensemble Pairings and Mechanism Probes (Management)

The management ensemble sensitivity combines SFT model probabilities by averaging the four-class softmax outputs. Table SM6a reports all two-model pairs, together with the three-model and four-model averages. The results show that ensembling can provide a limited gain in management, but the gain is small relative to the single-model SFT results and is not a primary endpoint. Cross-field ensemble sensitivity is reported separately in SM7/ST14 and shows similarly limited and field-dependent gains.

**Table SM6a. All two-model SFT ensembles in management, with the three-model and four-model ensemble for completeness.**

| Model 1 | Model 2 | Accuracy (120 items) |
|---|---|---|
| GPT-4.1-nano (SFT) | Qwen3-30B-A3B (SFT) | 60.83% |
| GPT-4.1 (SFT) | Qwen3-30B-A3B (SFT) | 60.00% |
| GPT-4.1 (SFT) | Qwen3-4B (SFT) | 60.00% |
| GPT-4.1-nano (SFT) | Qwen3-4B (SFT) | 60.00% |
| GPT-4.1-nano (SFT) | GPT-4.1 (SFT) | 59.17% |
| Qwen3-30B-A3B (SFT) | Qwen3-4B (SFT) | 59.17% |
| Three-model (nano + Q4B + Q30B SFT) | – | 60.00% |
| Four-model (incl. GPT-4.1 SFT) | – | 60.00% |

The accuracy spread across pairs is narrow (59.17–60.83%, range 1.7 percentage points), and the three-model and four-model averages do not materially improve on the two-model pairs. Probability-averaging ensembles may provide limited additional gains in some settings, but the improvement is small and inconsistent across fields. The cross-field analog is therefore reported only as a sensitivity analysis in SM7 and ST14.

## Management mechanism and robustness probes

The following management-only probes support the main-text mechanism figure and are not part of the eight-field primary evaluator set. They are included here so that the management depth analyses remain traceable within the SI.

**Table SM6b. Management mechanism and robustness probes.**

| Probe | Design | Key result | Role in SI |
|---|---|---|---|
| RL reasoning probe | Chain-of-thought RL checkpoint on the 120-article management benchmark | Run-pooled accuracy 40.3%, above the frontier mean but below every SFT single model (55.0–59.2%) | Tests whether articulated reasoning optimization recovers the same signal as SFT; objective and reward are specified below |
| GPT-5.5 chat-vs-reasoning evaluation | GPT-5.5 comparison spanning all eight field benchmarks, chat/log-probability classification with reasoning disabled versus high-reasoning inference with eight runs per item | GPT-5.5 chat/log-probability: 32.3% item-weighted; GPT-5.5 High pitch-mean: 31.2%; paired $t$ test $P = 0.0468$, high reasoning lower; fixed-denominator majority-vote McNemar $P = 0.0116$ | Tests whether current frontier reasoning improves evaluative judgment outside the SFT/RL training setting |
| Pairwise head-to-head transfer | GPT-4.1 SFT, Gemini 3.1 Pro, GPT-5.2 High, and GPT-4.1 base on 300 pairwise comparisons from the March 2026 management pairwise run | SFT GPT-4.1: 84.33% overall and 78.67% on adjacent-tier pairs; Gemini 3.1 Pro: 77.33%; GPT-5.2 High: 78.67%; GPT-4.1 base: 76.00% | Tests transfer from four-class classification to a label-free comparison format; the later GPT-5.5 evaluation was run on the four-class benchmark rather than this pairwise task |
| Compressed-input transfer | Full research pitch vs compressed short idea statement on the same 120 articles | GPT-4.1 SFT: 55.0% -> 49.2%; GPT-4.1-nano SFT: 57.5% -> 33.3% | Tests whether the learned signal survives removal of rich contextual detail |
| Temporal stability | Older 2015–2020 trace slice vs recent trace slice | Qwen3-30B-A3B SFT: 46.7% older vs 58.3% recent; matched two-model ensemble: 47.5% older vs 60.8% recent | Shows signal persistence with calibration drift under older institutional traces |
| SFT consensus | Agreement among four management SFT checkpoints | 4/4 consensus: 70.59% accuracy (36/51) at 42.50% coverage; >=3/4 consensus: 66.0% at 80.8% coverage; 2/4 split: 34.8% | Supports selective triage using cross-model agreement |

## RL reasoning probe: objective, reward, and training configuration

The RL reasoning probe was a management-only mechanism test, not part of the eight-field primary evaluator set. It asks whether explicit chain-of-thought policy optimization can recover the same institutional-trace signal captured by direct SFT. We trained reasoning-enabled Qwen3-4B-Thinking and Qwen3-32B checkpoints with a modified Group Relative Policy Optimization (GRPO)-style objective.

The RL probe used a DAPO-style improvement to the GRPO objective, with group-sampled rollouts, clipped policy-ratio updates, token-level normalization, and asymmetric clipping. Evaluation then used the same 120-article management benchmark and the same eight-sample stochastic protocol used for frontier reasoning controls; Figure 5a reports the run-pooled accuracy.

For each prompt $q$, the policy sampled a group of $G$ chain-of-thought completions $o_i$, each ending with one of the four tier labels. The token-level policy ratio for token $t$ in completion $i$ was

$$r_{i,t}(\theta)=\frac{\pi_\theta(o_{i,t}/q,o_{i,<t})}{\pi_{ref}(o_{i,t}/q,o_{i,<t})}.$$

The GRPO-style clipped surrogate was averaged over generated tokens rather than over completions:

$$L_{GRPO}(\theta)=-\frac{1}{\sum_{i=1}^{G}|o_i|}\sum_{i=1}^{G}\sum_{t=1}^{|o_i|}min\left(r_{i,t}(\theta)\hat{A}_i,clip\left(r_{i,t}(\theta),1-\epsilon,1+\epsilon+\epsilon_{higher}\right)\hat{A}_i\right).$$

This token-level normalization prevents longer reasoning traces from receiving larger update weight solely because they contain more tokens. The asymmetric upper clip $1+\epsilon+\epsilon_{higher}$ gives positive-advantage trajectories a wider update range than negative-advantage trajectories.

Rewards were ordinal and consistency-gated. The final label parsed from the completion, $\hat{y}_i^{label}$, and the label implied by the reasoning chain, $\hat{y}_i^{reasoning}$, had to agree for the completion to receive reward:

$$R(o_i,y)=1\left[\hat{y}_i^{label}=\hat{y}_i^{reasoning}\right]\cdot r(\hat{y}_i^{label},y),$$

with

$$r(\hat{y},y)=\begin{cases}1.0, & |\hat{y}-y|=0\\ 0.3, & |\hat{y}-y|=1\\ 0, & |\hat{y}-y|\geq 2.\end{cases}$$

Group-normalized advantages were computed as

$$\hat{A}_i=\frac{R_i-\mu_G}{\sigma_G+\epsilon},$$

with per-group reward mean $\mu_G$ and standard deviation $\sigma_G$. The consistency gate suppresses reward when the reasoning chain and final label diverge; the adjacent-tier partial credit preserves the ordinal structure of the four-class task.

**Algorithm SM6. Adaptive RL sampling and update.**

1. For each training prompt, draw K = 8 diagnostic completions with the current policy.
2. Estimate prompt-level accuracy from the diagnostic completions.
3. Route prompts with estimated accuracy below tau = 0.2 to guided mode during training; above-threshold prompts use pure GRPO or are skipped once reliably solved.
4. In guided mode, generate hindsight-style oracle hints from failed drafts and the ground-truth tier label, and prepend the hint during the training rollout only.
5. Sample the GRPO group, parse the final tier labels, apply the consistency-gated ordinal reward, normalize advantages within the group, and update with the token-normalized objective above.
6. Remove all guidance at evaluation time; the reported benchmark uses ordinary chain-of-thought sampling followed by final-label parsing.

**Table SM6c. RL mechanism-probe objective and configuration.**

| Component | Setting |
|---|---|
| Scope | Management-only mechanism probe; not included in the eight-field primary evaluator set |
| Base checkpoints | Reasoning-enabled Qwen3-4B-Thinking and Qwen3-32B |
| Objective | Modified GRPO-style policy objective with no KL penalty, token-level normalization, and asymmetric clipping |
| Reward | Consistency-gated ordinal reward: exact tier match = 1, adjacent tier = 0.3, distance >=2 = 0, multiplied by a reasoning-label consistency indicator |
| Sampling | K = 8 diagnostic completions per prompt; prompts below tau = 0.2 routed to guided mode during training |
| Learning rates | Qwen3-4B-Thinking: 5e-5; Qwen3-32B: 1e-5 |
| Optimizer and precision | AdamW, batch size 32, betas 0.9/0.99, weight decay 0.05, max gradient norm 1.0, bf16 mixed precision |
| Clipping | epsilon = 0.2; epsilon_higher = 0.1 |
| Infrastructure | 8 x A100 GPUs; FlashAttention-2; SGLang inference; DDP for Qwen3-4B-Thinking and FSDP for Qwen3-32B |
| Training duration | Under one week for Qwen3-4B-Thinking; about one week for Qwen3-32B |
| Evaluation | Eight stochastic samples per pitch on the 120-article management benchmark; parsed final labels aggregated as in the frontier-control protocol |

## Supplementary Methods 7 (SM7): Cross-Field Calibration and Selective Prediction

### Cross-field ensembling

For each of the eight fields we computed single-model SFT accuracy for each of the three cost-effective SFT models (GPT-4.1-nano, Qwen3-4B, Qwen3-30B-A3B), all three two-model probability-averaging pairs, and the full three-model ensemble. The validation set is fixed at 200 items per non-management field and 120 items in management, so per-field accuracy lifts are directly comparable after accounting for the management sample size.

The full per-field table is reported as ST14 and visualized as SF9. The qualitative pattern is small and field-specific: management (+1.6 pp), economics (+3.5 pp), and communication (+1.5 pp) improve modestly, while the remaining five non-management fields are flat or negative. The cross-field claim therefore relies on the best single SFT per field (Qwen3-30B-A3B in six of eight fields, Qwen3-4B in management, and GPT-4.1-nano in business and finance); ensembling is a sensitivity check only.

### Cross-field calibration

For each (SFT model, field) cell we compute confidence as exp(max log-probability) and report (i) accuracy, (ii) mean confidence on correct vs incorrect predictions, (iii) the gap between them and its Mann–Whitney U p-value, and (iv) ECE in 10 equal-width bins over [0.25, 1.00]. The 24 (model, field) cells all show a positive correct-vs-incorrect confidence gap at $P$ <= 0.014, indicating that SFT models reliably encode internal evidence about which of their own predictions are likely to be correct. Across-field heterogeneity is substantial: the mean confidence gap ranges from +0.066 (management) to +0.231 (psychology), and ECE ranges from 0.043 (psychology) to 0.189 (management). The summary calibration table is reported as ST2 and supports the main-text calibration results.

### Selective prediction

Selective prediction was computed as defined in SM4. The full per-(model, field) top-K confidence coverage curves are reported as ST16. The dominant pattern is that selective prediction recovers >=80% accuracy at modest coverage (<=35% of items) for the majority of (SFT model, field) cells, including in fields whose full-coverage prediction accuracy is otherwise in the 50–60% range. The qualitative implication is that even in the harder fields, SFT confidence usefully ranks predictions and enables selective deployment.

### Field-level rank correlations

Across the eight fields, accuracy and ECE rank-correlate at Spearman $r$ = -0.381 ($P$ = 0.352), and accuracy and confidence-gap rank-correlate at $r$ = +0.333 ($P$ = 0.420). Both are in the expected direction (higher-accuracy fields are better calibrated and have larger discrimination gaps) but neither reaches significance with $N$ = 8 fields. This is consistent with the field-heterogeneity argument in SM8: the cross-field accuracy spread (55.0%–85.5%) is dominated by field-level achievable ceiling rather than by the SFT method.

## Supplementary Methods 8 (SM8): Label-Noise Ceiling and Field Heterogeneity

Publication outcomes are not solely determined by research-idea quality. Execution fidelity, writing quality, reviewer-manuscript fit, and editorial discretion all contribute to final publication decisions, while the standardized inputs used here capture only the idea dimension. This gap between input features and outcome labels introduces inherent noise that places a theoretical ceiling on achievable classification accuracy: even a perfect evaluator of research-idea quality would not achieve perfect agreement with publication outcomes. Several factors contribute to this noise floor: a strong research idea may be published in a lower-tier journal due to poor execution, while a modest idea may reach a top-tier journal through exceptional methods and writing; the standardized inputs strip execution information and therefore cannot account for this variance. Publication decisions also depend partly on reviewer-manuscript fit and editorial discretion, which introduce stochastic variation unrelated to the underlying idea. Some journals sit near boundaries between adjacent tiers, so any deterministic mapping necessarily compresses a continuous quality distribution into discrete classes.

This noise floor is field-specific. Fields with more standardized publication conventions, more consistent review norms, or sharper tier boundaries should have a higher achievable ceiling; fields with more diffuse boundaries or more variable review norms should have a lower ceiling. The cross-field results are consistent with this: psychology (multidisciplinary), where publication conventions (structured abstracts, explicit hypotheses, consistent IMRaD formatting) are most standardized, achieves 85.5% SFT prediction accuracy, the highest of any field; public administration, where the journal hierarchy is shallower and several exceptional-tier journals overlap thematically with adjacent fields, achieves 55.0%, the lowest of the seven non-management fields. We do not interpret the cross-field accuracy spread as evidence that SFT works less well in some fields. We interpret it as evidence that the achievable ceiling differs by field, in line with the noise floor and field-heterogeneity argument above. Observed accuracies should therefore be interpreted relative to each field's ceiling, not against a 100% standard. Critically, this noise affects all evaluated systems equally (frontier models, fine-tuned models, and base controls), so all relative performance comparisons remain internally valid within field.

## Supplementary Methods 9 (SM9): Cross-Field Transfer of Management-Trained SFT Models

### Motivation

The eight-field SFT results in ST1 / ST13 are obtained by training one SFT model per field on that field's institutional traces. SM9 reports a stricter exploratory probe: how do SFT models trained *only* on management traces perform when evaluated zero-shot on each of the other seven fields' held-out validation sets, with no target-field exposure during fine-tuning? This single-source transfer experiment isolates whether the management-learned alignment captures any cross-field-portable component of evaluative judgment; it is not part of the primary cross-field evidence, which comes from field-specific SFT.

### Procedure

Four management-only SFT checkpoints were prepared: GPT-4.1 (SFT), GPT-4.1-nano (SFT), Qwen3-30B-A3B-Instruct (SFT), and Qwen3-4B-Instruct (SFT); training details follow SM1, with the corpus drawn from the 19-journal management source universe (ST3). These checkpoints were trained with the general social-science prompt used for SFT and base-model evaluation, not the management-specific frontier-control prompt. Each checkpoint was therefore applied zero-shot to each of the seven non-management fields' held-out validation sets (N = 200 per field) using that same general social-science evaluation prompt (SM2), avoiding a prompt-shift confound between training and transfer inference. All transfer evaluations used deterministic first-token classification over the four-tier label vocabulary; Qwen outputs used a token-prefix label resolver to handle tokenizer truncation of multi-token labels. For economics, GPT-4.1 predictions were regenerated from the same log-probability representation used by the SFT evaluator, reproducing the documented 43.5% prediction accuracy exactly (87 of 200). No economics, business-and-finance, communication, political-science, psychology, public-administration, or sociology pitches were seen during management fine-tuning. The held-out validation sets used here are the same ones used for in-domain SFT evaluation in ST1, ST13, and ST14, so per-field transfer accuracy and in-domain best-SFT accuracy are directly comparable on the same items.

All four management-SFT checkpoints have full cross-transfer coverage on the seven non-management fields (4 architectures × 7 fields = 28 rows). The same restricted-vocabulary first-token argmax prediction protocol was applied across hosted and locally served checkpoints. Qwen outputs were resolved to the four-class label space using a token-prefix matcher that achieved 100% resolution on a held-out 50-prompt validation. All 28 cross-transfer rows enter the SM9 result summary; the per-architecture, per-field cells are reported in ST17.

**Result summary**

The completed 4 × 7 transfer results are reported in ST17 and visualized in SF3. Descriptively, GPT-4.1 has the highest seven-field mean transfer accuracy (37.9%) and the smallest in-domain → cross-field drop-off (−17.1 pp), while the three smaller fine-tunes have lower mean transfer (32.4–33.3%). Because the four management-trained architectures have different management in-domain baselines, ST17 reports each transfer accuracy alongside that architecture's management reference and the transfer gap. Cross-architecture contrasts are interpreted descriptively rather than as a common-baseline model ranking. ST17 includes the cross-architecture accuracy-vector correlations as a compact diagnostic, not as a primary claim.

Field-distance correlations are not used as primary evidence because the transfer probe has only seven target fields per architecture. Per-tier breakdowns show systematic over-prediction of *strong* and under-prediction of *limited* across all seven destinations, so per-field tier-distribution priors may also contribute to the variation in transfer accuracy.

The full per-field transfer accuracy table with Wilson 95% confidence intervals and in-domain → cross-field transfer gaps is reported in ST17. Underlying per-prediction records and per-field summary statistics are listed in the Data Availability statement.

## Supplementary Tables

### Supplementary Table 1 (ST1): Cross-Field Accuracy by Evaluator Class (Eight Fields)

**Table ST1. Cross-field accuracy. Best single-model SFT accuracy and best available non-SFT control per field (architecture-matched base, frontier, or chat/log-probability control; 200 articles per field for each of the seven non-management fields, 120 articles in management; tier balance is 50 articles per tier for the non-management fields and 30 articles per tier in management).**

| Field | Best SFT model | SFT accuracy | Best non-SFT control | Control accuracy |
|---|---|---|---|---|
| Psychology | Qwen3-30B-A3B | 85.5% | Gemini 3.1 Pro | 33.5% |
| Economics | Qwen3-30B-A3B | 69.5% | Gemini 3.1 Pro | 44.5% |

| Field | Best SFT model | SFT accuracy | Best non-SFT control | Control accuracy |
|---|---|---|---|---|
| Communication | Qwen3-30B-A3B | 67.5% | Qwen3-30B-A3B (base) | 31.5% |
| Sociology | Qwen3-30B-A3B | 65.5% | GPT-5.5 High | 31.9% |
| Political Science | Qwen3-30B-A3B | 58.5% | Gemini 3.1 Pro | 35.0% |
| Management | Qwen3-4B† | 59.2% | Gemini 3.1 Pro | 38.8% |
| Business and Finance | GPT-4.1-nano | 55.5% | Gemini 3.1 Pro | 33.5% |
| Public Administration | Qwen3-30B-A3B | 55.0% | Gemini 3.1 Pro | 30.0% |

†In management, Qwen3-4B is the sole best single SFT at 59.2%, with Qwen3-30B-A3B at 58.3%, GPT-4.1-nano at 57.5%, and GPT-4.1 at 55.0%. The primary cross-field summaries use the best single SFT in every field; ensemble sensitivity is separated in ST14 and SF9.

Management base-architecture cells in ST13 (Qwen3-30B-A3B base 22.5%; Qwen3-4B base 26.7%; GPT-4.1-nano base 25.0%) are derived under the same primary 120–150-word research-pitch representation and the first-token log-probability resolver used elsewhere in the SI. In economics, Gemini 3.1 Pro is the best frontier control at 44.5% and GPT-5.5 High is the high-reasoning evaluation control at 36.3% pitch-mean accuracy on the same evaluation set. In sociology, GPT-5.5 High narrowly exceeds Gemini 3.1 Pro (31.9% versus 31.0%) and is therefore the best frontier control in ST1. The historical GPT-5.2 chat/log-probability run reached 33.5% in economics, while the GPT-5.5 chat/log-probability run reached 38.0%. The reproducibility package contains the prediction records for SFT, base, and frontier controls in this table.

## Supplementary Table 2 (ST2): Confidence Calibration by Field

**Table ST2. Confidence calibration by field, all three SFT architectures. Mean confidence gap (mean confidence on correct minus mean confidence on incorrect predictions) is reported as the per-field mean across the three SFT architectures (Qwen3-30B-A3B, Qwen3-4B, GPT-4.1-nano), matching the per-field-mean gap reported in the main-text calibration results. Expected calibration error (ECE) is reported per architecture, computed in 10 equal-width bins over the [0.25, 1.00] confidence range. Max *P* is the largest (worst) Mann–Whitney *U* two-sided *P* value across the three architectures within the field; all 24 (architecture, field) cells have $P \leq 0.014$ (the corresponding within-field maximum is shown). The main-text calibration anchor is the GPT-4.1-nano ECE column because GPT-4.1-nano achieves the lowest mean cross-field ECE among the three SFT architectures.**

| Field | Conf. gap (mean) | ECE (Qwen3-30B-A3B) | ECE (Qwen3-4B) | ECE (GPT-4.1-nano) | Max *P* |
|---|---|---|---|---|---|
| Psychology, multidisciplinary | +0.237 | 0.045 | 0.065 | 0.045 | <0.001 |
| Sociology | +0.176 | 0.086 | 0.180 | 0.098 | <0.001 |
| Economics | +0.164 | 0.178 | 0.106 | 0.084 | <0.001 |
| Public Administration | +0.130 | 0.141 | 0.115 | 0.092 | <0.001 |
| Communication | +0.115 | 0.099 | 0.051 | 0.142 | <0.001 |
| Business and Finance | +0.116 | 0.219 | 0.110 | 0.103 | <0.001 |
| Political Science | +0.083 | 0.147 | 0.127 | 0.078 | 0.003 |
| Management | +0.067 | 0.146 | 0.358 | 0.096 | 0.014 |

The mean confidence gap ranges from +0.067 (management) to +0.237 (psychology). The per-field-mean ECE ranges from 0.097 (communication) to 0.200 (management), and GPT-4.1-nano has the lowest average ECE across fields (0.092). Across the 24 (architecture, field) cells, the largest *P* value is 0.014 (Qwen3-4B in management); all other cells have $P \leq 0.005$. GPT-4.1-nano achieves the lowest ECE in five fields and ties for lowest in psychology; the exceptions are sociology (Qwen3-30B) and communication (Qwen3-4B). Management ECE is largest for Qwen3-4B (0.358) and lowest for GPT-4.1-nano (0.096), motivating the GPT-4.1-nano main-text anchor for the cross-field calibration figure.

As an all-field log-probability comparator, GPT-5.5 chat/log-probability was evaluated on the same eight benchmarks with deterministic decoding and reasoning disabled. Management used the expert-rubric prompt; the seven other fields used the general social-science prompt. It remained near chance at full coverage (mean accuracy 32.4%; per-field range 28.5%–38.0%) and was severely overconfident under the same four-class confidence definition (mean ECE = 0.626; per-field range 0.584–0.660). These records are used as the grey min–max band and dashed mean curve in Figure 4b. Compared with the earlier GPT-5.2 chat/log-probability evaluation, GPT-5.5 improved mean accuracy by +4.1 percentage points and reduced mean ECE from 0.701 to 0.626, but it remained far below the SFT models and did not produce reliable selective triage. Because the management GPT-5.5 row used the expert-rubric prompt while the historical GPT-5.2 chat row used the general prompt, the seven-non-management rows provide the cleaner model-progress comparison; the management row is retained as a prompt-plus-model sensitivity. The sampled frontier-reasoning cohort is not included in the calibration table because those evaluations produced sampled labels rather than token-probability records; the separate chat/log-probability track is inventoried in ST19b.

**Table ST2b. GPT-5.5 all-field evaluation. GPT-5.5 chat/log-probability used reasoning disabled and supplies the confidence records used in Figure 4b. GPT-5.5 High used eight high-reasoning samples per item; pitch-mean accuracy averages correctness over the eight runs within item before field aggregation. Majority accuracy excludes tied pitches, following the majority-vote convention used elsewhere in the SI. GPT-5.2 chat/log-probability is retained as a historical comparator for model-progress sensitivity only.**

| Field | *n* | GPT-5.2 chat acc. | GPT-5.5 chat/logprob acc. | Delta | GPT-5.5 chat ECE | GPT-5.5 High pitch-mean acc. | GPT-5.5 High majority acc. |
|---|---|---|---|---|---|---|---|
| Mgmt | 120 | 27.5% | 34.2% | +6.7 pp | 0.602 | 27.1% | 25.2% |
| Econ | 200 | 33.5% | 38.0% | +4.5 pp | 0.584 | 36.3% | 35.5% |
| Bus. & Fin. | 200 | 27.0% | 33.0% | +6.0 pp | 0.623 | 33.3% | 34.3% |
| Comm. | 200 | 27.0% | 30.0% | +3.0 pp | 0.660 | 29.2% | 29.4% |
| Pol. Sci. | 200 | 27.0% | 32.0% | +5.0 pp | 0.634 | 32.4% | 33.2% |
| Psych. | 200 | 28.0% | 31.5% | +3.5 pp | 0.641 | 29.5% | 30.2% |
| Pub. Admin. | 200 | 29.0% | 28.5% | -0.5 pp | 0.639 | 27.8% | 28.0% |
| Sociol. | 200 | 27.5% | 32.0% | +4.5 pp | 0.626 | 31.9% | 31.6% |

Field abbreviations in ST2b are used only to keep the wide table legible in DOCX: Mgmt, Management; Econ, Economics; Bus. & Fin., Business and Finance; Comm., Communication; Pol. Sci., Political Science; Psych., Psychology (multidisciplinary); Pub. Admin., Public Administration; Sociol., Sociology.

Across all 1,520 items, GPT-5.5 chat/log-probability accuracy was 32.3% item-weighted and GPT-5.5 High pitch-mean accuracy was 31.2% (difference -1.2 pp for high reasoning; paired $t$ test $P = 0.0468$). A fixed-denominator majority-vote McNemar test likewise showed no reasoning gain ($P = 0.0116$, high reasoning lower). Prompt comparability is exact for the seven non-management GPT-5.2-to-GPT-5.5 chat/log-probability rows; the management chat row combines model change with a prompt change and is retained as a sensitivity row rather than a strict model-progress estimate. A strict GPT-5.2 High versus GPT-5.5 High eight-run comparison is available only in management: GPT-5.2 High pitch-mean 31.0% versus GPT-5.5 High 27.1% (difference -4.0 pp, paired $t$ test $P = 0.176$). This management-only high-reasoning classification comparison is distinct from the Figure 5b pairwise task; no GPT-5.5 pairwise result is included. We therefore do not claim an all-field GPT-5.2 High versus GPT-5.5 High reasoning comparison.

## Supplementary Table 3 (ST3): Management Journal-to-Tier Mapping (19 Journals)

**Table ST3. Management 19-journal source universe. Tier semantics: Exceptional / Strong / Fair / Limited (mapped from top / top- / good / fair source labels).**

| Tier | Journals (N) |
|---|---|
| Exceptional (6) | Academy of Management Journal; Academy of Management Review; Administrative Science Quarterly; Journal of Applied Psychology; Organization Science; Strategic Management Journal |
| Strong (3) | Journal of Management; Organizational Behavior and Human Decision Processes; Personnel Psychology |
| Fair (5) | Human Resource Management; Human Relations; Journal of Management Studies; Journal of Organizational Behavior; Leadership Quarterly |
| Limited (5) | Group & Organization Management; Journal of Business and Psychology; Journal of Managerial Psychology; Journal of Organizational Behavior Management; Journal of Personnel Psychology |

Held-out benchmark: 120 articles balanced 30 per tier. Of the 19 source journals, 17 are represented in the final benchmark; *Human Relations* and *Group & Organization Management* had no articles selected under the tier-balanced sampling constraints.

## Supplementary Table 4 (ST4): Economics Journal-to-Tier Mapping

**Table ST4. Economics journal-to-tier mapping. Tier rationale: Exceptional contains the "Top 5" economics journals (universal cross-subfield consensus[45]); Strong contains AEJ-series and major subfield flagships; Fair contains established Q1 journals; Limited captures more specialized or lower-impact outlets.**

| Tier | Journals (N) |
|---|---|
| Exceptional (5) | American Economic Review; Econometrica; Journal of Political Economy; Quarterly Journal of Economics; Review of Economic Studies |
| Strong (18) | American Economic Journal: Applied Economics; American Economic Journal: Economic Policy; American Economic Journal: Macroeconomics; American Economic Journal: Microeconomics; Econometric Theory; Economic Journal; Games and Economic Behavior; International Economic Review; Journal of Development Economics; Journal of Econometrics; Journal of Economic Theory; Journal of International Economics; Journal of Labor Economics; Journal of Public Economics; RAND Journal of Economics; Review of Economic Dynamics; Theoretical Economics; Journal of the European Economic Association |
| Fair (7) | Brookings Papers on Economic Activity; European Journal of Health Economics; International Journal of Emerging Markets; Journal of Economic History; Journal of Population Economics; New Political Economy; Quarterly Review of Economics and Finance |
| Limited (8) | Agricultural Economics; Amfiteatru Economic; ASTIN Bulletin; Journal of Cultural Economics; Journal of the Japanese and International Economies; Local Economy; Post-Soviet Affairs; Quantitative Economics |

Held-out benchmark: 200 articles balanced 50 per tier.

## Supplementary Table 5 (ST5): Business and Finance Journal-to-Tier Mapping

**Table ST5. Business and finance journal-to-tier mapping (35 journals; Exceptional = 3, Strong = 8, Fair = 10, Limited = 14).**

| Tier | Journals |
|---|---|
| Exceptional (3) | Journal of Financial Economics; Journal of Finance; Review of Financial Studies |
| Strong (8) | Journal of Financial and Quantitative Analysis; Journal of Banking & Finance; Journal of Corporate Finance; Journal of Monetary Economics; Journal of Money, Credit and Banking; Review of Finance; Journal of Risk and Insurance; Journal of Financial Intermediation |
| Fair (10) | Journal of Empirical Finance; Financial Management; Mathematical Finance; Journal of Financial Markets; Journal of Real Estate Finance and Economics; Journal of Behavioral and Experimental Finance; Emerging Markets Review; European Accounting Review; Financial Analysts Journal; Global Finance Journal |
| Limited (14) | Borsa Istanbul Review; Sustainability Accounting Management and Policy Journal; Journal of Sustainable Finance & Investment; Meditari Accountancy Research; International Journal of Accounting; Australian Accounting Review; Financial Accountability & Management; International Journal of Auditing; Journal of Accounting Literature; International Journal of Managerial Finance; Journal of Behavioral Finance; Fiscal Studies; Journal of Risk Finance; Journal of Information Systems |

Held-out benchmark: 200 articles balanced 50 per tier. Several limited-tier journals are accounting outlets cross-listed under JCR's Business, Finance category; their primary disciplinary identity is accounting rather than finance, and per-field sensitivity checks treat them as a flagged subset.

## Supplementary Table 6 (ST6): Communication Journal-to-Tier Mapping

**Table ST6. Communication journal-to-tier mapping (33 journals; Exceptional = 5, Strong = 10, Fair = 10, Limited = 8).**

| Tier | Journals |
| --- | --- |
| Exceptional (5) | Journal of Communication; Communication Research; Communication Theory; Human Communication Research; New Media & Society |
| Strong (10) | Journal of Computer-Mediated Communication; Digital Journalism; Journalism & Mass Communication Quarterly; Information Communication & Society; Journalism; Media Culture & Society; Communication Monographs; Mass Communication and Society; International Journal of Press/Politics; Communication Methods and Measures |
| Fair (10) | Television & New Media; Journal of Public Relations Research; European Journal of Communication; Research on Language and Social Interaction; Convergence; Media and Communication; Journal of Language and Social Psychology; Journal of Advertising Research; Communication Culture & Critique; Communication and Critical-Cultural Studies |
| Limited (8) | Asian Journal of Communication; Revista Latina de Comunicacion Social; Javnost - The Public; Journal of Business and Technical Communication; Technical Communication Quarterly; Business and Professional Communication Quarterly; Journal of Family Communication; International Journal of Sport Communication |

Held-out benchmark: 200 articles balanced 50 per tier.

## Supplementary Table 7 (ST7): Political Science Journal-to-Tier Mapping

**Table ST7. Political science journal-to-tier mapping (40 journals; Exceptional = 4, Strong = 9, Fair = 18, Limited = 9).**

| Tier | Journals |
| --- | --- |
| Exceptional (4) | American Political Science Review; American Journal of Political Science; Journal of Politics; International Organization |
| Strong (9) | Annual Review of Political Science (review-heavy; flagged); Comparative Political Studies; Journal of Peace Research; Public Opinion Quarterly; Political Science Research and Methods; International Studies Quarterly; Politics & Gender; Comparative Politics; Political Theory |
| Fair (18) | Contemporary Security Policy; International Journal of Press/Politics; Global Environmental Politics; Review of International Organizations; Regulation & Governance; International Studies Review; British Journal of Politics & International Relations; Review of Policy Research; Party Politics; Policy and Society; Politics & Society; Political Research Quarterly; Politics and Governance; Quarterly Journal of Political Science; Geopolitics; International Political Science Review; Latin American Politics and Society; Political Science Quarterly |
| Limited (9) | Frontiers in Political Science; Problems of Post-Communism; Contemporary Politics; Scandinavian Political Studies; Swiss Political Science Review; International Journal of Transitional Justice; Journal of International Relations and Development; Economics & Politics; Ethics & International Affairs |

Held-out benchmark: 200 articles balanced 50 per tier.

## Supplementary Table 8 (ST8): Psychology (Multidisciplinary) Journal-to-Tier Mapping

**Table ST8. Psychology (multidisciplinary) journal-to-tier mapping (30 journals; Exceptional = 4, Strong = 4, Fair = 9, Limited = 13).**

| Tier | Journals |
| --- | --- |
| Exceptional (4) | Psychological Bulletin (review-heavy; flagged); Psychological Science; Psychological Review; Annual Review of Psychology (review-heavy; flagged) |
| Strong (4) | American Psychologist; Psychological Methods; Nature Reviews Psychology (review-heavy; flagged); Journal of Environmental Psychology |
| Fair (9) | Journals of Gerontology Series B; Journal of Happiness Studies; Journal of Positive Psychology; Aggression and Violent Behavior (review-mixed; flagged); Psychology of Women Quarterly; Psychology of Sexual Orientation and Gender Diversity; European Psychologist; Technology, Mind, and Behavior; Journal of Community Psychology |
| Limited (13) | Computers in Human Behavior Reports; Qualitative Research in Psychology; Psychosocial Intervention; Adversity and Resilience Science; Psychological Reports; Scandinavian Journal of Psychology; Australian Psychologist; Journal of Reproductive and Infant Psychology; Psychology of Religion and Spirituality; Zeitschrift fur Psychologie; Ecopsychology; Psychologica Belgica; Journal of Pacific Rim Psychology |

Held-out benchmark: 200 articles balanced 50 per tier. *Perspectives on Psychological Science* (APS flagship, JIF 8.4) was flagged as an exceptional-tier omission during reconciliation but was not added to the source list to avoid mid-study tier-list expansion.

## Supplementary Table 9 (ST9): Public Administration Journal-to-Tier Mapping

**Table ST9. Public administration journal-to-tier mapping (26 journals; Exceptional = 3, Strong = 5, Fair = 7, Limited = 11).**

| Tier | Journals |
|---|---|
| Exceptional (3) | Public Administration Review; Journal of Public Administration Research and Theory; Governance |
| Strong (5) | Public Administration (Wiley); Public Management Review; Journal of European Public Policy; Journal of Policy Analysis and Management; Regulation & Governance |
| Fair (7) | Policy Sciences; American Review of Public Administration; Policy and Politics; Review of Public Personnel Administration; Public Policy and Administration; Perspectives on Public Management and Governance; Journal of Public Policy |
| Limited (11) | Climate Policy (field-mismatch; flagged); Journal of Accounting and Public Policy (field-mismatch; flagged); Science and Public Policy; Social Policy & Administration; Journal of Social Policy; Public Money & Management; Local Government Studies; Nonprofit Management & Leadership; Journal of Comparative Policy Analysis; Journal of Chinese Governance; Global Public Policy and Governance |

Held-out benchmark: 200 articles balanced 50 per tier. Two limited-tier journals (Climate Policy; Journal of Accounting and Public Policy) are environmental-science and accounting outlets cross-listed under public administration; both are flagged for sensitivity analysis.

## Supplementary Table 10 (ST10): Sociology Journal-to-Tier Mapping

**Table ST10. Sociology journal-to-tier mapping (45 journals; Exceptional = 4, Strong = 11, Fair = 20, Limited = 10).**

| Tier | Journals |
|---|---|
| Exceptional (4) | American Sociological Review; American Journal of Sociology; Annual Review of Sociology (review-heavy; flagged); Social Forces |
| Strong (11) | Journal of Marriage and Family; Journal of Health and Social Behavior; Sociological Methods & Research; Gender & Society; Social Networks; Social Problems; Theory and Society; Sociological Theory; Sociology (BSA); European Sociological Review; Socio-Economic Review |
| Fair (20) | Socius; Social Indicators Research; Information Communication & Society; Social Science Research; Sociological Quarterly; Politics & Society; Rural Sociology; European Journal of Social Theory; Journal of Cultural Economy; European Societies; American Journal of Cultural Sociology; Media Culture & Society; Agriculture and Human Values; Qualitative Sociology; International Political Sociology; Sociology of Race and Ethnicity; Race & Class; Journal for the Scientific Study of Religion; Mobilization; Cultural Sociology |
| Limited (10) | Human Studies; Annals of Tourism Research (field-mismatch; flagged); Scandinavian Journal of Hospitality and Tourism (field-mismatch; flagged); Social Science Quarterly; Human Ecology; Critical Sociology; Social Justice Research; Community Work & Family; Sociological Spectrum; Young |

Held-out benchmark: 200 articles balanced 50 per tier. Two limited-tier journals (*Annals of Tourism Research*; *Scandinavian Journal of Hospitality and Tourism*) are tourism/hospitality outlets cross-listed under sociology; both are flagged for sensitivity analysis.

## Supplementary Table 11 (ST11): Per-Field Training Corpus and Validation Set Sizes

**Table ST11. Training corpus and held-out validation sizes by field.**

| Field | Training rows (post-split) | Held-out validation N | Tier balance (validation) |
|---|---|---|---|
| Management | 4,479 | 120 | 30 per tier |
| Economics | 5,593 | 200 | 50 per tier |
| Business and finance | 4,550 | 200 | 50 per tier |
| Communication | 2,587 | 200 | 50 per tier |
| Political science | 2,931 | 200 | 50 per tier |

| Field | Training rows (post-split) | Held-out validation N | Tier balance (validation) |
|---|---|---|---|
| Psychology, multidisciplinary | 2,476 | 200 | 50 per tier |
| Public administration | 2,251 | 200 | 50 per tier |
| Sociology | 2,094 | 200 | 50 per tier |

Validation rows were held out before any SFT training and were never seen during fine-tuning. Field-specific training corpora are disjoint from validation by construction; no benchmark articles appear in any training slice.

## Supplementary Table 12 (ST12): Per-Tier Precision/Recall/F1 for the Best Single SFT Model in Each Field

**Table ST12. Per-tier precision, recall, and F1 for the best single SFT model per field.**

Among best-single-SFT configurations, Qwen3-30B-A3B-Instruct leads in six of eight fields, Qwen3-4B leads in management, and GPT-4.1-nano leads in business and finance (55.5% vs 53.5% for Qwen3-30B). ST12 therefore reports the best single SFT for each field; ensemble sensitivity is separated in ST14 and SF9. All values are percentages.

| Field | Evaluator | Tier | Precision | Recall | F1 |
|---|---|---|---|---|---|
| Management | Qwen3-4B SFT | Exceptional | 61.5 | 80.0 | 69.6 |
| Management | Qwen3-4B SFT | Strong | 56.7 | 56.7 | 56.7 |
| Management | Qwen3-4B SFT | Fair | 45.7 | 53.3 | 49.2 |
| Management | Qwen3-4B SFT | Limited | 87.5 | 46.7 | 60.9 |
| Economics | Qwen3-30B SFT | Exceptional | 63.9 | 46.0 | 53.5 |
| Economics | Qwen3-30B SFT | Strong | 47.6 | 78.0 | 59.1 |
| Economics | Qwen3-30B SFT | Fair | 97.4 | 74.0 | 84.1 |
| Economics | Qwen3-30B SFT | Limited | 90.9 | 80.0 | 85.1 |
| Business and finance | GPT-4.1-nano SFT | Exceptional | 57.0 | 90.0 | 69.8 |
| Business and finance | GPT-4.1-nano SFT | Strong | 46.5 | 40.0 | 43.0 |
| Business and finance | GPT-4.1-nano SFT | Fair | 41.9 | 26.0 | 32.1 |
| Business and finance | GPT-4.1-nano SFT | Limited | 70.2 | 66.0 | 68.0 |
| Communication | Qwen3-30B SFT | Exceptional | 52.7 | 58.0 | 55.2 |
| Communication | Qwen3-30B SFT | Strong | 73.0 | 54.0 | 62.1 |
| Communication | Qwen3-30B SFT | Fair | 61.2 | 82.0 | 70.1 |
| Communication | Qwen3-30B SFT | Limited | 92.7 | 76.0 | 83.5 |
| Political science | Qwen3-30B SFT | Exceptional | 69.7 | 46.0 | 55.4 |
| Political science | Qwen3-30B SFT | Strong | 48.1 | 76.0 | 58.9 |
| Political science | Qwen3-30B SFT | Fair | 56.9 | 66.0 | 61.1 |
| Political science | Qwen3-30B SFT | Limited | 76.7 | 46.0 | 57.5 |
| Psychology, multidisciplinary | Qwen3-30B SFT | Exceptional | 89.8 | 88.0 | 88.9 |
| Psychology, multidisciplinary | Qwen3-30B SFT | Strong | 84.9 | 90.0 | 87.4 |
| Psychology, multidisciplinary | Qwen3-30B SFT | Fair | 79.6 | 86.0 | 82.7 |
| Psychology, multidisciplinary | Qwen3-30B SFT | Limited | 88.6 | 78.0 | 83.0 |
| Public administration | Qwen3-30B SFT | Exceptional | 56.8 | 50.0 | 53.2 |
| Public administration | Qwen3-30B SFT | Strong | 44.7 | 68.0 | 54.0 |
| Public administration | Qwen3-30B SFT | Fair | 50.0 | 30.0 | 37.5 |
| Public administration | Qwen3-30B SFT | Limited | 72.0 | 72.0 | 72.0 |

| Field | Evaluator | Tier | Precision | Recall | F1 |
|---|---|---|---|---|---|
| Sociology | Qwen3-30B SFT | Exceptional | 67.6 | 50.0 | 57.5 |
| Sociology | Qwen3-30B SFT | Strong | 53.7 | 58.0 | 55.8 |
| Sociology | Qwen3-30B SFT | Fair | 63.2 | 72.0 | 67.3 |
| Sociology | Qwen3-30B SFT | Limited | 78.8 | 82.0 | 80.4 |

Management per-tier metrics use the Qwen3-4B single SFT, matching ST1 and the best-single-SFT convention used for the other fields. Per-tier metrics for the in-domain Qwen3-30B SFT on economics are computed from the in-domain economics checkpoint's first-token log-probabilities using argmax over softmax-normalized restricted-vocabulary log-probabilities with the fixed alphabetical tie-break (exceptional < fair < limited < strong); aggregate prediction accuracy on this prediction set is 69.5% (139 / 200), matching the headline economics best-SFT accuracy in ST1 and ST13.

## Supplementary Table 13 (ST13): Eight-Field SFT-vs-Base Lift Detail

**Table ST13. SFT lift over architecture-matched base, per field and per SFT model.**

Lift is defined as prediction accuracy of the SFT model minus prediction accuracy of the architecture-matched base model on the same held-out validation set, in percentage points. The values are split into architecture-specific panels so the DOCX rendering preserves readable field names.

**Table ST13a. Qwen3-30B-A3B SFT versus architecture-matched base.**

| Field | SFT accuracy | Base accuracy | Lift (pp) |
|---|---|---|---|
| Management | 58.3 | 22.5 | +35.8 |
| Economics | 69.5 | 25.5 | +44.0 |
| Business and finance | 53.5 | 28.5 | +25.0 |
| Communication | 67.5 | 31.5 | +36.0 |
| Political science | 58.5 | 24.5 | +34.0 |
| Psychology, multidisciplinary | 85.5 | 31.5 | +54.0 |
| Public administration | 55.0 | 23.5 | +31.5 |
| Sociology | 65.5 | 28.0 | +37.5 |

**Table ST13b. Qwen3-4B SFT versus architecture-matched base.**

| Field | SFT accuracy | Base accuracy | Lift (pp) |
|---|---|---|---|
| Management | 59.2 | 26.7 | +32.5 |
| Economics | 64.0 | 25.0 | +39.0 |
| Business and finance | 53.5 | 25.5 | +28.0 |
| Communication | 61.0 | 25.0 | +36.0 |
| Political science | 50.5 | 25.0 | +25.5 |
| Psychology, multidisciplinary | 76.0 | 25.0 | +51.0 |
| Public administration | 51.0 | 25.0 | +26.0 |
| Sociology | 51.0 | 24.5 | +26.5 |

**Table ST13c. GPT-4.1-nano SFT versus architecture-matched base.**

| Field | SFT accuracy | Base accuracy | Lift (pp) |
|---|---|---|---|
| Management | 57.5 | 25.0 | +32.5 |
| Economics | 68.5 | 25.0 | +43.5 |
| Business and finance | 55.5 | 26.0 | +29.5 |
| Communication | 60.5 | 24.0 | +36.5 |
| Political science | 54.0 | 26.0 | +28.0 |

| Field | SFT accuracy | Base accuracy | Lift (pp) |
|---|---|---|---|
| Psychology, multidisciplinary | 81.5 | 30.0 | +51.5 |
| Public administration | 53.5 | 24.0 | +29.5 |
| Sociology | 54.0 | 26.5 | +27.5 |

All three management base controls (Qwen3-30B-A3B, Qwen3-4B, GPT-4.1-nano) use the primary research-pitch representation and the deterministic first-token log-probability resolver used elsewhere in the SI. The mean SFT-vs-base lift across the seven non-management fields x three SFT architectures (21 cells) is **+35.2 percentage points**; this is the headline cross-field SFT-effect statistic referenced in the main-text cross-field SFT-effect summary. The eight-field mean across all 24 (SFT model, field) cells is +35.0 pp (i.e., management's own SFT lift is essentially at the cross-field grand mean). We do not report a six-field subset mean here because excluding management and economics is not part of the primary cross-field estimand and can obscure the full eight-field pattern. Per-architecture seven-non-management-field means are Qwen3-30B +37.4 pp, GPT-4.1-nano +35.1 pp, Qwen3-4B +33.1 pp.

## Supplementary Table 14 (ST14): All Pairwise SFT Ensemble Combinations, Cross-Field

**Table ST14. Single-model accuracy, two-model pair accuracy, and three-model accuracy for every field x ensemble combination.**

All values are prediction accuracy on the field's held-out validation set. Pairs and triples use probability averaging (SM7). "Pair lift" = best two-model pair minus best single. "Triple lift" = three-model minus best single. The table is split into compact panels so the DOCX rendering preserves readable field names and model-pair labels.

**Table ST14a. Best single, best pair, and lift summary.**

| Field | Best single | Best pair | Three-model | Pair lift (pp) | Triple lift (pp) |
|---|---|---|---|---|---|
| Management | 59.2% (Q4B) | 60.8% (nano+Q30B) | 60.0% | **+1.6** | +0.8 |
| Economics | 69.5% (Q30B) | 73.0% (Q30B+nano) | 70.0% | **+3.5** | +0.5 |
| Business and finance | 55.5% (nano) | 54.5% (Q4B+nano) | 55.0% | **−1.0** | −0.5 |
| Communication | 67.5% (Q30B) | 69.0% (Q30B+Q4B) | 66.0% | **+1.5** | −1.5 |
| Political science | 58.5% (Q30B) | 57.0% (Q30B+Q4B) | 57.0% | **−1.5** | −1.5 |
| Psychology, multidisciplinary | 85.5% (Q30B) | 85.5% (Q30B+nano) | 85.5% | **+0.0** | +0.0 |
| Public administration | 55.0% (Q30B) | 54.5% (Q4B+nano) | 54.0% | **−0.5** | −1.0 |
| Sociology | 65.5% (Q30B) | 63.5% (Q30B+nano) | 58.5% | **−2.0** | −7.0 |

**Table ST14b. Single-model SFT accuracies by field.**

| Field | GPT-4.1-nano | Qwen3-4B | Qwen3-30B |
|---|---|---|---|
| Management | 57.5% | 59.2% | 58.3% |
| Economics | 68.5% | 64.0% | 69.5% |
| Business and finance | 55.5% | 53.5% | 53.5% |
| Communication | 60.5% | 61.0% | 67.5% |
| Political science | 54.0% | 50.5% | 58.5% |
| Psychology, multidisciplinary | 81.5% | 76.0% | 85.5% |
| Public administration | 53.5% | 51.0% | 55.0% |
| Sociology | 54.0% | 51.0% | 65.5% |

**Table ST14c. Two-model pair accuracies by field.**

| Field | nano+Q4B | nano+Q30B | Q4B+Q30B |
|---|---|---|---|
| Management | 60.0% | 60.8% | 59.2% |

| Field | nano+Q4B | nano+Q30B | Q4B+Q30B |
|---|---|---|---|
| Economics | 71.5% | 73.0% | 68.5% |
| Business and finance | 54.5% | 54.0% | 53.0% |
| Communication | 61.0% | 65.5% | 69.0% |
| Political science | 52.0% | 57.0% | 57.0% |
| Psychology, multidisciplinary | 81.0% | 85.5% | 85.0% |
| Public administration | 54.5% | 53.5% | 52.5% |
| Sociology | 54.0% | 63.5% | 59.0% |

Validation set: 200 items per non-management field; 120 items in management. The qualitative split is that ensembling helps where no single SFT model dominates: management gains +1.6 pp, and among the seven non-management fields, economics gains +3.5 pp and communication +1.5 pp, while the remaining five non-management fields are flat or negative. The mechanistic reading is that ensembling helps when no single model dominates (SM7).

## Supplementary Table 15 (ST15): Top-1+2 Accuracy Diagnostic by Field

**Table ST15. Top-1+2 accuracy (Top-1 OR Top-2 prediction matches the ground-truth tier), by SFT model, for fields with available second-label records.**

Top-1+2 accuracy treats the second-most-probable label as also acceptable. Boundary disagreements between adjacent tiers are routine even among human reviewers, so Top-1+2 is the natural ordinal-evaluation analog for a four-class ordinal label space. Values are percentages.

| Field | Qwen3-30B SFT | Qwen3-4B SFT | GPT-4.1-nano SFT |
|---|---|---|---|
| Business and finance | 85.5 | 82.5 | 85.0 |
| Communication | 89.0 | 88.5 | 90.0 |
| Political science | 84.5 | 79.5 | 82.0 |
| Psychology, multidisciplinary | 96.5 | 92.0 | 94.5 |
| Public administration | 74.5 | 74.5 | 74.5 |
| Sociology | 84.5 | 77.0 | 75.0 |

The field-level Top-1+2 values are high across the available records, indicating that adjacent-tier confusions are the dominant remaining error mode for SFT and that the SFT signal often places near-correct predictions within one tier. We do not report a subset mean because this diagnostic table excludes management and economics for record-availability reasons and is not used as a primary cross-field summary.

## Supplementary Table 16 (ST16): Selective Prediction Coverage Curves Across Fields and SFT Models

**Table ST16. SFT accuracy at top-K confidence coverage thresholds, per (model, field).**

Confidence is exp(max log-probability) over the four-class restricted vocabulary; predictions are sorted in descending order of confidence and accuracy is computed at each coverage level.

**Table ST16a. Qwen3-30B SFT — selective prediction by field.**

| Field | Top-10% | Top-15% | Top-20% | Top-25% | Top-33% | Top-50% | Top-100% |
|---|---|---|---|---|---|---|---|
| Management | 100.0 | 88.9 | 75.0 | 73.3 | 71.8 | 70.0 | 58.3 |
| Economics | 100.0 | 100.0 | 97.5 | 98.0 | 98.5 | 89.0 | 69.5 |
| Business and finance | 95.0 | 90.0 | 92.5 | 80.0 | 74.2 | 67.0 | 53.0 |
| Communication | 100.0 | 96.7 | 92.5 | 92.0 | 83.3 | 79.0 | 67.5 |
| Political science | 80.0 | 73.3 | 67.5 | 72.0 | 74.2 | 69.0 | 59.0 |
| Psychology, multidisciplinary | 100.0 | 100.0 | 97.5 | 98.0 | 97.0 | 98.0 | 86.0 |
| Public administration | 95.0 | 93.3 | 90.0 | 86.0 | 81.8 | 69.0 | 52.5 |

| Field | Top-10% | Top-15% | Top-20% | Top-25% | Top-33% | Top-50% | Top-100% |
|---|---|---|---|---|---|---|---|
| Sociology | 95.0 | 96.7 | 95.0 | 94.0 | 87.9 | 83.0 | 63.0 |

**Table ST16b. Qwen3-4B SFT — selective prediction by field.**

| Field | Top-10% | Top-15% | Top-20% | Top-25% | Top-33% | Top-50% | Top-100% |
|---|---|---|---|---|---|---|---|
| Management | 91.7 | 83.3 | 75.0 | 80.0 | 84.6 | 68.3 | 59.2 |
| Economics | 95.0 | 96.7 | 97.5 | 96.0 | 92.4 | 84.0 | 62.5 |
| Business and finance | 85.0 | 83.3 | 85.0 | 82.0 | 74.2 | 68.0 | 53.5 |
| Communication | 95.0 | 93.3 | 95.0 | 88.0 | 81.8 | 72.0 | 61.0 |
| Political science | 75.0 | 70.0 | 65.0 | 58.0 | 59.1 | 58.0 | 50.5 |
| Psychology, multidisciplinary | 100.0 | 100.0 | 95.0 | 96.0 | 97.0 | 96.0 | 77.0 |
| Public administration | 90.0 | 83.3 | 77.5 | 72.0 | 68.2 | 63.0 | 51.5 |
| Sociology | 100.0 | 100.0 | 92.5 | 88.0 | 81.8 | 67.0 | 51.5 |

**Table ST16c. GPT-4.1-nano SFT — selective prediction by field.**

| Field | Top-10% | Top-15% | Top-20% | Top-25% | Top-33% | Top-50% | Top-100% |
|---|---|---|---|---|---|---|---|
| Management | 100.0 | 94.4 | 79.2 | 76.7 | 69.2 | 66.7 | 57.5 |
| Economics | 100.0 | 100.0 | 100.0 | 96.0 | 93.9 | 85.0 | 68.5 |
| Business and finance | 90.0 | 83.3 | 87.5 | 82.0 | 74.2 | 66.0 | 56.0 |
| Communication | 100.0 | 90.0 | 82.5 | 84.0 | 78.8 | 67.0 | 59.5 |
| Political science | 90.0 | 86.7 | 77.5 | 72.0 | 71.2 | 65.0 | 53.5 |
| Psychology, multidisciplinary | 100.0 | 100.0 | 100.0 | 100.0 | 98.5 | 96.0 | 81.0 |
| Public administration | 95.0 | 80.0 | 77.5 | 74.0 | 71.2 | 63.0 | 54.0 |
| Sociology | 100.0 | 96.7 | 97.5 | 94.0 | 84.8 | 74.0 | 54.5 |

The dominant pattern is that selective prediction recovers >=80% accuracy at modest coverage for many (SFT model, field) cells, including for fields whose full-coverage prediction accuracy is otherwise in the 50–60% range. At 25% coverage, examples reaching at least 80% accuracy include management with Qwen3-4B, business and finance with all three SFT architectures, communication with all three SFT architectures, economics with all three SFT architectures, psychology with all three SFT architectures, sociology with all three SFT architectures, and public administration with Qwen3-30B. The qualitative implication is that even in the harder fields, SFT confidence usefully ranks predictions and enables selective deployment.

The main-text top-10% summary (94.6%) is the mean across all 24 model-field cells in ST16. Figure 4b displays the Qwen3-30B-A3B slice as a representative architecture and adds the all-field GPT-5.5 chat/log-probability comparator as a min–max range band plus mean curve; Figure 4d summarizes the per-field top-25% operating point.

## Supplementary Table 17 (ST17): Cross-Field Transfer of Management-SFT Models (Four Architectures × Seven Fields)

**Table ST17. Exploratory cross-field transfer of management-trained SFT models. Per-field prediction accuracy of each management-only-trained SFT checkpoint evaluated zero-shot on each non-management field's held-out validation set (N = 200 per field), alongside the in-domain best SFT accuracy from ST1 and the transfer gap (transfer minus in-domain best). All values are percentages; 95% confidence intervals are Wilson score intervals on n = 200 (cross-field cells) and n = 120 (management in-domain reference column).**

**Panel a. Mgmt-SFT GPT-4.1 → 7 fields.**

| Field | Transfer accuracy | 95% CI | In-domain best SFT (ST1) | Transfer gap (pp) |
|---|---|---|---|---|
| Public administration | 33.0% | [26.9, 39.8] | 55.0% | −22.0 |
| Business and finance | 39.0% | [32.5, 45.9] | 55.5% | −16.5 |
| Economics | 43.5% | [36.8, 50.4] | 69.5% | −26.0 |
| Sociology | 35.0% | [28.7, 41.8] | 65.5% | −30.5 |
| Communication | 31.5% | [25.5, 38.2] | 67.5% | −36.0 |

| Field | Transfer accuracy | 95% CI | In-domain best SFT (ST1) | Transfer gap (pp) |
|---|---|---|---|---|
| Psychology, multidisciplinary | 43.5% | [36.8, 50.4] | 85.5% | −42.0 |
| Political science | 40.0% | [33.5, 46.9] | 58.5% | −18.5 |
| **Mean (seven fields)** | **37.9%** | — | **65.3%** | **−27.4** |

In-domain reference: GPT-4.1 management SFT = 55.0% (n = 120); in-domain → cross-field mean drop-off = −17.1 pp.

**Panel b. Mgmt-SFT GPT-4.1-nano → 7 fields.**

| Field | Transfer accuracy | 95% CI | In-domain best SFT (ST1) | Transfer gap (pp) |
|---|---|---|---|---|
| Public administration | 25.0% | [19.5, 31.4] | 55.0% | −30.0 |
| Business and finance | 30.0% | [24.1, 36.7] | 55.5% | −25.5 |
| Economics | 37.0% | [30.6, 43.9] | 69.5% | −32.5 |
| Sociology | 34.0% | [27.8, 40.8] | 65.5% | −31.5 |
| Communication | 33.0% | [26.9, 39.8] | 67.5% | −34.5 |
| Psychology, multidisciplinary | 32.0% | [25.9, 38.8] | 85.5% | −53.5 |
| Political science | 41.0% | [34.4, 47.9] | 58.5% | −17.5 |
| **Mean (seven fields)** | **33.1%** | — | **65.3%** | **−32.2** |

In-domain reference: GPT-4.1-nano management SFT = 57.5% (n = 120); in-domain → cross-field mean drop-off = −24.4 pp.

**Panel c. Mgmt-SFT Qwen3-30B-A3B → 7 fields.**

| Field | Transfer accuracy | 95% CI | In-domain best SFT (ST1) | Transfer gap (pp) |
|---|---|---|---|---|
| Public administration | 25.0% | [19.5, 31.4] | 55.0% | −30.0 |
| Business and finance | 31.5% | [25.5, 38.2] | 55.5% | −24.0 |
| Economics | 38.0% | [31.6, 44.9] | 69.5% | −31.5 |
| Sociology | 38.5% | [32.0, 45.4] | 65.5% | −27.0 |
| Communication | 29.0% | [23.2, 35.6] | 67.5% | −38.5 |
| Psychology, multidisciplinary | 30.0% | [24.1, 36.7] | 85.5% | −55.5 |
| Political science | 41.0% | [34.4, 47.9] | 58.5% | −17.5 |
| **Mean (seven fields)** | **33.3%** | — | **65.3%** | **−32.0** |

In-domain reference: Qwen3-30B-A3B management SFT = 58.3% (n = 120); in-domain → cross-field mean drop-off = −25.0 pp.

**Panel d. Mgmt-SFT Qwen3-4B → 7 fields.**

| Field | Transfer accuracy | 95% CI | In-domain best SFT (ST1) | Transfer gap (pp) |
|---|---|---|---|---|
| Public administration | 25.5% | [20.0, 32.0] | 55.0% | −29.5 |
| Business and finance | 31.5% | [25.5, 38.2] | 55.5% | −24.0 |
| Economics | 36.0% | [29.7, 42.9] | 69.5% | −33.5 |
| Sociology | 33.5% | [27.3, 40.3] | 65.5% | −32.0 |
| Communication | 31.0% | [25.0, 37.7] | 67.5% | −36.5 |
| Psychology, multidisciplinary | 32.5% | [26.4, 39.3] | 85.5% | −53.0 |
| Political science | 37.0% | [30.6, 43.9] | 58.5% | −21.5 |
| **Mean (seven fields)** | **32.4%** | — | **65.3%** | **−32.9** |

In-domain reference: Qwen3-4B management SFT = 59.2% (n = 120); in-domain → cross-field mean drop-off = −26.8 pp.

**Panel e. Cross-architecture accuracy-vector correlations.**

Pearson correlations are computed across the seven non-management fields.

| | GPT-4.1 SFT | GPT-4.1-nano SFT | Qwen3-30B-A3B SFT | Qwen3-4B SFT |
|---|---|---|---|---|
| GPT-4.1 SFT | 1.000 | +0.449 | +0.436 | +0.626 |
| GPT-4.1-nano SFT | — | 1.000 | +0.891 | +0.955 |
| Qwen3-30B-A3B SFT | — | — | 1.000 | +0.914 |
| Qwen3-4B SFT | — | — | — | 1.000 |

The three smaller fine-tuned architectures (GPT-4.1-nano, Qwen3-30B-A3B, Qwen3-4B) produce near-collinear per-field accuracy patterns (pairwise Pearson $r \in [+0.89, +0.96]$); the flagship GPT-4.1 SFT departs from this cluster ($r \in [+0.44, +0.63]$). This diagnostic is reported to characterize the exploratory transfer probe only. It is consistent with the in-domain → cross-field drop-off pattern in panels a–d, where the flagship's 17.1 pp drop is the smallest of the four, but the primary cross-field claim remains field-specific SFT rather than management-to-field transfer.

*Cross-transfer compute-access footnote.* The two Qwen3 management-SFT checkpoints (Qwen3-30B-A3B and Qwen3-4B) used the same restricted-vocabulary first-token argmax prediction protocol applied to GPT-4.1 and GPT-4.1-nano. Qwen outputs were resolved to the four-class label space using a token-prefix matcher that achieved 100% resolution on a held-out 50-prompt validation. All four architectures' per-field cells include Wilson 95% confidence intervals; bootstrap summaries are available in the underlying records. The 5,600 prediction records supporting this probe (4 architectures × 7 fields × 200 articles) are included in the reproducibility package.

All transfer evaluations used the same general social-science evaluation prompt (SM2) and deterministic first-token classification. Hosted and locally served checkpoints used matched restricted-vocabulary resolvers, with prefix matching for Qwen labels. No target-field pitches were seen during management fine-tuning. The target-field baseline column reports each destination field's best in-domain SFT from ST1, while the panel-specific note under each architecture reports that management-trained model's own in-domain management reference; these two baselines are separated because the four source architectures have different management starting points. Field-distance correlations are not used as primary evidence because the transfer probe has only seven target fields per architecture.

## Supplementary Table 18 (ST18): Source-to-Unified Label Normalization

**Table ST18. Human survey label → model / analysis label mapping. Human raters used the field-facing survey labels shown in the left column. These responses were mapped once into the unified model / analysis label space before scoring.**

| Human survey label | Model / analysis label |
|---|---|
| Top | exceptional |
| Top- | strong |
| Good | fair |
| Fair | limited |

The key clarification is that human-facing *Fair* is the lowest survey option and maps to model / analysis *limited*, not to model / analysis *fair*. Model evaluators output or are scored over exceptional, strong, fair, and limited directly. This direct-label protocol is used because Top and especially Top- are poor labels for first-token log-probability scoring, whereas the analysis labels are distinct and parseable across SFT, base, chat log-probability, and sampled frontier outputs.

## Supplementary Table 19 (ST19): Model Inventory and Access Window

**Table ST19. Evaluator-model inventory. The evaluator set comprises three classes: frontier reasoning models accessed via API (eight chain-of-thought samples per pitch, modal vote or pitch-mean accuracy), chat/log-probability models accessed via API (single-pass top-token log-probability classification), and four SFT architectures fine-tuned on field-specific institutional traces (single-pass log-probability classification). The original management frontier cohort was accessed in March 2026; the GPT-5.5 evaluation was accessed separately in May 2026.**

**Table ST19a. Frontier reasoning models.**

| Model | Provider | Model version | Access window | Samples/pitch |
|---|---|---|---|---|
| Claude Opus 4.6 | Anthropic | claude-4.6-opus-20260205 | March 2026 | 8 |
| GPT-5.2 High | OpenAI | gpt-5.2 (reasoning=high) | March 2026 | 8 |
| Gemini 2.5 Pro | Google | gemini-2.5-pro | March 2026 | 8 |
| Gemini 3.1 Pro | Google | gemini-3.1-pro-preview-20260219 | March 2026 | 8 |
| Qwen 3.5 Plus | Alibaba | qwen3.5-plus-02-15 | March 2026 | 8 |
| DeepSeek V3.2 | DeepSeek | deepseek-v3.2-speciale-20251201 | March 2026 | 8 |
| Seed 2.0 Pro | ByteDance | doubao-seed-2-0-pro-260215 | March 2026 | 8 |
| MiniMax M2.5 | MiniMax | minimax-m2.5 | March 2026 | 8 |
| Kimi K2.5 | Moonshot AI | kimi-k2.5 | March 2026 | 8 |
| Grok 4.1 Fast | xAI | grok-4.1-fast | March 2026 | 8 |
| GLM-5 | Zhipu AI | glm-5 | March 2026 | 8 |
| GPT-5.5 High | OpenAI | GPT-5.5 high-reasoning configuration | May 2026 | 8 |

The March 2026 management cohort comprises the first 11 rows and is used for the broad management heterogeneity, collapse-pattern, and pairwise-comparison analyses. GPT-5.5 High is the May 2026 high-reasoning row used for the all-field GPT-5.5 chat-versus-high-reasoning comparison and as the OpenAI reasoning reference in the cross-field frontier-control figure. GPT-5.2 High remains the historical March 2026 management-cohort and pairwise comparator.

**Table ST19b. Chat / log-probability evaluators.**

| Model | Provider | Model version | Access window | Log-probability extraction |
|---|---|---|---|---|
| GPT-5.2 (chat) | OpenAI | gpt-5.2 | March 2026 | Top-token log-probabilities |
| GPT-5.5 (chat/logprob) | OpenAI | GPT-5.5 chat/log-probability configuration | May 2026 | Top-token log-probabilities |
| Kimi K2 (chat) | Moonshot AI | kimi-k2-0905-preview | March 2026 | Top-token log-probabilities |
| DeepSeek Chat | DeepSeek | deepseek-v3.2 | March 2026 | Top-token log-probabilities |

GPT-5.5 chat/logprob is used as the all-field log-probability comparator in Figure 4b. The reproducibility package includes item-level GPT-5.5 chat/log-probability predictions, selective-prediction summaries, and plotting summaries. GPT-5.2 chat is retained only as a historical comparison for model-progress sensitivity; the management row in that historical comparison used the general social-science prompt, whereas the GPT-5.5 management evaluation used the expert-rubric management prompt, so management-specific GPT-5.2-to-GPT-5.5 changes should be read as prompt-plus-model changes. The reasoning-model rows in ST19a belong to sampled frontier-reasoning evaluations; those sampled evaluations do not provide token log-probability records and are therefore not used for calibration or selective-prediction curves.

**Table ST19c. SFT architectures (n = 24 field-specific core checkpoints plus one management-only GPT-4.1 checkpoint).**

| Architecture | Type | Source | Training surface | Per-field corpus size (range) |
|---|---|---|---|---|
| Qwen3-30B-A3B-Instruct (2507) | Open-weight MoE (30B total / 3B active) | Alibaba | Local TRL on 8 × A100 | 2,094 – 5,593 |
| Qwen3-4B-Instruct (2507) | Open-weight dense | Alibaba | Local TRL on 1 × A100 | 2,094 – 5,593 |
| GPT-4.1-nano | Proprietary | OpenAI | OpenAI Fine-Tuning API | 2,094 – 5,593 |
| GPT-4.1 (management only) | Proprietary | OpenAI | OpenAI Fine-Tuning API | 4,479 (management corpus) |

GPT-4.1 SFT was trained on management only (used for management mechanism probes and the cross-field-transfer test reported in SM9). Qwen3-4B, Qwen3-30B-A3B, and GPT-4.1-nano SFT checkpoints exist for each of the eight fields. Per-field training corpus sizes are detailed in ST11. Inference for SFT models uses the four-class restricted-vocabulary first-token argmax described in SM2.

## Supplementary Table 20 (ST20): Human-Panel Filtering Sensitivity

**Table ST20. Sensitivity of human-panel accuracy to perfunctory-rater filtering. Junior raters who spent fewer than one minute on average per pitch were excluded from the filtered junior panel as perfunctory; no quality filter was applied to the expert panel because experts were recruited through direct one-to-one professional contact and individual time investment was not used as a quality cue. The unfiltered junior individual mean is 25.3% and the filtered junior individual mean is 31.7% (*P* = 0.066, marginally significant); the unfiltered expert majority vote on the 89 non-tied pitches is 41.6% (primary expert metric) and the filtered expert majority vote on 68 non-tied pitches is 39.7% (sensitivity check). The junior majority vote on the filtered panel reaches 40.8% on 103 non-tied pitches.**

| Panel | Filter | N raters | Individual mean accuracy | Majority-vote accuracy | Non-tied N (majority vote) | Role |
|---|---|---|---|---|---|---|
| Experts | Unfiltered | 48 | 36.2% | **41.6%** | 89 | Primary expert benchmark |
| Experts | Filtered (≥ 1 min/pitch) | 38 | — | 39.7% | 68 | Filtering sensitivity |
| Juniors | Unfiltered | 175 | 25.3% (*P* = 0.066) | — | — | Filtering sensitivity |
| Juniors | Filtered (≥ 1 min/pitch) | 174 | 31.7% | 40.8% | 103 | Primary junior benchmark |

The unfiltered–filtered junior individual-mean difference (25.3% vs 31.7%) is marginally significant at *P* = 0.066, indicating that perfunctory raters are a small but real source of accuracy depression in the junior panel. The unfiltered–filtered expert difference goes the opposite direction (unfiltered majority-vote 41.6% > filtered 39.7%) because the expert filter is null and the gap reflects only the per-rater minimum-coverage threshold used to define the non-tied subset in each panel; the primary expert benchmark is therefore unfiltered (41.6%, n = 89). These values are derived from the aggregation rules described in SM3.

## Supplementary Table 21 (ST21): Cross-Architecture Agreement (Management Benchmark)

Independent SFT models converge on a shared evaluative signal, an order of magnitude tighter than human raters.

**Table ST21. Cross-architecture agreement on the 120-article management benchmark. Pairwise Cohen's κ is computed on the four-class first-token argmax predictions of each pair; the human contrast row is multi-rater Fleiss' κ on the same benchmark, which is therefore a directional rather than strictly comparable contrast (Cohen's κ for two raters versus Fleiss' κ for many). All AI–AI pairs share the same item set; the human-pool κ values use the modal-truncation variant (truncate each pitch to the modal rater count: 3 for experts, 21 for juniors).**

| Comparison | Cohen's κ (AI–AI) or Fleiss' κ (humans) | Agreement % | Notes |
|---|---|---|---|
| GPT-4.1 SFT vs GPT-4.1-nano SFT | 0.500 | 63.33% | AI–AI |
| GPT-4.1 SFT vs Qwen3-30B-A3B SFT | 0.556 | 67.50% | AI–AI |
| GPT-4.1 SFT vs Qwen3-4B SFT | 0.510 | 64.17% | AI–AI |
| GPT-4.1-nano SFT vs Qwen3-30B-A3B SFT | 0.604 | 70.83% | AI–AI; primary ensemble pair |
| GPT-4.1-nano SFT vs Qwen3-4B SFT | 0.603 | 70.83% | AI–AI |
| Qwen3-30B-A3B SFT vs Qwen3-4B SFT | 0.511 | 64.17% | AI–AI |
| Expert panel (multi-rater) | 0.049 | – | Fleiss' κ, 118 pitches at modal n = 3 raters |
| Junior panel (multi-rater) | 0.030 | – | Fleiss' κ, 71 pitches at modal n = 21 raters |

The six pairwise SFT κ values lie in [0.500, 0.604] (mean 0.547); the two human-pool Fleiss' κ values lie in [0.030, 0.049]. The directional contrast is robust to the human-truncation choice: the min-truncation variants (κ = 0.104 for the expert panel at n = 2 per pitch on all 120 pitches; κ = 0.033 for the junior panel at n = 15 per pitch on all 120) remain an order of magnitude below the SFT pairwise band. The contrast is descriptive of agreement only — it is not used as a test of whether SFT models recover a *correct* signal (that role belongs to ST22–ST24 and to the accuracy and McNemar comparisons in the main text).

## Supplementary Table 22 (ST22): Per-Class Accuracy under SFT 4/4 Consensus (Management Benchmark)

Where all four SFT architectures agreed, accuracy was sharpest at the quality extremes.

**Table ST22. Per-class accuracy when all four management SFT checkpoints (GPT-4.1, GPT-4.1-nano, Qwen3-30B-A3B, Qwen3-4B) produce the same first-token argmax label. Coverage is the fraction of each tier's 30 ground-truth items on which the four checkpoints unanimously agree; consensus accuracy is exact-tier match within that subset.**

| Ground-truth tier | Tier *N* | Consensus *n* (4/4 agree) | Coverage | Consensus accuracy |
|---|---|---|---|---|
| Exceptional | 30 | 14 | 46.67% | **100.00%** (14/14) |
| Strong | 30 | 12 | 40.00% | 50.00% (6/12) |
| Fair | 30 | 9 | 30.00% | 66.67% (6/9) |
| Limited | 30 | 16 | 53.33% | 62.50% (10/16) |
| **All tiers** | **120** | **51** | **42.50%** | **70.59%** (36/51) |

The all-tiers row summarizes the subset on which all four SFT checkpoints produced the same first-token label. Per-tier breakdown shows that the consensus subset's accuracy is concentrated at the *exceptional* end (14 of 14 correct) with the lowest accuracy at *strong*, where the field-level boundary is hardest. The tier-extreme pattern is consistent with the cross-field selective-prediction result (Figure 4d, SI ST16): high-confidence routing is most reliable when the tier signal is strongest, which on this benchmark is at the field-defining extreme rather than at the field-internal *strong*-versus-*exceptional* boundary.

## Supplementary Table 23 (ST23): AI–Human Error Complementarity and Either-Correct Upper Bound on the Expert-Comparable Subset

SFT and senior-expert errors decompose largely into disjoint subsets; the either-correct upper bound on the expert-comparable subset is 77.53%, indicating substantial complementary signal.

**Table ST23. Joint-error decomposition for the SFT 2-model ensemble (GPT-4.1-nano + Qwen3-30B-A3B) versus the unfiltered 48-expert majority vote on the 89 non-tied management pitches. The either-correct upper-bound row gives the retrospective accuracy obtained by counting an item correct whenever either evaluator was right; it is reported as an upper-bound decomposition rather than as a deployed metric.**

| Cell | Definition | *n* | % of 89 |
|---|---|---|---|
| Both correct | SFT ensemble correct **and** expert majority correct | 24 | 26.97% |
| AI only correct | SFT ensemble correct **and** expert majority wrong | 32 | 35.96% |
| Expert only correct | SFT ensemble wrong **and** expert majority correct | 13 | 14.61% |
| Both wrong | Neither evaluator correct | 20 | 22.47% |
| **Either-correct upper bound** | Either correct (= both-correct + AI-only + expert-only) | 69 | **77.53%** |
| — | — | — | — |
| SFT ensemble alone | (both-correct + AI-only) / 89 | 56 | 62.92% |
| Expert majority alone | (both-correct + expert-only) / 89 | 37 | 41.57% |

The disjoint-correct cells (AI-only-correct = 32; expert-only-correct = 13) are larger than the joint-correct cell (24) and only slightly smaller than the joint-wrong cell (20), so the two evaluator classes do not collapse to a single shared signal even where they are individually correct. The 14.6 percentage-point gap between the either-correct upper bound (77.53%) and SFT-ensemble-alone (62.92%) is a retrospective bound on the complementary signal observed in this subset; it does not specify an implementable routing rule.

## Supplementary Table 24 (ST24): Pairwise McNemar Tests on the Management Benchmark

Pairwise paired tests on the same items confirm the SFT advantage is significant against every comparator class.

**Table ST24. Pairwise McNemar tests of the primary SFT 2-model ensemble (GPT-4.1-nano + Qwen3-30B-A3B) against four comparator classes on the management benchmark. Each test compares the ensemble's correctness vector against the comparator's correctness vector on the paired subset where both predictions are defined; *N* is that paired subset size. McNemar $\chi^2$ uses the continuity-corrected statistic on the discordant cells. Raw two-sided *P* values are reported; with Bonferroni correction across the four contrasts, all four still reject $H_0$ at $\alpha = 0.05$.**

| Comparator | Paired *N* | SFT ensemble accuracy | Comparator accuracy | Δ (pp) | McNemar $\chi^2$ | Raw *P* |
|---|---|---|---|---|---|---|
| Frontier mean (per-article plurality of 11 reasoning models) | 114 | 61.40% | 38.60% | +22.81 | 10.78 | **0.00103** |
| Best frontier (Gemini 3.1 Pro, single-shot per pitch) | 120 | 60.83% | 43.33% | +17.50 | 6.78 | **0.00922** |
| Expert majority vote (unfiltered, ties excluded) | 89 | 62.92% | 41.57% | +21.35 | 7.20 | **0.00729** |
| Junior majority vote (filtered, ties excluded) | 103 | 62.14% | 40.78% | +21.36 | 9.59 | **0.00196** |

The frontier-mean row uses per-article plurality across the 11-model reasoning cohort as the comparator (alphabetical tie-break, ties excluded from the paired subset, $N = 114$); this is the operationalization that supports a paired McNemar test, distinct from the scalar 11-model individual-mean accuracy reported in the main text (which cannot be paired-tested). The best-frontier row pairs SFT against single-shot Gemini 3.1 Pro per-pitch predictions and is therefore reported in the same single-shot space as the McNemar comparator rather than the eight-sample pitch-mean used for the headline frontier comparison in §2 of the main text. Both human-majority rows pair SFT against the unfiltered-expert and filtered-junior majority votes already reported in §2 and SM3.

*Cohort and protocol footnote.* ST24 frontier comparators are computed on single-shot per-pitch predictions from the same 11-model reasoning cohort used in main-text Section 2. The 31.1% frontier-mean and 38.8% best-frontier figures cited in main text §2 use eight-sample pitch-mean aggregation per Methods. The two are different operationalisations of “frontier accuracy” — single-shot per-pitch predictions are required for paired McNemar testing, whereas the eight-sample pitch-mean is the primary headline metric — rather than competing estimates of the same quantity, and neither is reducible to the other.

## Supplementary Figures

### Supplementary Figure 1 (SF1): Cross-Field SFT vs Control Accuracy (Eight Fields)

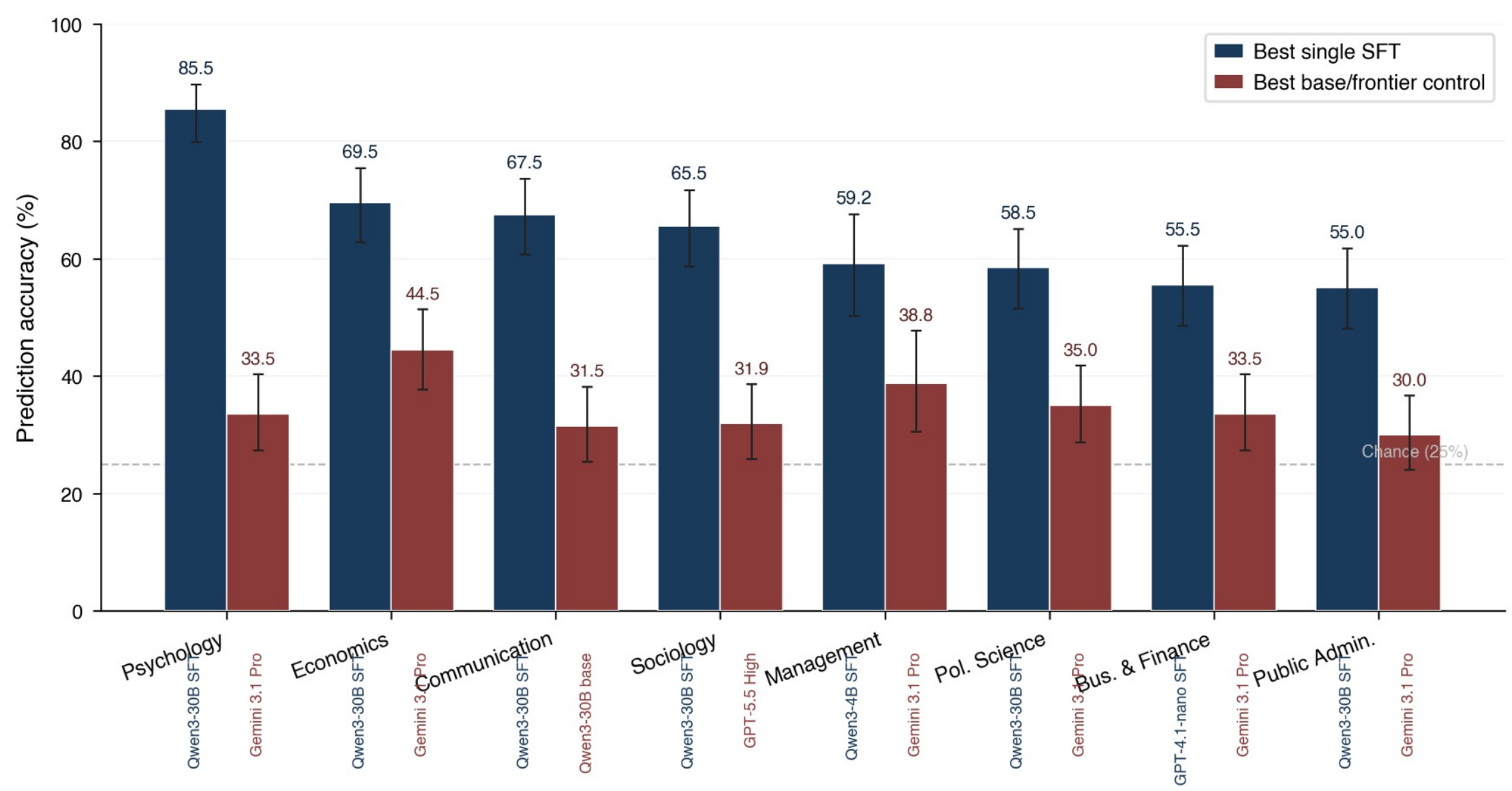

**Supplementary Figure SF1 | Cross-field SFT vs control accuracy across eight social-science fields.** Per-field prediction accuracy on the held-out four-tier benchmark for the best single SFT model (blue) and the best base/frontier control (red), ordered top-to-bottom by SFT accuracy. The model name underneath each bar identifies which architecture supplied the best SFT and the best control. Error bars are 95% Wilson binomial intervals on $n = 200$ articles per field except management ($n = 120$). Dashed horizontal line, chance (25%). The best single SFT is Qwen3-30B-A3B in six fields, Qwen3-4B in management, and GPT-4.1-nano in business and finance; the best control is Gemini 3.1 Pro in six fields, GPT-5.5 High in sociology, and the architecture-matched Qwen3-30B base in communication. Numbers replot Supplementary Table ST1.

### Supplementary Figure 2 (SF2): SFT-vs-Base Lift Heatmap (Eight Fields × Three Architectures)

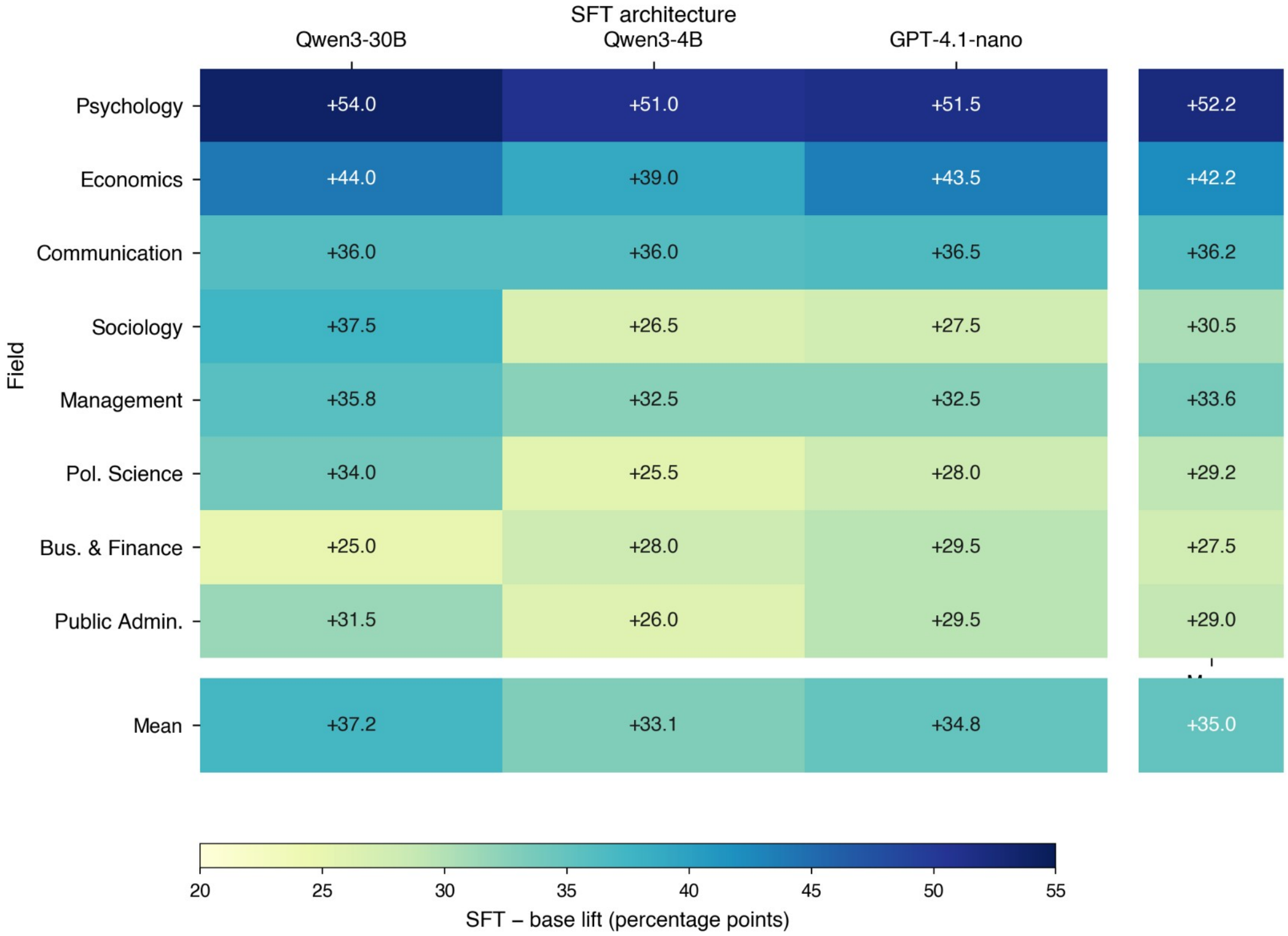

**Supplementary Figure SF2 | SFT-vs-base lift heatmap (8 fields × 3 architectures).**
Each cell is the SFT − architecture-matched base accuracy lift, in percentage points, on the held-out four-tier benchmark. Rows are fields ordered by descending best-SFT accuracy; columns are the three SFT architectures (Qwen3-30B-A3B, Qwen3-4B, GPT-4.1-nano). Right-edge "Mean" strip shows the cross-architecture mean lift per field; bottom "Mean" strip shows the cross-field mean lift per architecture; corner cell shows the grand mean over the 24 (architecture, field) cells. The colour ramp saturates between 20 and 55 percentage points. Numbers replot Supplementary Table ST13.

## Supplementary Figure 3 (SF3): Management-SFT Cross-Field Transfer Matrix (Four Architectures × Seven Non-Management Fields)

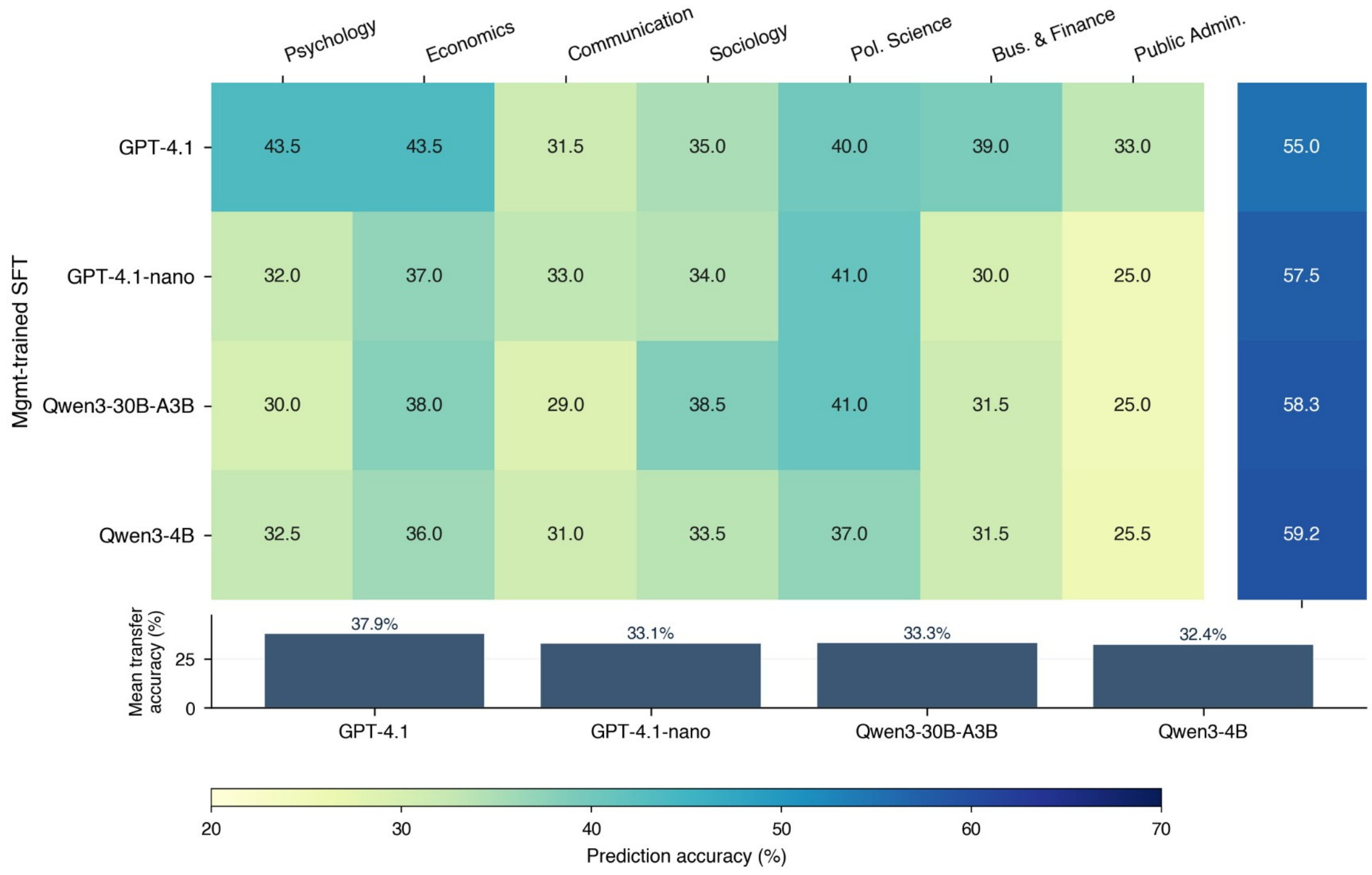

**Supplementary Figure SF3 | Management-SFT cross-field transfer matrix (4 architectures × 7 non-management fields).**

Each main-grid cell is the prediction accuracy of a management-only-trained SFT checkpoint evaluated zero-shot on a non-management field's held-out validation set (n = 200 per cell). Rows are the four management-SFT architectures (GPT-4.1, GPT-4.1-nano, Qwen3-30B-A3B, Qwen3-4B); columns are the seven non-management fields. The right-edge anchor strip shows in-domain management accuracy (n = 120) for the same architecture. The bottom bar gives the mean transfer accuracy across the seven non-management fields per architecture. The colour ramp saturates between 20% and 70%. GPT-4.1, the largest architecture, achieves the highest mean transfer (37.9%) and the smallest in-domain -> out-of-domain drop-off (-17.1 pp). Numbers replot the four panels of Supplementary Table ST17.

## Supplementary Figure 4 (SF4): Per-Field Reliability Diagrams for the Best-Calibrated SFT Architecture (GPT-4.1-nano)

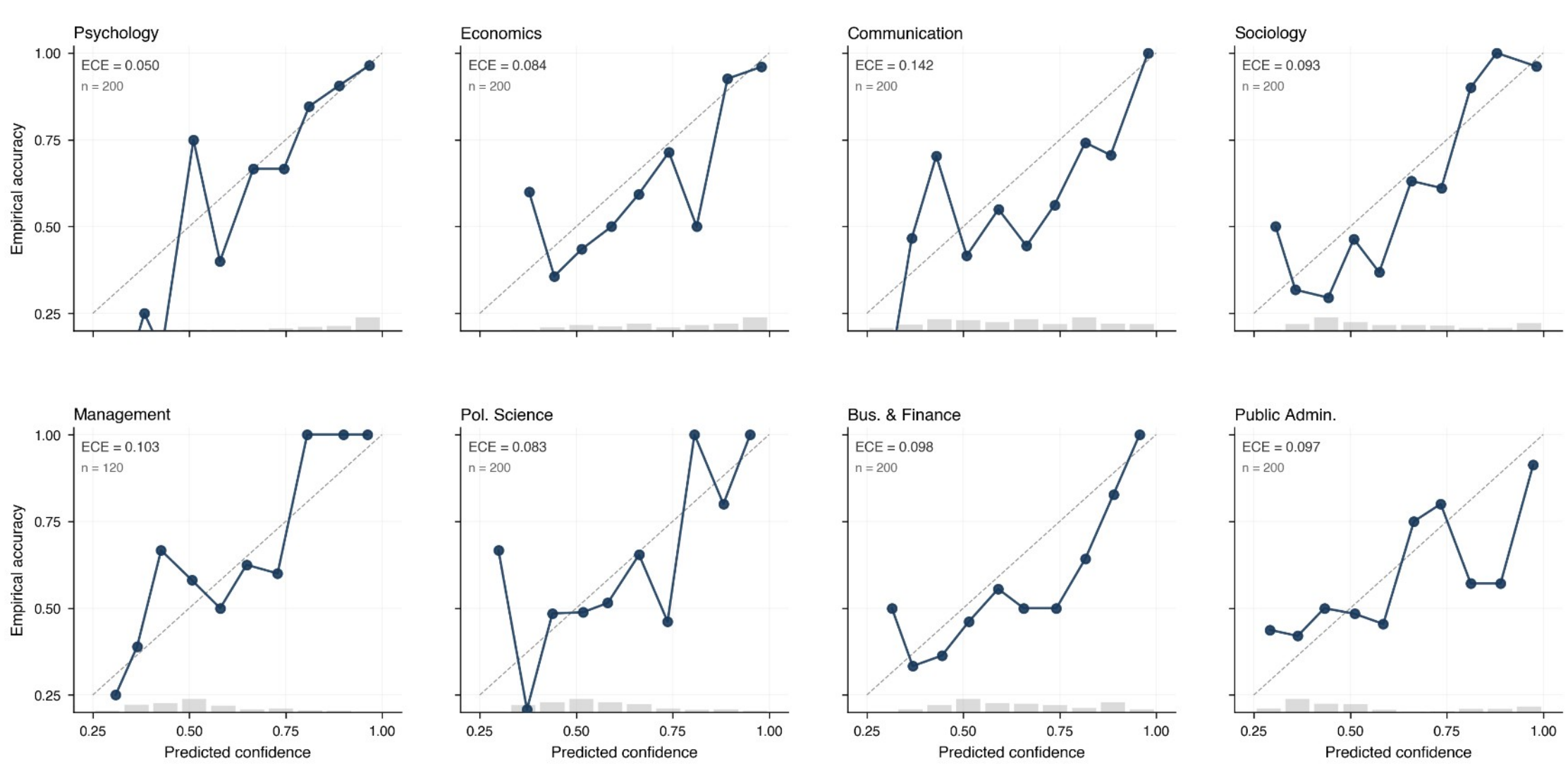

**Supplementary Figure SF4 | Per-field reliability diagrams for the best-calibrated SFT architecture (GPT-4.1-nano).**

Eight panels, one per social-science field, ordered as in main-text figures. Each panel plots the empirical accuracy in 10 equal-width confidence bins on the [0.25, 1.00] interval, where confidence is exp(max log-probability) over the four-class restricted vocabulary. Filled circles connect the (mean confidence, empirical accuracy) value of each non-empty bin; the dashed grey line is the y = x identity (perfect calibration); small grey histogram bars at the bottom of each panel are the per-bin counts. Per-field expected calibration error (ECE) and the n per panel are printed in the upper-left corner. n = 200 per field except management (n = 120). The flatter, near-diagonal trace in psychology and the steeper, more dispersed trace in management track the per-field ECE values reported in Supplementary Table ST2.

## Supplementary Figure 5 (SF5): Selective-Prediction Coverage Curves (Three SFT Architectures × Eight Fields)

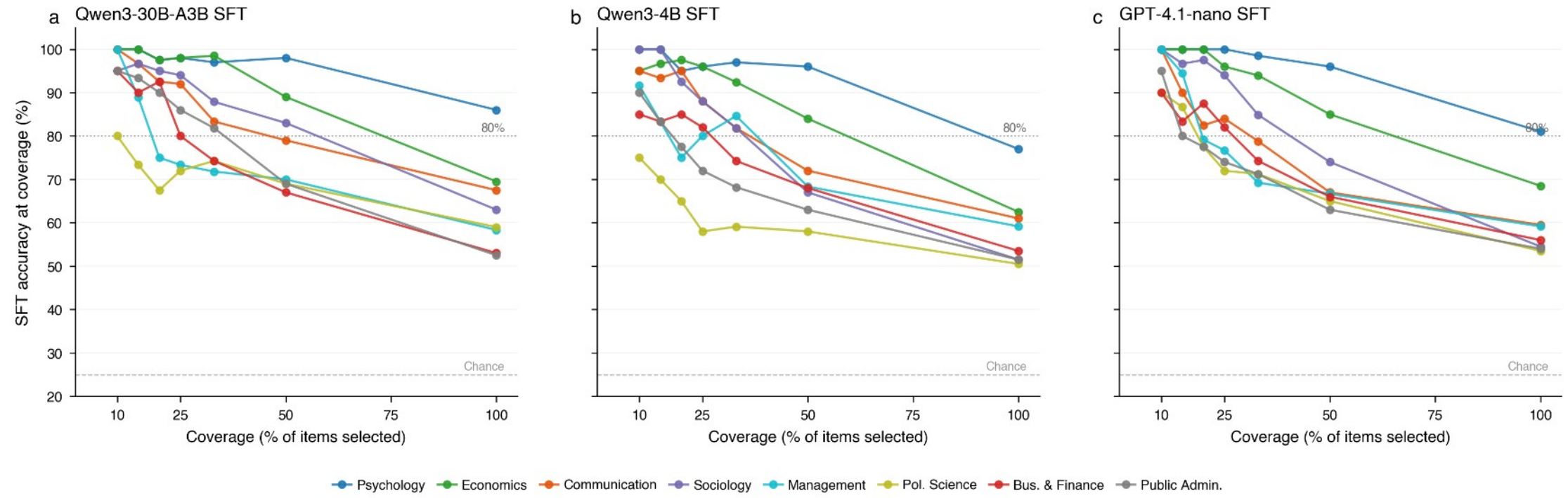

**Supplementary Figure SF5 | Selective-prediction coverage curves (3 SFT architectures × 8 fields).**

a–c, One panel per SFT architecture (Qwen3-30B-A3B, Qwen3-4B, GPT-4.1-nano). Within each panel, every line is one of the eight fields. The x-axis is coverage: the percentage of held-out items selected, in descending order of confidence; the y-axis is prediction accuracy on the selected subset. Confidence is exp(max log-probability) over the four-class restricted vocabulary. Markers sit at the discrete coverage thresholds tabulated in Supplementary Table ST16 (10, 15, 20, 25, 33, 50, 100%); the dotted horizontal reference is at 80% accuracy and the dashed reference is at chance (25%). The dominant pattern is that selective prediction recovers >=80% accuracy at modest coverage in the majority of (architecture, field) cells. n = 200 per non-management field, n = 120 in management.

## Supplementary Figure 6 (SF6): Prediction-Distribution Comparison Across Evaluator Classes (Management Benchmark)

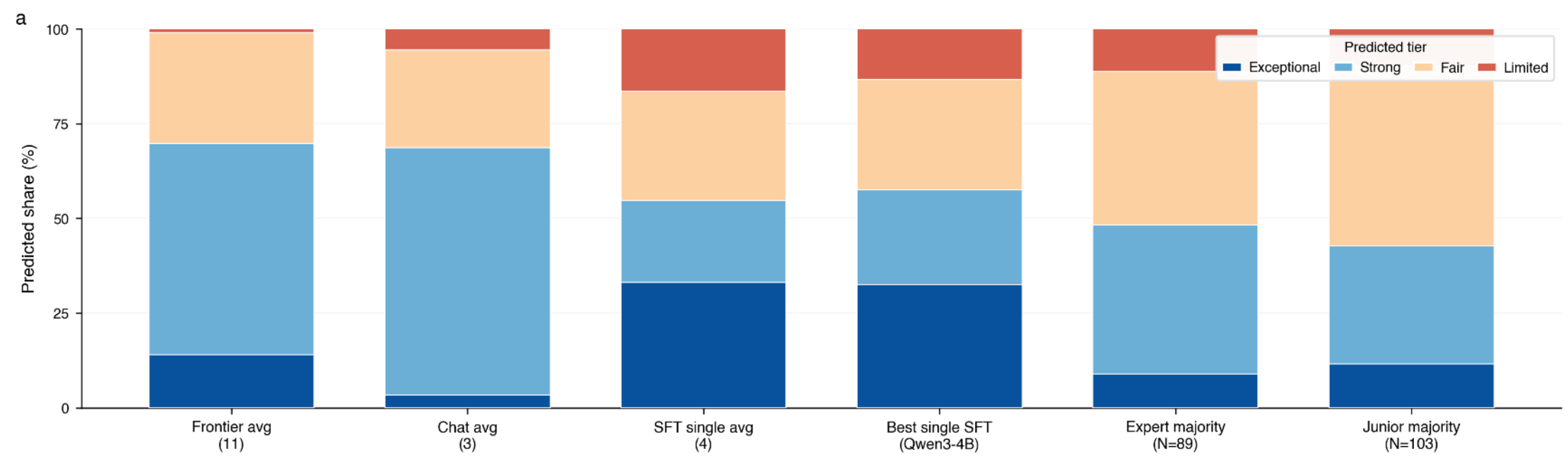

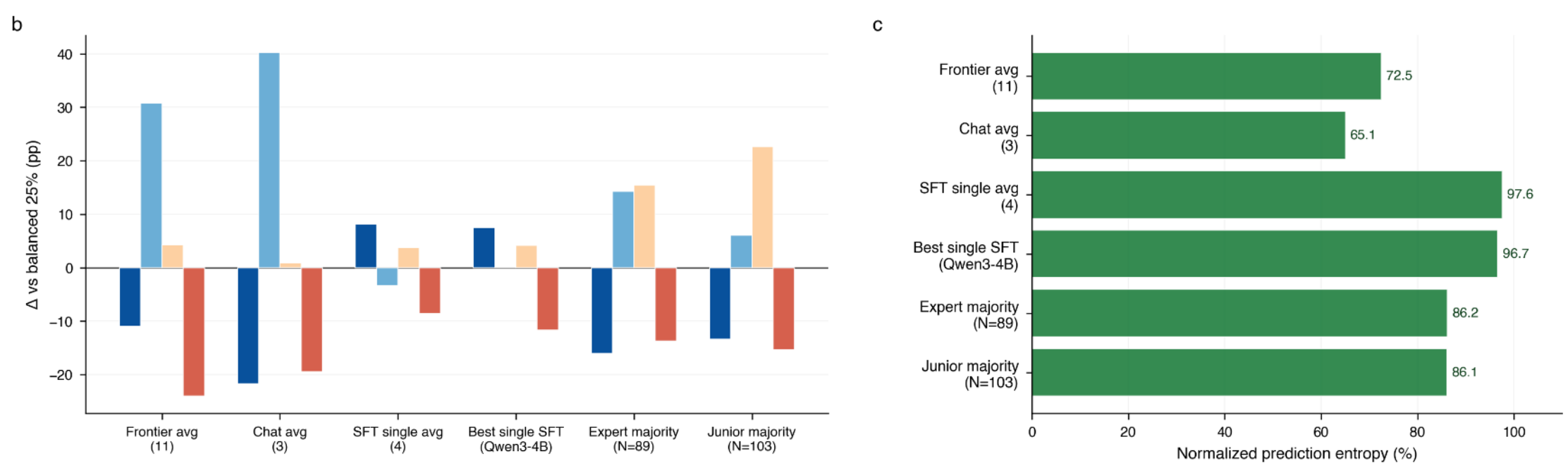

**Supplementary Figure SF6 | Prediction-distribution comparison across evaluator classes (management benchmark).**

a, 100%-stacked predicted-tier shares for six evaluator classes — frontier average (11), chat average (3), four-SFT-single average, best single SFT (Qwen3-4B), expert majority vote (N = 89 non-tied), junior majority vote (N = 103 non-tied). b, Tier-bias profile: deviation in percentage points from the balanced 25%-per-tier reference, by tier within each class. c, Distribution-concentration index: normalized Shannon entropy of the predicted-tier distribution per class (0% = single-tier output, 100% = uniform). The four-SFT-single average and best single SFT approach 100% normalised entropy and a near-zero tier-bias signature, while frontier-average, chat-average, and both human majority votes pile predictions on the middle tiers. n = 120 management articles per AI class.

## Supplementary Figure 7 (SF7): Frontier-Collapse-Metric Landscape (Eleven Frontier Reasoning Models)

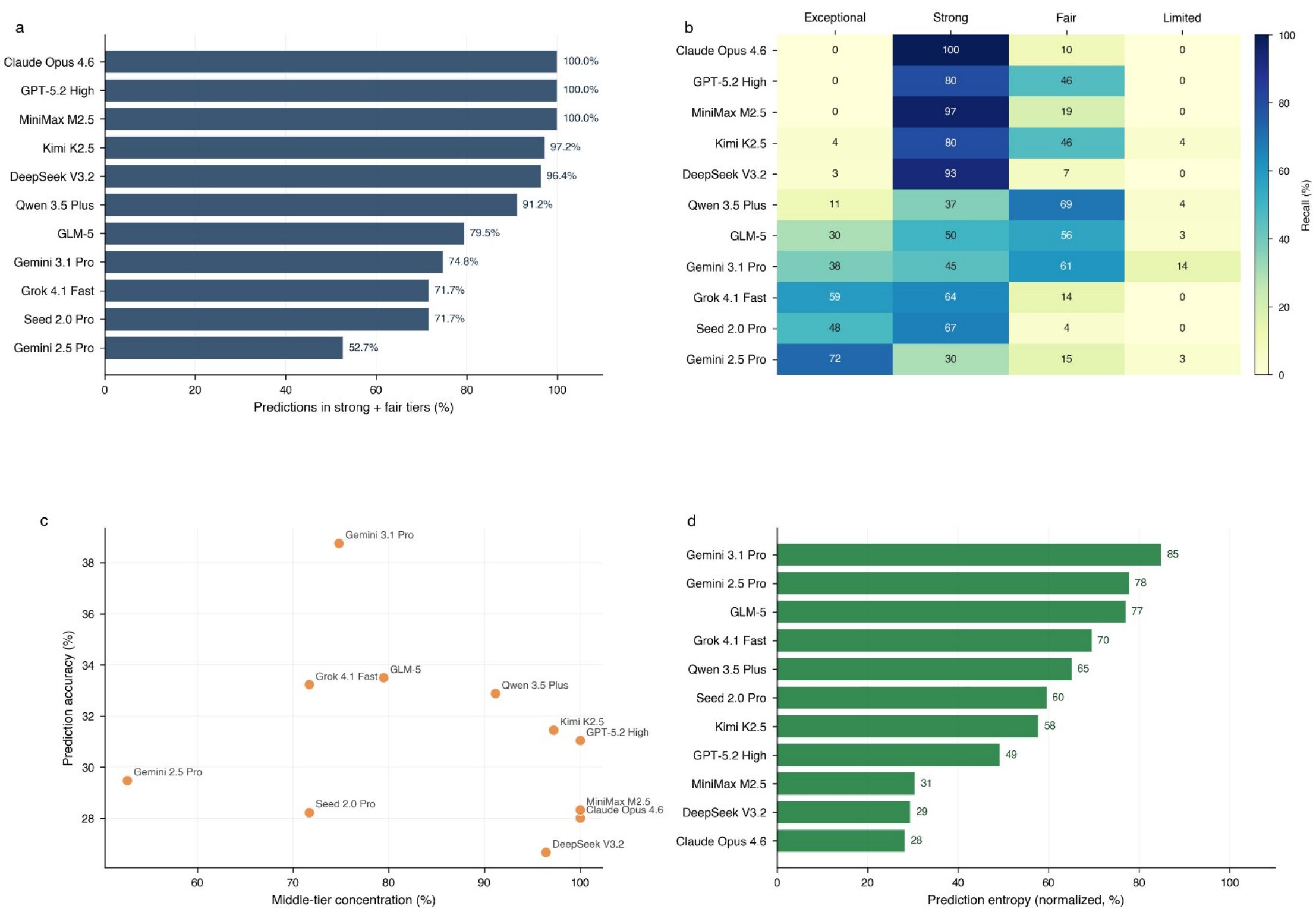

**Supplementary Figure SF7 | Frontier-collapse-metric landscape (eleven frontier reasoning models on the management benchmark).**

a, Middle-tier concentration: percentage of each model's predictions falling into the *strong* + *fair* tiers, sorted descending. b, Tier-recall heatmap: per-tier recall (%) for each model, rows ordered as in a. c, Concentration-vs-accuracy scatter: each marker is one frontier model. d, Distribution-balance ranking: normalized Shannon entropy of the predicted-tier distribution per model. n = 120 per model. Numbers are derived from the pre-specified eight-sample frontier aggregation protocol used elsewhere in the manuscript; per-model accuracy, macro-F1, and sample sizes are listed in the Data Availability statement.

## Supplementary Figure 8 (SF8): Per-Field Confusion Matrices for the Best Single SFT Model in Each of the Eight Fields

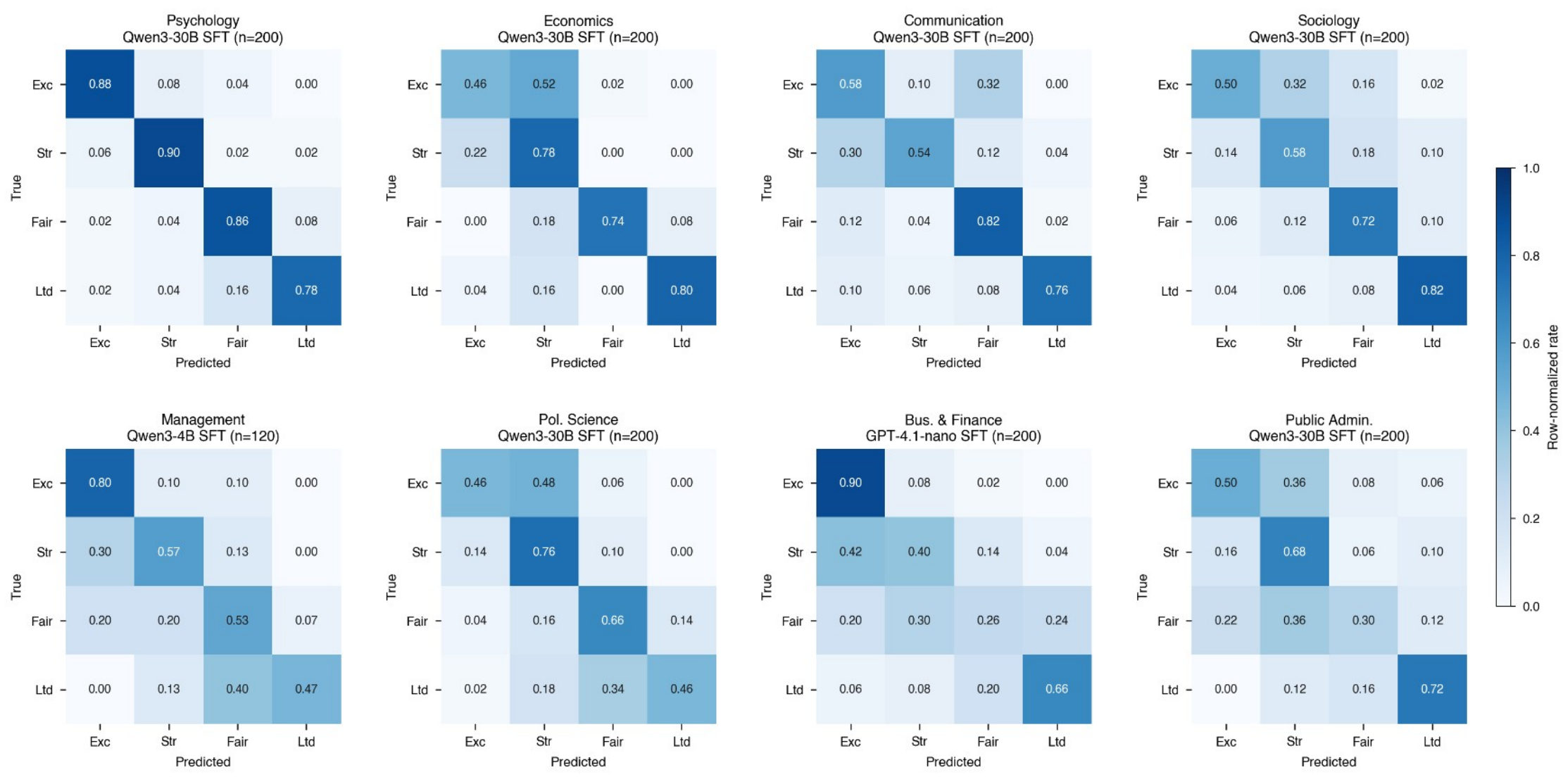

**Supplementary Figure SF8 | Per-field confusion matrices for the best single SFT model in each of the eight fields.**

Eight 4×4 row-normalized confusion matrices, one per field. Rows are the ground-truth tier (exceptional → limited); columns are the SFT model's chosen prediction. Cells are within-row predicted-class rates; each row sums to 1 by construction. Diagonal entries are tier-level recall from the same prediction resolver used for Supplementary Table ST12. The per-field best single SFT is shown above each panel: Qwen3-4B SFT for management ($n$ = 120); GPT-4.1-nano SFT for business and finance ($n$ = 200); Qwen3-30B-A3B SFT for the remaining six fields ($n$ = 200 each). Cell colour is shared across all eight panels (saturating between 0 and 1). The dominant per-tier confusions concentrate on adjacent-tier substitutions (strong → fair and fair → strong), consistent with the field-level Top-1+2 diagnostic reported in Supplementary Table ST15. A systematic *under-prediction of exceptional combined with over-prediction of strong* is visible in three of the seven cross-field cells whose best single SFT is Qwen3-30B-A3B — economics, political science and public administration — manifested as exceptional-row recall ≤ 0.50 paired with 76–82 articles (out of 200) routed to strong. Psychology, sociology, communication, business and finance, and management do not show this pattern (their exceptional-row recall is 0.50–0.90 with markedly lower strong-column volume).

## Supplementary Figure 9 (SF9): Cross-Field Ensemble Sensitivity

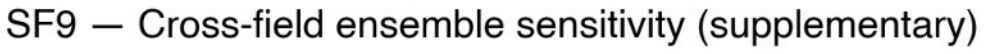

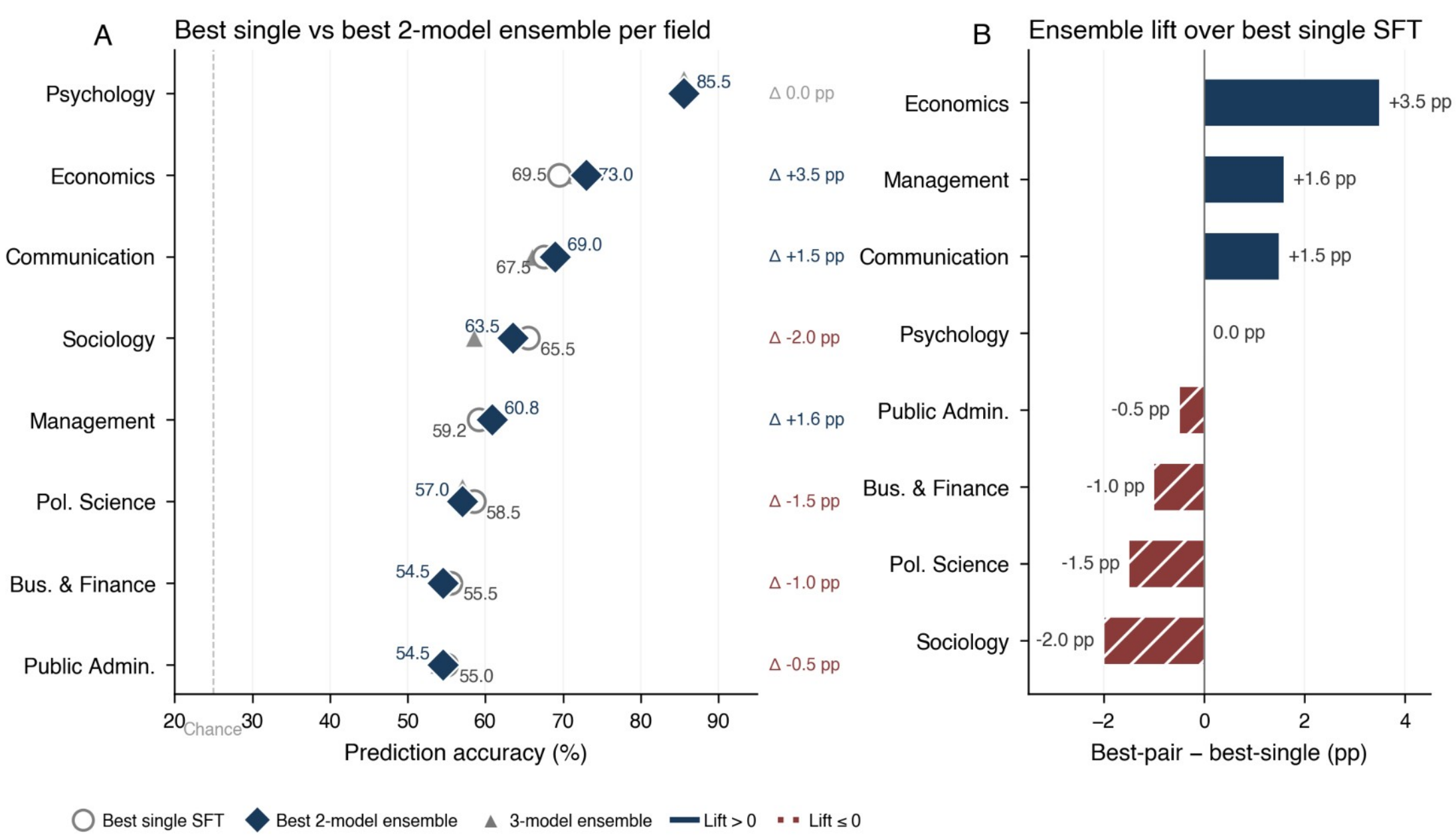

**Supplementary Figure SF9 | Ensemble lift is small and field-specific.**

a, Per-field dumbbell plot comparing best single-model SFT accuracy to best two-model pair-ensemble accuracy, with the three-model ensemble shown as a marker. Probability averaging is the aggregator (SM7). b, Pair-ensemble lift over the best single SFT model. Management (+1.6 pp), economics (+3.5 pp), and communication (+1.5 pp) improve modestly; the remaining five fields are flat or negative. The primary cross-field evaluator is therefore the best single SFT model per field, and the two-model ensemble is reported only as a sensitivity analysis.

## Supplementary Figure 10 (SF10): Frontier Cohort Diagnostics (11 Models)

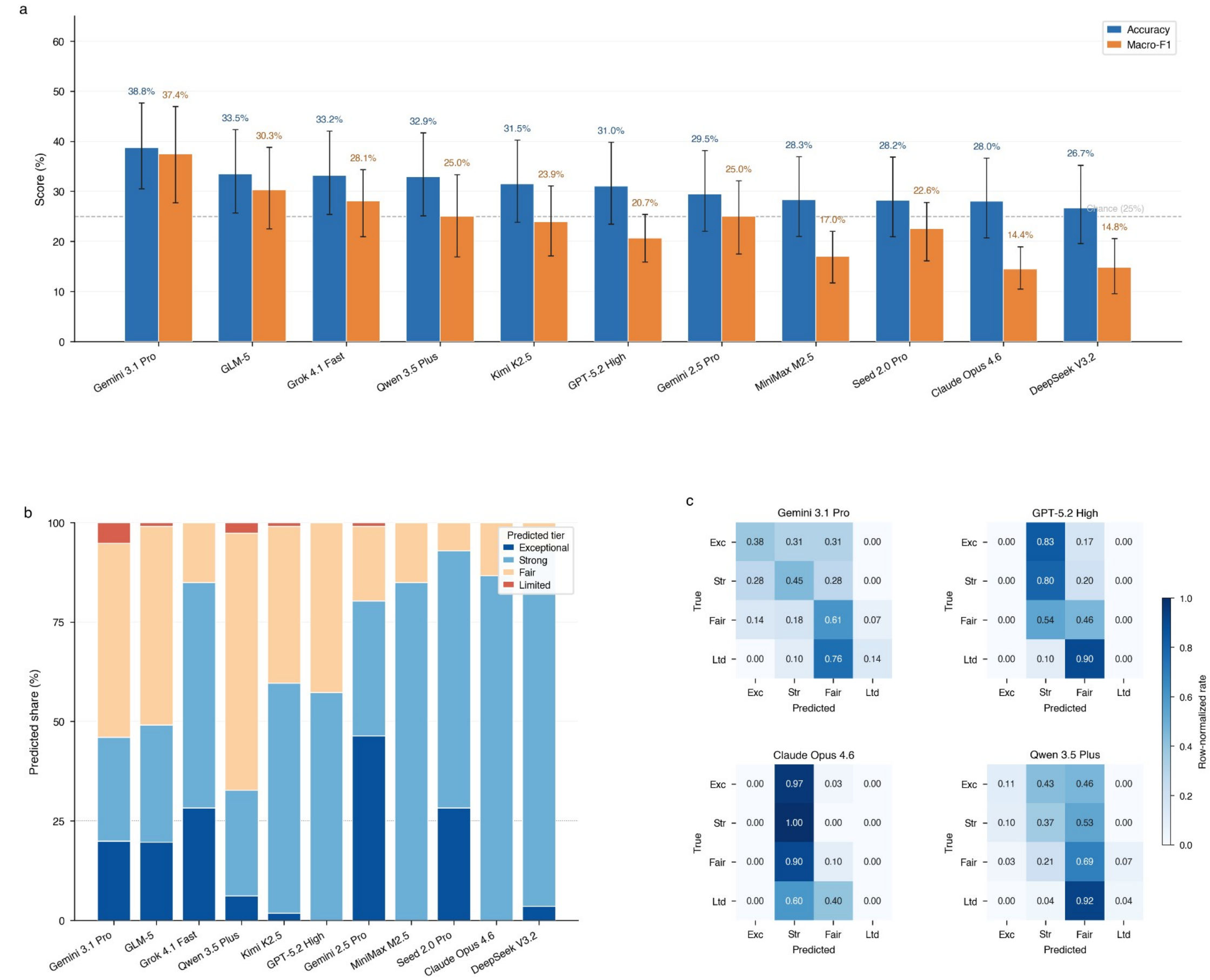

**Supplementary Figure SF10 | Frontier-model deep-dive on the 120-article management benchmark.**

a, Leaderboard for the eleven frontier systems benchmarked in the main text (Gemini 3.1 Pro, GLM-5, Grok 4.1 Fast, Qwen 3.5 Plus, Kimi K2.5, GPT-5.2 High, Gemini 2.5 Pro, MiniMax M2.5, Seed 2.0 Pro, Claude Opus 4.6, DeepSeek V3.2), sorted by prediction accuracy (descending). Paired bars show accuracy (blue) and macro-F1 (orange) on the four-tier classification (*exceptional*, *strong*, *fair*, *limited*); error bars are 95% Wilson binomial intervals on accuracy and 95% percentile bootstrap intervals on macro-F1 (1,000 draws, seed 42; n = 120 per model). Dashed horizontal line, chance (25%). b, 100%-stacked predicted-tier distribution for the same eleven models in the same order, illustrating systematic mid-tier piling and the rarity of *limited* predictions across the frontier (n per model = 109–120 valid discrete predictions after tie removal). c, Row-normalized confusion matrices for four representatives covering the dominant collapse modes: Gemini 3.1 Pro (best frontier), GPT-5.2 High (mid-tier clustering), Claude Opus 4.6 (strong-ceiling) and Qwen 3.5 Plus (open-weight reference). Cell values give the within-row predicted-class rate (each row sums to 1); the colour ramp is shared across panels and saturates between 0 and 1. Numbers are derived from the pre-specified eight-sample frontier aggregation protocol used elsewhere in the manuscript; sample sizes are the primary benchmark size and the valid discrete-prediction count after tie removal. Per-model accuracy, macro-F1, and sample sizes are listed in the Data Availability statement.

## Supplementary Figure 11 (SF11): GPT-5.5 All-Field Evaluation

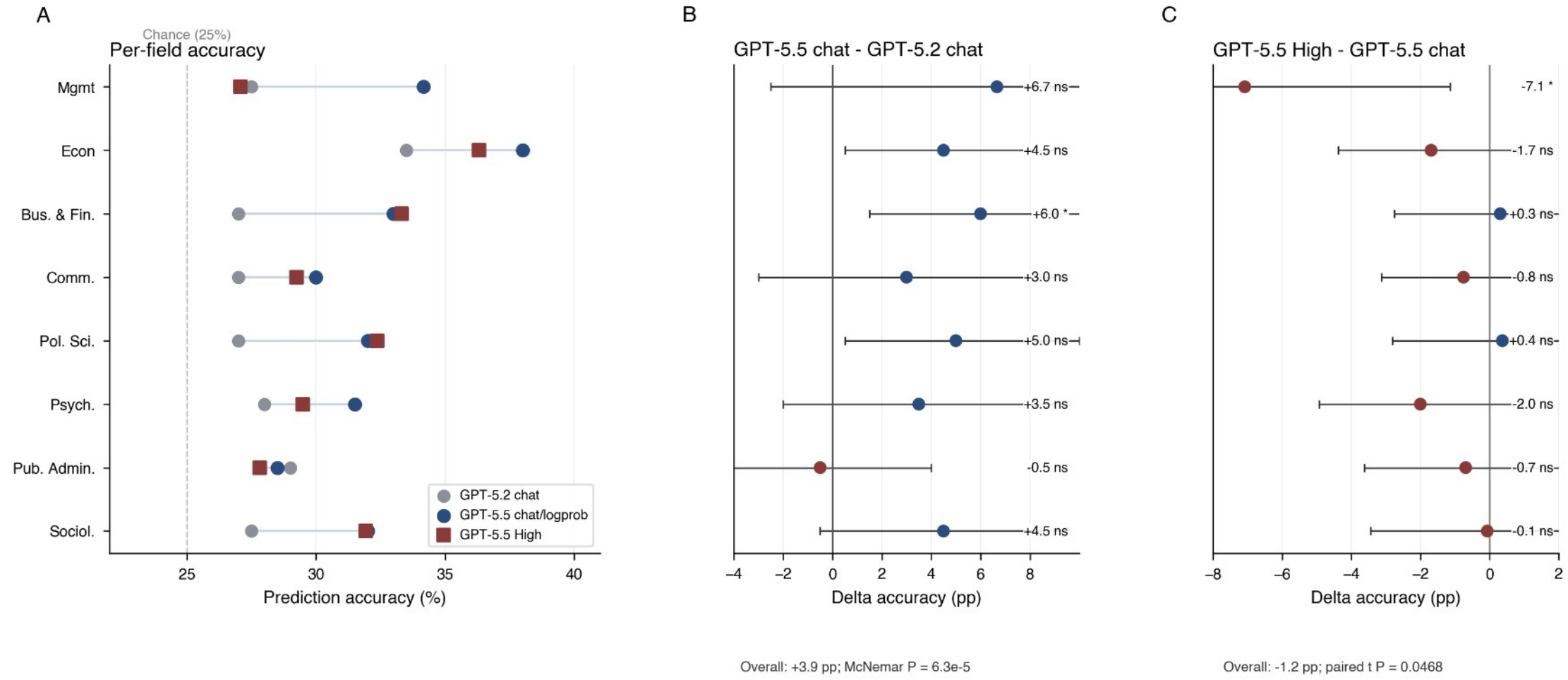

**Supplementary Figure SF11 | GPT-5.5 all-field evaluation.**

a, Per-field prediction accuracy on the held-out four-tier benchmarks for the historical GPT-5.2 chat/log-probability comparator, GPT-5.5 chat/log-probability classification with reasoning disabled, and GPT-5.5 High pitch-mean accuracy from eight high-reasoning samples per item. Dashed vertical line, chance (25%). b, Field-level change from GPT-5.2 chat/log-probability to GPT-5.5 chat/log-probability; seven non-management rows use the same general prompt, while management combines model change with the expert-rubric prompt used for the main management frontier benchmark. Points and horizontal whiskers show paired item-level mean differences and 95% paired-bootstrap confidence intervals; significance labels use two-sided exact McNemar tests on paired correctness (asterisk, $P < 0.05$; ns, $P >= 0.05$). Because the intervals and labels use different paired procedures, significance should be read from the stated test rather than inferred from whether the bootstrap interval crosses zero. The item-weighted all-field chat/log-probability change was +3.9 percentage points (95% CI +2.0 to +5.9; McNemar $P$ = 6.3e-5). c, Field-level change from GPT-5.5 chat/log-probability to GPT-5.5 High pitch-mean accuracy. Points and horizontal whiskers show paired item-level mean differences and 95% paired-bootstrap confidence intervals; significance labels use paired $t$ tests on item-level high pitch-mean accuracy minus chat correctness. Across all 1,520 items, GPT-5.5 chat/log-probability accuracy was 32.3% item-weighted versus 31.2% for GPT-5.5 High pitch-mean accuracy (difference -1.2 percentage points; 95% CI -2.3 to 0.0; paired $t$ test $P$ = 0.0468). No all-field GPT-5.2 High comparison is plotted or claimed.

## Supplementary References

## Data Availability

All datasets, model predictions, training-corpus summaries, journal-tier mappings, summary statistics, and analysis scripts required to reproduce the reported numbers in this Supplementary Information are provided in the accompanying public reproducibility package (https://github.com/FutureTech-OB/ai-taste). The release includes (i) the held-out research-idea pitch corpora and per-pitch tier labels for all eight fields (management N = 120; each non-management field N = 200); (ii) per-(model, field) prediction records, including SFT, base, chat/log-probability, GPT-5.5 high-reasoning, and management RL mechanism-probe outputs; (iii) public model aliases and provider/model access information where redistribution is permitted, with provider-specific checkpoint identifiers and model weights not redistributed where licensing or provider terms prohibit redistribution; (iv) summary statistics files supporting each Supplementary Table and Figure; and (v) validation and reproduction scripts for the released machine-readable tables and figure asset bundle. Data are released under CC BY 4.0 and code under MIT.